\documentclass[letterpaper]{article} % DO NOT CHANGE THIS
\usepackage[preprint]{aaai2027}
\usepackage[hyphens]{url}  % DO NOT CHANGE THIS
\usepackage{graphicx} % DO NOT CHANGE THIS
\usepackage{natbib}  % DO NOT CHANGE THIS AND DO NOT ADD ANY OPTIONS TO IT
\usepackage{caption} % DO NOT CHANGE THIS AND DO NOT ADD ANY OPTIONS TO IT
\usepackage{algorithm}
\usepackage{algorithmic}

\usepackage{newfloat}
\usepackage{listings}
\DeclareCaptionStyle{ruled}{labelfont=normalfont,labelsep=colon,strut=off} % DO NOT CHANGE THIS
\floatstyle{ruled}
\newfloat{listing}{tb}{lst}{}
\floatname{listing}{Listing}

\usepackage{booktabs}

\usepackage{microtype}
\usepackage{graphicx}
\usepackage{subfigure}
\usepackage{booktabs} % for professional tables

\usepackage[most]{tcolorbox} 
\usepackage{algorithm}
\usepackage{algorithmic}
\definecolor{algobackground}{RGB}{245, 248, 250} 
\definecolor{algoframeline}{RGB}{210, 220, 230}
\usepackage[most]{tcolorbox}
\definecolor{algoback}{RGB}{248, 250, 252} 
\definecolor{algoframe}{RGB}{71, 85, 105} 
\usepackage{amsmath}
\usepackage{amssymb}
\usepackage{mathtools}
\usepackage{amsthm}
\usepackage{caption}
\usepackage[capitalize,noabbrev]{cleveref}
\usepackage{threeparttable} 
\usepackage{multirow}
\usepackage{colortbl}  
\usepackage{footmisc} 
\usepackage{newfloat}
\usepackage{listings}
\usepackage{makecell}
\usepackage{amsmath, amsfonts, bm}
\usepackage{algorithm}
\usepackage{algorithmic}
\usepackage{amssymb, bbding}
\usepackage{longtable}
\usepackage{subcaption} 

\usepackage{marvosym}
\usepackage{xcolor}
\definecolor{DeepBlue}{RGB}{46,84,151}
\definecolor{SoftBrown}{RGB}{186,145,103}
\definecolor{DarkGreen}{RGB}{84,130,52}
\usepackage{arydshln}
\usepackage{makecell}
\usepackage{multirow}
\usepackage{nicefrac}       % compact symbols for 1/2, etc.
\usepackage{dsfont}
\usepackage{caption}
\usepackage{tabularx}
\usepackage{enumerate}
\usepackage{thmtools}
\usepackage{thm-restate}
\usepackage{tcolorbox}

\tcolorboxenvironment{datageneration}{
  colback=rgb{0.7765,0.9373,0.8078}
  boxrule=1pt, % Change this value to set the border thickness
  boxsep=1pt,
  left=2pt,right=2pt,top=2pt,bottom=2pt,
  oversize=2pt,
  sharp corners,
  before skip=\topsep,
  after skip=\topsep,
}

\tcbset{
  mybox/.style={
    colback=blue!5,
    boxrule=1pt, % Change this value to set the border thickness
    boxsep=1pt,
    left=10pt, % Increased left margin
    right=10pt, % Increased right margin
    top=10pt, % Increased top margin
    bottom=10pt, % Increased bottom margin
    oversize=2pt,
    arc=4pt, % Set this value to adjust the radius of the corners
    before skip=10pt, % Increased space before the box
    after skip=10pt, % Increased space after the box
    baseline=3.5ex % Adjust this value to set line spacing
  }
}

\newcommand{\eqmark}{\textsuperscript{\dag}}
\newcommand{\coremark}{\textsuperscript{\ddag}}
\newcommand{\corrmark}{\textsuperscript{*}}
\newcounter{corefn}
\newcommand{\corecontrib}{%
  \ifnum\value{corefn}=0
    \footnote{These authors are core experimental contributors.}%
    \setcounter{corefn}{\value{footnote}}%
  \else
    \footnotemark[\value{corefn}]%
  \fi
}

\theoremstyle{plain}
\newtheorem{theorem}{Theorem}[section]

\theoremstyle{definition}

\theoremstyle{remark}

\usepackage[textsize=tiny]{todonotes}
\usepackage{url}
\usepackage{fontawesome}

\newcommand{\pdflink}[2]{%
  \leavevmode
  \pdfstartlink attr{/Border [0 0 0]} user {/Subtype /Link /A << /S /URI /URI (#1) >>}%
  #2%
  \pdfendlink
}

\NewDocumentCommand\emojidizzy{}{
  \includegraphics[height=1.8em]{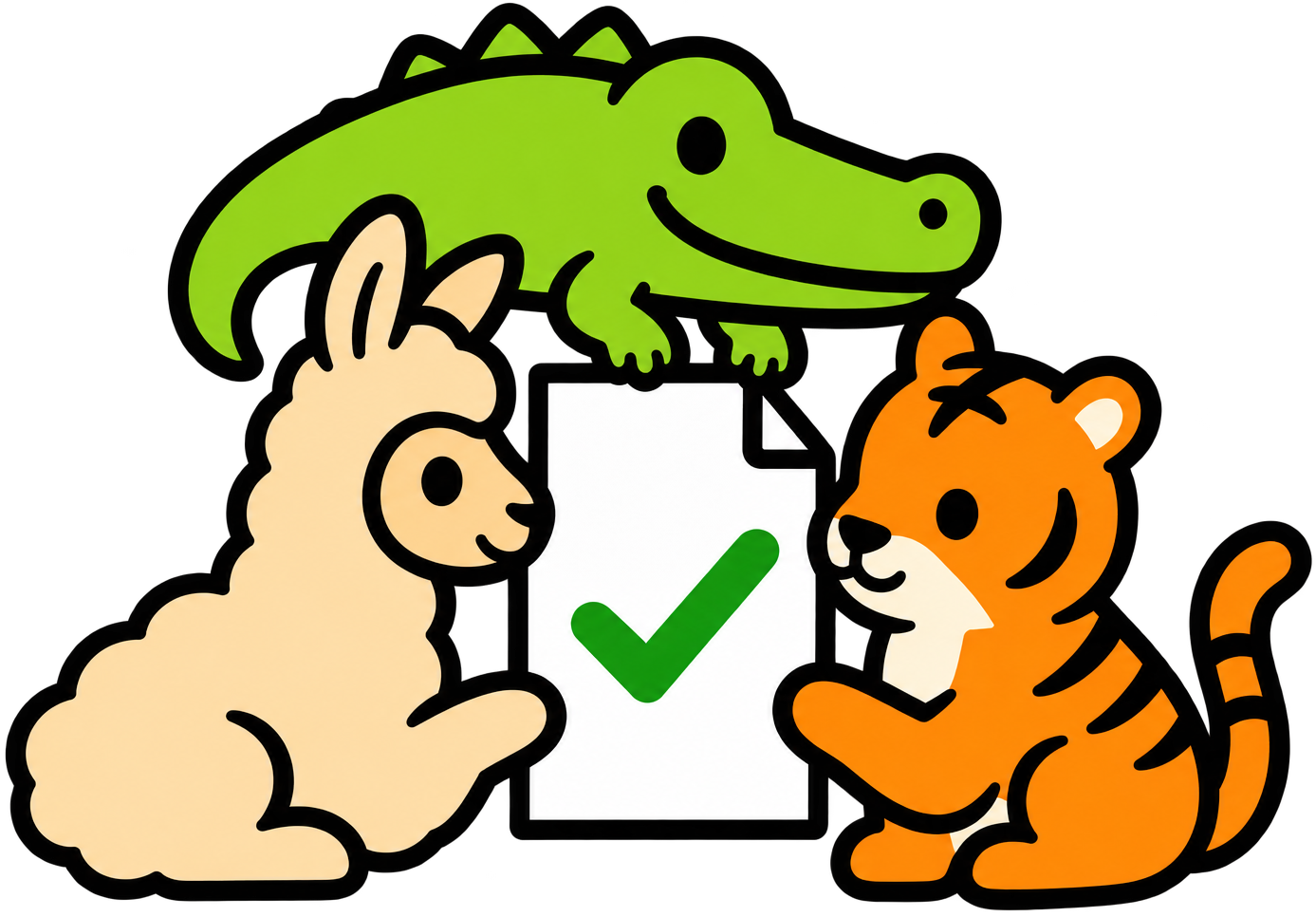}
}

\title{\emojidizzy Scoring, Reasoning, and Selecting the Best! \\
            Ensembling Large Language Models via a Peer-Review Process}

\author{
\normalsize
Zhijun Chen\textsuperscript{\rm 1, 2}\eqmark,
Zeyu Ji\textsuperscript{\rm 1}\eqmark\coremark,
Qianren Mao\textsuperscript{\rm 3},
Hao Wu\textsuperscript{\rm 4},
Jinhuan Song\textsuperscript{\rm 5}\coremark,
Junhang Cheng\textsuperscript{\rm 1},
Bangjie Qin\textsuperscript{\rm 6},
Zhuoran Li\textsuperscript{\rm 1},
Jingzheng Li\textsuperscript{\rm 3},
Kai Sun\textsuperscript{\rm 4},
Zizhe Wang\textsuperscript{\rm 7},
Yikun Ban\textsuperscript{\rm 1},
Zhu Sun\textsuperscript{\rm 8},
Xiangyang Ji\textsuperscript{\rm 7},
Hailong Sun\textsuperscript{\rm 1}\corrmark,
Xiao Huang\textsuperscript{\rm 2}
}
\affiliations{
\small
\textsuperscript{\rm 1}Beihang University, Beijing, China\\
\textsuperscript{\rm 2}The Hong Kong Polytechnic University, Hong Kong, China\\
\textsuperscript{\rm 3}Zhongguancun Laboratory, Beijing, China\\
\textsuperscript{\rm 4}Xi'an Jiaotong University, Xi'an, China\\
\textsuperscript{\rm 5}Beijing University of Posts and Telecommunications, Beijing, China\\
\textsuperscript{\rm 6}Hong Kong University of Science and Technology, Hong Kong, China\\
\textsuperscript{\rm 7}Tsinghua University, Beijing, China\\
\textsuperscript{\rm 8}Singapore University of Technology and Design, Singapore\\
\{zhijunchen, zeyuji, yikunb, sunhl\}@buaa.edu.cn,
xiao.huang@polyu.edu.hk\\[0.5em]
{\normalsize
\pdflink{https://zeyuji.github.io/LLM-PeerReview/}{\textcolor{blue!60!black}{\faGlobe\enspace Website}}
\qquad
\pdflink{https://github.com/zeyuji/LLM-PeerReview/}{\textcolor{blue!60!black}{\faGithub\enspace Code}}
}
}

\begin{document}

\maketitle
{\renewcommand{\thefootnote}{\fnsymbol{footnote}}%
\footnotetext[2]{Equal contribution.}%
\footnotetext[3]{Core experimental contributors.}%
\footnotetext[1]{Corresponding author.}%
}

\begin{abstract}
We propose LLM-PeerReview, an unsupervised LLM Ensemble method that selects the most ideal response from multiple LLM-generated candidates for each query, harnessing the collective wisdom of multiple models with diverse strengths.
LLM-PeerReview is built on a novel, peer-review-inspired framework that offers a transparent and interpretable mechanism, while remaining fully unsupervised for flexible adaptability and generalization.
Specifically, it operates in three stages:
For \textit{scoring}, we use the emerging LLM-as-a-Judge technique to evaluate each response by reusing multiple LLMs at hand;
For \textit{reasoning}, we can apply a straightforward averaging strategy or a principled graphical model-based truth inference algorithm to aggregate multiple scores to produce a final score for each response;
Finally, the highest-scoring response is \textit{selected} as the best ensemble output.
LLM-PeerReview is conceptually simple and empirically powerful.
Our results across four datasets show that the two variants of the proposed approach outperform the advanced model Smoothie-Global by 6.9\% and 7.3\% points, cross diverse task types including factual recall QA, math reasoning, and instruction following.

\textbf{Notably,} we also establish a carefully curated benchmark suite for LLM Ensemble, integrating 12 methods across four classic datasets and three task families, all evaluated under a rigorous and consistent protocol.
We hope this repository will help researchers reproduce the LLM Ensemble baselines.

\end{abstract}


\section{Introduction}
\label{sec: Introduction}

The artificial intelligence domain has undergone a massive transformation recently, driven by the emergence of Large Language Models (LLMs) such as Gemini~\cite{team2023gemini}, GPT-4~\cite{achiam2023gpt}, Llama~\cite{touvron2023llama}, and DeepSeek~\cite{liu2024deepseek}.
The success of these models has triggered a surge in research activity, with over 182,000 models now available on Hugging Face.

Behind this research enthusiasm, we can observe two main points:
% ~\cite{jiang2023llm}:
1) \textit{Persistent performance concerns}:
Although LLMs can be easily deployed for zero-shot or in-context few-shot inference, they still face common performance issues, such as limited accuracy, hallucinations, and misalignment with human goals;
2) \textit{The varying strengths and weaknesses of LLMs}: 
These models display significant behavioral differences, primarily driven by variations in their architecture, scale, training data, dictionary, tokenization and methodology. 
Thus, their responses to the same prompt often diverge.
With the above two points in mind and inspired by the spirit of Ensemble Learning~\cite{dong2020survey}, it is reasonable to suggest that it is advantageous to simultaneously consider multiple LLMs  (which are usable out-of-the-box) and leverage their distinct strengths.
This concept is the core focus of the burgeoning field of LLM Ensemble~\cite{chen2025harnessing}.

\begin{figure*}[tb]
\centering
\includegraphics[width=0.6567\linewidth]{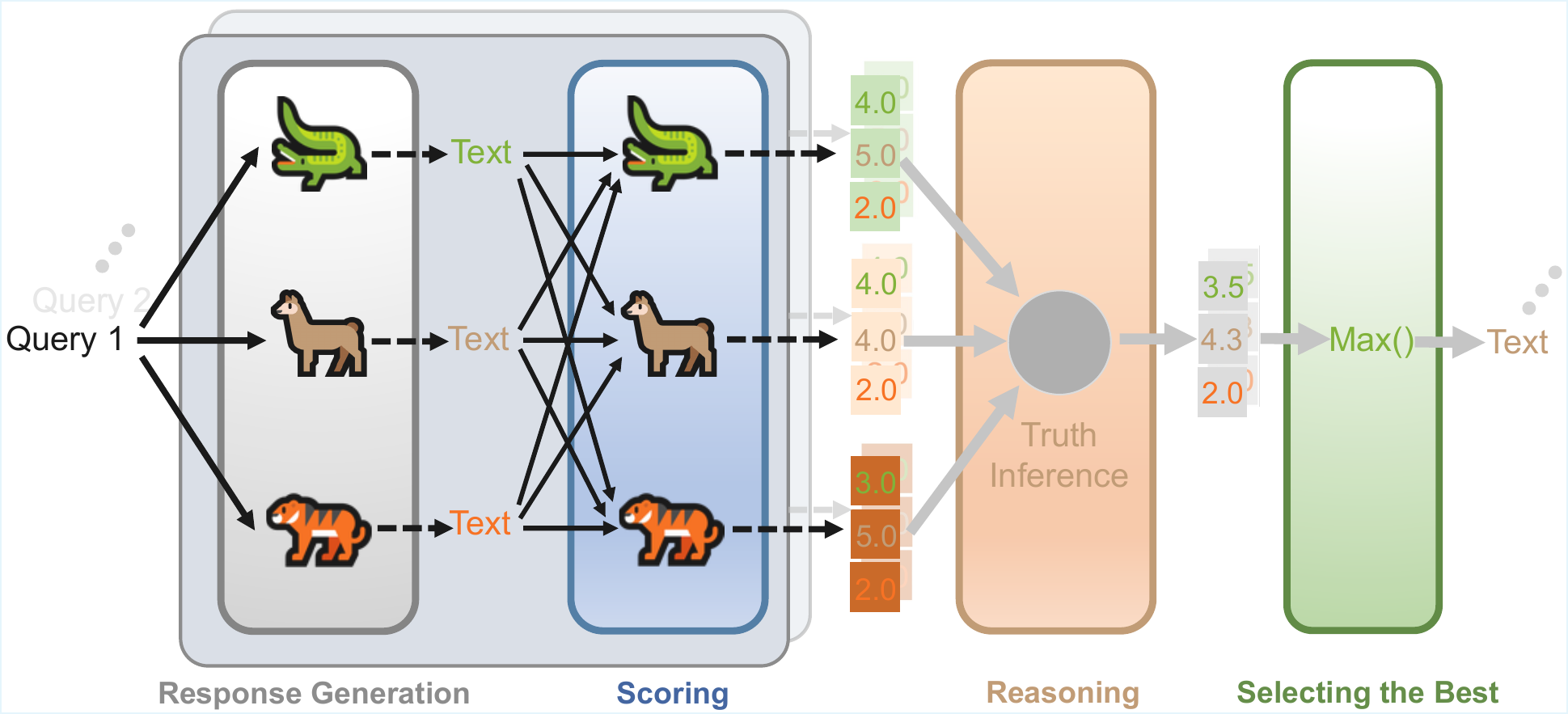}
\caption{The proposed LLM-PeerReview  contains three steps:
\textbf{(1) Scoring}: 
For a given query, after each LLM independently generates a response (analogous to a submitted academic paper), LLM-PeerReview applies the LLM-as-a-Judge technique (and the proposed flipped-triple scoring trick), treating each model as a reviewer to assign scores to all candidate responses;
\textbf{(2) Reasoning}: 
LLM-PeerReview then uses a truth inference algorithm—analogous to a senior reviewer—to estimate a final score for each response;
% (Notably, for the variant LLM-PeerReview-W, the inference algorithm is performed using score information across all queries, allowing the model to learn each LLM’s scoring behavior using global information from the dataset);
% thereby enabling fine-grained, reliability-aware score aggregation);
\textbf{(3) Selecting the best:}
Finally, for each query, LLM-PeerReview selects the response with the highest final score as the ensemble output—analogous to how a senior reviewer chooses the most ideal paper from a specific submission pool.
}
\label{fig: method overview}
\end{figure*}

As LLM Ensemble gains increasing attention, one well-established class of solutions—\textit{ensemble-after-inference} (also known as \textit{post-hoc ensemble}) methods—has emerged. 
These methods include the following two categories:
% \vspace{-5pt}
\begin{itemize}
\item 
\textbf{Selection-then-regeneration approach}~\cite{jiang2023llm,lv2024urg,tekin2024llm}, first employs a pre-trained ``PairRanker'' module to select the top-K candidate responses—those deemed most likely to be of high quality—from a pool of LLM-generated responses. 
This selected subset is then fed into another fine-tuned LLM (e.g., Flan-T5-XL~\cite{chung2024scaling}) to synthesize a \textit{final} response.
While these methods have attracted significant attention~\cite{jiang2023llm}, they rely on carefully curated task-specific training data and the need to fine-tune an additional LLM, \textit{limiting their generalization and adaptability}.
\item 
\textbf{Similarity-based selection approach}~\cite{li2024more,guha2024smoothie,si2023getting}, instead, is mostly fully unsupervised~\cite{guha2024smoothie,si2023getting}.
These methods follow a simple principle: \textit{for a given query, select the response with the highest total similarity to all other responses}.
While such methods pioneered unsupervised post-hoc LLM Ensemble, their design remains coarse-grained---they rely on the naive similarity-based selection strategy~\cite{li2024more,guha2024smoothie}, along with shallow similarity measure of BLEU~\cite{li2024more} and limited informational utilization~\cite{guha2024smoothie}.
So, the true potential of the ensemble remains largely untapped.
\end{itemize}

% \vspace{-6pt}
When we revisit this research problem, we ask the most fundamental question: \textit{In the real world, how would humans select the most ideal text from a set of candidate texts?}  
Perhaps the most immediate and relatable real-world example is: the academic peer-review process.
Motivated by this, we propose a new, fully unsupervised, training-free LLM Ensemble approach called LLM-PeerReview.

Specifically, LLM-PeerReview comprises three sequential modules:
(1) \textit{Scoring (analogous to paper reviewing)}:
Given multiple candidate responses to the same query, we adopt the LLM-as-a-Judge approach—leveraging available LLMs as evaluators that assess each response and assign a score (e.g., 5.0 indicating Strong Accept, 4.0 indicating Weak Accept, etc.).
To enhance scoring precision and mitigate inherent biases, we further propose the \textit{flipped-triple scoring trick}—a cornerstone mechanism that fundamentally determines the efficacy  of the entire framework;
(2) \textit{Reasoning (analogous to final score estimation made by the senior reviewer)}:
Besides being able to perform direct averaging calculations, we can invoke graphical model-based truth inference techniques from weak supervision literature, to perform reliability-aware weighted score aggregation, deriving a final score for each response; 
(3) \textit{Selection (analogous to final decision made by the senior reviewer)}:
This step is analogous to how a senior reviewer or area chair selects the most suitable paper from a specific submission pool.
For each query, we identify the highest-scoring response as the final ensemble result. 
% once all scores have been inferred.

% \textbf{Contribution.}
% LLM-PeerReview is built upon a transparent and interpretable framework that mimics the peer-review process, where its fully unsupervised nature provides flexibility and generalizability across various tasks and datasets.
% Within the methodology, we employ the LLM-as-a-Judge approach to leverage collective intelligence of LLMs for refined evaluation, moving beyond coarse metrics like BLEU, and propose a novel pivotal flipped-triple scoring trick to reduce scoring bias.
% Further, we provide both straightforward averaging and graphical model-based weighted methods to aggregate multiple scoring signals.
% Extensive experiments show that LLM-PeerReview substantially outperforms recent advanced similarity-based methods~\cite{li2024more,guha2024smoothie}.
% % and all  individual LLMs.

\textbf{Contribution.}
(1) 
We propose the first peer-review inspired framework that systematically maps the academic peer-review process to LLM Ensemble, which offers a new, unsupervised, interpretable paradigm beyond existing methods.
Further, we develop two variants, including a principled graphical-model-based method for a more refined instantiation of the framework;
(2) We introduce a novel \textit{flipped-triple scoring} technique to solve the coupling of position and scoring biases in LLMs. 
 Not only is it simple, but this technique is also decisive for the efficacy of the proposed framework;
% To our knowledge, no equivalent debiasing design has previously been studied in the LLM-as-a-Judge field. 
 % The "flipped-triple" scoring trick is a novel procedural control specifically designed to solve the coupling of position and scoring biases in LLMs. Not only is it simple, but this technique is also decisive for the efficacy of the proposed framework. 
 % To our knowledge, no equivalent technique currently exists in the LLM-as-a-Judge field. 
(3) Extensive experiments show that our methods outperform the strong Smoothie-Global~\cite{guha2024smoothie} by 6.9 and 7.3 points, and surpass the recent sophisticated token-level methods SAFE~\cite{yun2026when} and CoRE~\cite{zeng2026harnessing}, while outperforming the classical debate-based method MAD~\cite{liang2024encouraging} in both performance and efficiency; 
% (4) As a theoretical contribution (provided in the Appendix), we establish formal error--ambiguity decomposition theorems for the scoring stage, showing how judge quality and diversity jointly affect ensemble error.
% These theoretical results are  due to space constraints.
(4) To support reproducible research, we release a unified open-source benchmark suite implementing both LLM-PeerReview variants and 10 LLM Ensemble baselines across four datasets.

% The main contribution of LLM-PeerReview is a new unsupervised formulation of response selection as peer-review-based truth inference. Its novelty lies in how evaluation evidence is constructed and inferred, rather than in any component alone. Flipped-triple scoring converts biased point-wise judgments into structured comparative evidence: local triplets provide comparison context, while reversed orders reduce position bias. The weighted variant, LLM-PeerReview-W, then jointly infers latent response quality and judge-specific scoring behavior across queries, using soft interpolation to retain decimal-valued scores. Together, these steps turn imperfect and heterogeneous judgments into a reliability-aware decision without labels, reference answers, or additional model training. We also establish an error--ambiguity decomposition that explains how judge error and diversity affect the ensemble estimate. Experiments across four datasets and three task types show consistent gains over strong post-hoc baselines, with further favorable comparisons to recent token-level and debate-based methods.

\section{LLM-PeerReview}
\label{sec: LLM-PeerReview}
% \label{sec:llm-peerreview}

This section presents our proposed method, LLM-PeerReview, with an overview shown in Figure~\ref{fig: method overview}.
We begin by formalizing the  problem, followed by a detailed introduction of LLM-PeerReview's three components  in Sections~\ref{sec:scoring},~\ref{sec: Reasoning: a Truth Inference Process}, and~\ref{sec: Selecting the Best}.

% \label{sec:scoring}

\paragraph{Problem Formulation: (Unsupervised) LLM Ensemble.} 
Without access to any reference responses (i.e., ideal/truth responses), we are given a set of queries $\{\mathbf{x}^{(i)}\}_{i=1}^I$. 
% for a generative task.
We have access to $J$ large language models $\{\mathcal{M}_j\}_{j=1}^J$, where each model $\mathcal{M}_j$ generates a response  $\mathbf{r}^{(i,j)} = \mathcal{M}_{j}(\mathbf{x}^{(i)})$---which is often not ideal---for a given query $\mathbf{x}^{(i)}$.
Thus, for each query, we have a set of zero-shot inference responses $\mathbf{R}^{(i)}=[\mathbf{r}^{(i,1)}, \ldots, \mathbf{r}^{(i,J)}]$ from heterogeneous LLMs $\{\mathcal{M}_j\}_{j=1}^J$, while the underlying reference response $\mathbf{y}^{(i)}$ is \textit{unobserved} to us.
Our goal is to ensemble the LLM responses to produce a single, high-quality final response for each query $\mathbf{x}^{(i)}$, using the available data $\mathcal{D}=\{\mathbf{x}^{(i)}, \mathbf{R}^{(i)}\}_{i=1}^I$.

\subsection{Scoring}
% \label{sec: Scoring}
\label{sec:scoring}

As shown in Figure~\ref{fig: method overview} and within our proposed LLM-PeerReview, the scoring phase occurs after the LLMs have first generated responses to the input queries.
Using the LLM-as-a-Judge technique~\cite{gu2024survey}, each LLM judge can assign a \textit{point-wise score} to each response.
% , representing its overall quality. 
~\footnote{For example, the score can range from [1, 2, 3, 4, 5], representing the levels of [``\texttt{Very Poor}'', ``\texttt{Poor}'', ``\texttt{Acceptable}'', ``\texttt{Good}'', ``\texttt{Excellent}''].}

\paragraph{Flipped-triple scoring trick.}
The above  \textit{naive point-wise scoring} technique can provide scores; however, we propose a new technique called \textit{flipped triple scoring}, which we recommend when applying our approach.~\footnote{It is worth noting that each scoring prompt contains both the corresponding query and the response to be evaluated, rather than presenting the response alone. 
The scoring prompts that we designed and utilized in the experiments are provided in the Appendix.}
Specifically: (1) For multiple responses from different models to the same query $\mathbf{x}^{(i)}$, we first shuffle them; (2) Then, for each LLM judge $\mathcal{M}_{j^{\prime}}$, we score the response triplet $[\mathbf{r}^{(i,j-1)},\mathbf{r}^{(i,j)},\mathbf{r}^{(i,j+1)}]$ sequentially (with $J$ times), and for each iteration, we also score the flipped version (i.e., $[\mathbf{r}^{(i,j+1)},\mathbf{r}^{(i,j)},\mathbf{r}^{(i,j-1)}]$). 
As a result, each response receives six scores from the same LLM judge. 
We can simply average these scores to obtain a final score, which serves as the score $y^{(i,j;j^{\prime})}$ for response $\mathbf{r}^{(i,j)}$ by LLM judge $\mathcal{M}_{j^{\prime}}$.
In short, this technique mitigates two common scoring biases in LLM-as-a-Judge~\cite{wang2023large,zheng2023judging,gu2024survey}.
First, in point-wise scoring, models tend to show a consistent bias toward certain scores (e.g., consistently assigning a score of ``1''), as they evaluate a single response without the reference effect of multiple responses.
Also, when multiple responses (such as two or three) are presented for evaluation at one time, models frequently exhibit a \textit{position bias}, tending to favor responses that appear either at the beginning or at the end.

\paragraph{More  efficient scoring.}
The efficiency of the scoring stage can be readily improved in several ways.
For example, as discussed in Section~\ref{sec: others}, LLM-PeerReview can use a selected set of judges during scoring,  reducing computation while retaining strong performance.
Also, flipped-triple scoring can be implemented with a sliding window over candidate responses, avoiding exhaustive enumeration of response groups. 

% \paragraph{More efficient scoring.}
% The scoring stage can be made more efficient in two simple ways. First, flipped-triple scoring can use a sliding window over candidate responses, avoiding exhaustive enumeration of response groups. Second, as discussed in Section~\ref{sec:others}, LLM-PeerReview can use fewer judges during scoring, further reducing computation while retaining strong performance.

\subsection{Reasoning: a Truth Inference Process}
\label{sec: Reasoning: a Truth Inference Process}

% For the Reasoning process within LLM-PeerReview, as noted in Section ~\ref{sec: Introduction} 
and the caption of Figure~\ref{fig: method overview}, we can adopt different specific sub-components: (1) A default, straightforward, yet effective \textit{averaging method}, as described in Section 2.2.1; (2) A \textit{weighted method}, LLM-PeerReview-W, as detailed in Section 2.2.2.

\subsubsection{2.2.1 Reasoning for Default LLM-PeerReview}
\label{sec: Reasoning for Default LLM-PeerReview}

After the scoring phase, as shown in Figure~\ref{fig: method overview}, each response $\mathbf{r}^{(i,j)}$ corresponding to a query $\mathbf{x}^{(i)}$ receives multiple scores $\{y^{(i,j;j^{\prime})}\}_{j^{\prime}=1}^J$ provided by multiple LLM judges.
Then, how can we aggregate these scores meaningfully to compute a final, reliable score $Score(\mathbf{r}^{(i,j)})$ for each response $\mathbf{r}^{(i,j)}$, using scores $\{y^{(i,j;j^{\prime})}\}_{j^{\prime}=1}^J$? 
This problem is analogous to designing an algorithm that simulates a senior reviewer who consolidates evaluations from multiple reviewers with different scoring preferences and evaluation capabilities.
A straightforward and intuitive approach is \textit{averaging}~\cite{zheng2017truth,zhang2021wrench}—simply taking the mean of all the scores for a given response:
% \begin{small}
\begin{equation}
Score(\mathbf{r}^{(i,j)}) =\frac{1}{J}\sum_{j^{\prime}=1}^{J} y^{(i,j;j^{\prime})}.
\label{eq: obtain averaging scores}
\end{equation}
% \end{small}

\subsubsection{2.2.2 Reasoning for LLM-PeerReview-W}
\label{sec: Reasoning for LLM-PeerReview-W}

% \paragraph{Another variant: LLM-PeerReview-W.}
% We observe that the above averaging strategy assumes all models to be equally reliable.
% We further propose a weighted variant, referred to as \textit{LLM-PeerReview-W}.
% We invoke the well-established Dawid-Skene (DS) model~\cite{dawid1979maximum}, a canonical truth-inference graphical model widely used in weak supervision learning ~\cite{zhang2021wrench}.
% and adapt it to our context. 
Here,
we implement several necessary improvements to the  well-established DS model~\cite{dawid1979maximum}.
% particularly regarding the processing of regression scores.
In the following, we introduce the construction of the graphical model, and present the optimization objective and optimization. 
% (to obtain the final score  $Score(\mathbf{r}^{(i,j)})$ for each response $\mathbf{r}^{(i,j)}$ for each response $\mathbf{r}^{(i,j)}$).
\textit{In brief and intuitively, we perform two iterative steps: (1) inferring the true score category for each response, and (2) learning the parameters of the graphical model.}

\begin{figure}[t] 
    \centering
    \includegraphics[width=0.16\textwidth]{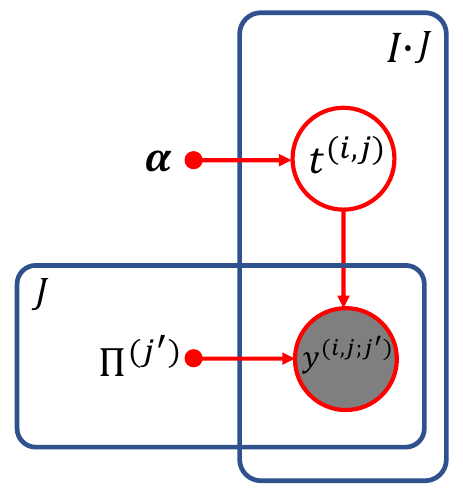} 
    \caption{Probabilistic graphical representation.}
    \label{fig: graphical model}
    % \vspace{-1em}
\end{figure}

% \paragraph{\hypertarget{target:graph_model}{Graphical model construction.}} 
\paragraph{Graphical model construction.}
Overall, to infer the underlying ``truth'' score (\textit{unobserved}) 
% behind the multiple weak/non-ideal score annotations 
% (\textit{observed}) 
for each response, we construct a \textit{latent variable graphical model}~\cite{bishop2006pattern}.
% that includes a \textit{latent variable} representing the truth score.
As depicted the graphical representation in Figure~\ref{fig: graphical model}, 
% we next introduce the probabilistic generative process we construct from truth scores to weak scores labeled by LLM judges.
% First, 
for each response $\mathbf{r}^{(i,j)}$, we assume that its true score $t^{(i,j)}$ is drawn from a categorical distribution:
\begin{equation}
t^{(i,j)} \sim \operatorname{Cat}(t^{(i,j)} ; \boldsymbol{\alpha} )\mathrm{,}
\label{eq: distribution of t}
\end{equation}
where the distribution is parametrized by $\boldsymbol{\alpha}$.
Next, 
% similar to the concept of \textit{confusion matrix} 
% % commonly used in machine 
% % % learning~\cite{bishop2006pattern,goodfellow2016deep},
% % learning
% ~\cite{bishop2006pattern},
we introduce an annotator-specific \textit{transition matrix} $\boldsymbol{\Pi}^{(j^{\prime})}$ to model the probability that an LLM confuses one score category for another, capturing its scoring tendencies:
% and potential biases:
\begin{equation}
p(y^{(i, j;j^{\prime})}=n | t^{(i,j)}=m ; \boldsymbol{\Pi}^{(j^{\prime})})=\pi_{m n}^{(j^{\prime})}\mathrm{,}
\label{eq: distribution of y}
\end{equation}
where $m, n \in\{1, \ldots, K\}$ and $K$ denotes the number of categories (i.e., the number of score levels).

\paragraph{Objective and optimization.}
% Based on the model construction above, 
The optimization objective is to maximize the log conditional likelihood of the observed scoring labels $ \mathbf{Y} = \{y^{(i,j;j^{\prime})} \mid 1 \le i \le I,\ 1 \le j \le J ,\ 1 \le j^{\prime} \le J \}$,
% contributed by $J$ LLM judges,
i.e., $\log p(\mathbf{Y} ; \Theta)$  w.r.t. the parameters $\Theta=\{\boldsymbol{\alpha}, \boldsymbol{\Pi}^{(1)}, \ldots, \boldsymbol{\Pi}^{(J)}\}$. 
% \paragraph{Optimization.}
% For optimization, 
% in brief, 
% As with most latent variable models,
% ~\cite{bishop2006pattern,dawid1979maximum},
We apply the Expectation-Maximization (EM) algorithm~\cite{dempter1977maximum} to solve the optimization problem.\footnote{We provide the detailed derivations in Appendix.}
First, the log-likelihood can be written as:
% \begin{small}
\begin{equation}
\log p(\mathbf{Y} ; \Theta)=\sum_{i=1}^I \sum_{j=1}^J \log p(\mathbf{y}^{(i, j)} ; \Theta)
\label{eq: likelihood}
\end{equation}
% \end{small}
where $ \mathbf{y}^{(i,j)} = \{y^{(i,j;j^{\prime})}\mid1 \le j^{\prime} \le J \}$ denotes the set of scores assigned to response $\mathbf{r}^{(i,j)}$.
Then, we use Jensen’s inequality to get the \textit{Evidence Lower Bound} (ELBO):
\begin{small}
\begin{equation}
\log p(\mathbf{Y} ; \Theta)\geq  \sum_{i=1}^{I}  \sum_{j=1}^{J}  \sum_{t^{(i,j)}} q(t^{(i,j)}) \log \frac{p(\mathbf{y}^{(i,j)}, t^{(i,j)}  ; \Theta)}{q(t^{(i,j)})},
\label{eq: elbo}
\end{equation}
\end{small}
where $q(t^{(i,j)})$ is a discrete distribution over the variable $t^{(i,j)}$.
In the following, we then proceed to apply the general EM recipe to perform iterative calculations.

% (concerning E-step and M-step) to solve the optimization problem  $\Theta := \underset{\Theta}{\operatorname{argmax}} \log p(\mathbf{Y} ; \Theta)$.

\paragraph{Optimization: E-step (inference).}
The posterior $q(t^{(i,j)})$ is obtained by using of Bayes's theorem given the parameters $\Theta=\{\boldsymbol{\alpha}, \boldsymbol{\Pi}^{(1)}, \ldots, \boldsymbol{\Pi}^{(J)}\}$ learned on the last M-step:
\begin{small}
\begin{equation}
\begin{aligned}
&q(t^{(i,j)}=k) :=p(t^{(i,j)}=k \mid \mathbf{y}^{(i,j)}; \Theta) \\
\ \qquad & \propto p(t^{(i,j)}=k; \boldsymbol{\alpha} ) \cdot  \prod_{j^{\prime}=1}^{J}  p(y^{(i, j;j^{\prime})} \mid t^{(i,j)}=k ; \boldsymbol{\Pi}^{(j^{\prime})}).
\end{aligned}
\label{eq: posterior 1}
\end{equation}
\end{small}
Given that we are likely to obtain decimal values rather than integers on $y^{(i, j;j^{\prime})}$ after the scoring phase, we make the adaptation:
\begin{small}
\begin{equation}
\begin{aligned}
q(t^{(i,j)}=k)
\propto  & p(t^{(i,j)}=k; \boldsymbol{\alpha} ) \cdot \prod_{j^{\prime}=1}^{J}  [\phi_{l}\cdot
 p(y^{(i, j;j^{\prime})}_{l} \mid t^{(i,j)}=k ; \\\boldsymbol{\Pi}^{(j^{\prime})})
 &  + \phi_{u}\cdot p(y^{(i, j;j^{\prime})}_{u} \mid t^{(i,j)}=k ; \boldsymbol{\Pi}^{(j^{\prime})})],
\end{aligned}
\label{eq: posterior 2}
\end{equation}
\end{small}
where $\phi_{l}$ and $\phi_{u}$ represent the confidences for the decimal $y^{(i, j;j^{\prime})}$ corresponding to its lower and upper nearest integer neighbors (i.e., $y^{(i, j;j^{\prime})}_{l}$, $y^{(i, j;j^{\prime})}_{u}$).

\paragraph{Optimization: M-step (learning).}
% Furthermore, by maximizing optimization objective in Equation~\ref{eq: elbo},
% and using the standard Lagrange multiplier method~\cite{bishop2006pattern}, 
Furthermore, we can obtain the closed-form solution
solutions for $\boldsymbol{\alpha}=\{\alpha^{(k)}\mid1 \le k \le K \}$ and $\{\boldsymbol{\Pi}^{(j^{\prime})}\}_{j^{\prime}=1}^J$ by the EM algorithm:
% for $\boldsymbol{\alpha}=\{\alpha^{(k)}\mid1 \le k \le K \}$ in Equation~\ref{eq: alpha} shown below; and by equating the gradient of Equation~\ref{eq: elbo}  to zero, we can obtain the closed-form solution for $\{\boldsymbol{\Pi}^{(j^{\prime})}\}_{j^{\prime}=1}^J$ in Equation~\ref{eq: pai}. 
% shown below.
% \begin{small}
\begin{equation}
\alpha_k=\frac{\sum_{i=1}^I \sum_{j=1}^J q(t^{(i,j)}=k)}{I\cdot J},
\label{eq: alpha}
\end{equation}
% \end{small}
\begin{small}
\begin{equation}
\pi_{m n}^{(j^{\prime})}=\frac{\sum_{i=1}^{I}\sum_{j=1}^{J}  q(t^{(i,j)}=m) \cdot\Psi(y^{(i,j;j^{\prime})},n) }{\sum_{i=1}^{I}\sum_{j=1}^{J} q(t^{(i,j)}=m) }\mathrm{,}
\label{eq: pai}
\end{equation}
\end{small}
where $\Psi(y^{(i,j;j^{\prime})})=[\phi_{l}\cdot \mathbb{I}(y^{(i,j;j^{\prime})}_l=n) + \phi_{u}\cdot \mathbb{I}(y^{(i,j;j^{\prime})}_u=n]$, and $\mathbb{I}(\cdot)$ is an indicator function that takes the value $1$ when the internal declaration is true, and $0$ otherwise.

\paragraph{Obtain the final score value for each response.}
\label{Obtain the final score value for each response}
After the EM-based optimization, we obtain the posterior probabilities $q(t^{(i,j)})$ over score categories.
% for each response.
Then, we compute the final score for each response:
% through the following simple summation:
% \begin{small}
\begin{equation}
Score(\mathbf{r}^{(i,j)})=\sum_{k=1}^K q(t^{(i,j)}=k) \cdot s_k,
\label{eq: obtain the final score value for each response}
\end{equation}
% \end{small}
where $s_k$ denotes the score value corresponding to the $k$-th scoring category.\footnote{For example, suppose that for a given response, we have $q(t^{(i,j)}=4)=0.5$, $q(t^{(i,j)}=5)=0.5$, along with the score values $s_4=4.0$, $s_5=5.0$.
Then, the final aggregated score is $Score^{(i,j)}=0.5 \cdot 4.0+0.5 \cdot 5.0=4.5$.}

\subsection{Selecting the Best}
\label{sec: Selecting the Best}

Finally, for each query $\mathbf{x}^{(i, j)}$, we can easily determine its optimal response: 
\begin{small}
\begin{equation}
\mathbf{r}_{\text {ensemble}}^{(i)}=
  \underset{\mathbf{r}^{(i, j)}}{\operatorname{argmax}}
\{Score(\mathbf{r}^{(i,j)}) \mid 1 \le j \le J \},
\label{eq: obtain the final results}
\end{equation}
\end{small}
which is selected as the final result after the ensemble.

% In summary, the overall procedure for our proposed LLM-PeerReview is provided in Algorithm~\ref{alg:algorithm}.

In summary, the overall procedure for our proposed LLM-PeerReview is provided as Algorithm 1 in the Appendix.

\subsection{Others}
\label{sec: others}

% \subsection{Others: (a) Using Less Judges for Efficiency}
% \label{sec: Others: (a) Using Less Judges for Efficiency}

\paragraph{(a) Using less judges for efficiency.}
Upon closer examination of our proposed LLM-PeerReview framework , it is evident that the scoring phase can selectively employ a subset of models rather than relying on the entire ensemble. When utilizing a model subset, the underlying logic of the reasoning process remains unchanged, allowing existing methods to function as intended. This flexibility facilitates an efficiency-performance trade-off during the deployment phase. 
Our experiments further demonstrate that, supported by the framework's effectiveness and the de-biasing capability of the flipped-triple scoring trick, LLM-PeerReview achieves  satisfactory performance even when scoring is conducted by a subset of models.

% \subsection{Others: (b) Theoretical Analysis on Judging Process}
% \label{sec: Others: (b) Theoretical Analysis on Judging Process}

\paragraph{(b) Theoretical analysis on judging process.}
Furthermore, in terms of the scoring process, we can choose combinations that exhibit greater diversity. We provide a corresponding proof in the Appendix: as the general performance of the judges improves and their diversity increases, the error of the final ensemble score decreases.

\begin{figure*}[!h]
\centering
\hspace{-1.8cm}
\begin{minipage}[t]{0.69\textwidth}
\begin{center}
\resizebox{0.89\linewidth}{!}{
\begin{tabular}{lccccc}
\toprule
\textbf{Method} &
\textbf{TriviaQA}$\uparrow$ & \textbf{GSM8k}$\uparrow$ & \textbf{MATH}$\uparrow$ & \textbf{AlpacaEval}$\uparrow$ &
\textbf{Average}$\uparrow$ \\
\midrule
\multicolumn{6}{l}{\textit{Single LLM}} \\
 Llama-3.1-8B-Instruct & 75.3 & 79.3 & 52.3 & 7.3 & 53.5 \\
Mistral-7B-Instruct & 72.7 & 64.3 & 26.5 & 10.4 & 43.5 \\
Qwen2-7B-Instruct & 63.0 & 88.5 & 59.8 & 15.2 & 56.6 \\
Qwen2.5-7B-Instruct & 62.5 & 91.5 & 69.3 & 27.6 & 62.7 \\
% Theoretical average & 68.4 & 80.9 & 51.9 & 15.1 & 54.1 \\
\midrule

\multicolumn{6}{l}{\textit{LLM Ensemble}} \\
Random \scriptsize{\cite{guha2024smoothie}}
& 68.4 \scriptsize{$\pm$ 0.3} & 81.2 \scriptsize{$\pm$ 1.2} & 52.2 \scriptsize{$\pm$ 1.1} & 15.2 \scriptsize{$\pm$ 0.6} & 54.2 \\
Smoothie-Global \scriptsize{\cite{guha2024smoothie}}
& 63.0 & 91.5 & 59.8 & 27.6 & 60.5 \\
Smoothie-Local \scriptsize{\cite{guha2024smoothie}}
& 73.6 & 85.5 & 61.8 & 18.3 & 59.8 \\
Agent-Forest \scriptsize{\cite{li2024more}}
& 70.5 & 86.8 & 61.0 & 22.1 & 60.1 \\
GaC \scriptsize{\cite{yu2024breaking}}
& 71.5 & 91.8 & 54.0 & 23.6 & 60.2 \\

% SAFE-GaC \scriptsize{\cite{yun2026when}}
% & 63.2 & 90.5 & 41.3 & - & - \\
% SAFE-UniTE \scriptsize{\cite{yun2026when}}
% & 63.3 & 91.5 & 48.5 & - & - \\
% CoRE-GaC \scriptsize{\cite{zeng2026harnessing}}
% & 64.0 & 92.0 & 63.5 & - & - \\
% CoRE-UniTE \scriptsize{\cite{zeng2026harnessing}}
% & 64.0 & 92.0 & 63.3 & - & - \\

SAFE-GaC \scriptsize{\cite{yun2026when}}
& 63.2 & 90.5 & 41.3 & 20.1 &  53.8\\
SAFE-UniTE \scriptsize{\cite{yun2026when}}
& 63.3 & 91.5 & 48.5 & 16.5 & 55.0 \\
CoRE-GaC \scriptsize{\cite{zeng2026harnessing}}
& 64.0 & 92.0 & 63.5 & 8.7 & 57.1 \\
CoRE-UniTE \scriptsize{\cite{zeng2026harnessing}}
& 64.0 & 92.0 & 63.3 & 8.7 & 57.0 \\

MAD \scriptsize{\cite{liang2024encouraging}}
& -& 91.8
 & 47.3 & - & - \\
\rowcolor[rgb]{0.867, 0.922, 0.969}
LLM-PeerReview & 76.9 \scriptsize{$\pm$ 0.1} & 92.7 \scriptsize{$\pm$ 0.3} & 69.5 \scriptsize{$\pm$ 0.2} & \textbf{30.4} \scriptsize{$\pm$ 0.1} & 67.4 \\
\rowcolor[rgb]{0.867, 0.922, 0.969}
LLM-PeerReview-W & \textbf{77.0} \scriptsize{$\pm$ 0.1} & \textbf{93.0} \scriptsize{$\pm$ 0.2} & \textbf{71.0} \scriptsize{$\pm$ 0.2} & 30.2 \scriptsize{$\pm$ 0.1} & \textbf{67.8} \\
\midrule

\multicolumn{6}{l}{\textit{LLM Ensemble (our variants)}} \\
Llama-3.1-8B-Selection & 76.5 \scriptsize{$\pm$ 0.2} & 90.8 \scriptsize{$\pm$ 0.6} & 68.8 \scriptsize{$\pm$ 0.5} & 29.6 \scriptsize{$\pm$ 0.3} & 66.4 \\
Mistral-7B-Selection & 75.6 \scriptsize{$\pm$ 0.3} & 90.8 \scriptsize{$\pm$ 0.1} & 66.4 \scriptsize{$\pm$ 0.3} & 25.9 \scriptsize{$\pm$ 0.4} & 64.7 \\
Qwen2-7B-Selection & 74.2 \scriptsize{$\pm$ 0.2} & 88.8 \scriptsize{$\pm$ 0.6} & 61.7 \scriptsize{$\pm$ 0.7} & 23.7 \scriptsize{$\pm$ 0.3} & 62.1 \\
Qwen2.5-7B-Selection & 75.5 \scriptsize{$\pm$ 0.2} & 92.1 \scriptsize{$\pm$ 0.4} & 66.2 \scriptsize{$\pm$ 0.6} & 28.1 \scriptsize{$\pm$ 0.1} & 65.5 \\
\bottomrule
\end{tabular}
}
\end{center}
% \begin{tablenotes}
%  % \footnotesize   
% \small
% % \normalsize
% % \large
% \item[1] $\dag$: ``-'' denotes that the method is not applicable to the corresponding task.
% \end{tablenotes}
\captionof{table}{Main results (\%). “-”: not applicable to the corresponding task.}
\label{table: main results}
\end{minipage}
\hspace{0.01\textwidth} 
\begin{minipage}[t]{0.24\textwidth}
\vspace{-3.2cm}
    \centering
    % \hfill
    \begin{minipage}[t]{\textwidth}
        % \centering
        \hspace{0.1\textwidth} 
        \includegraphics[width=0.83\textwidth]{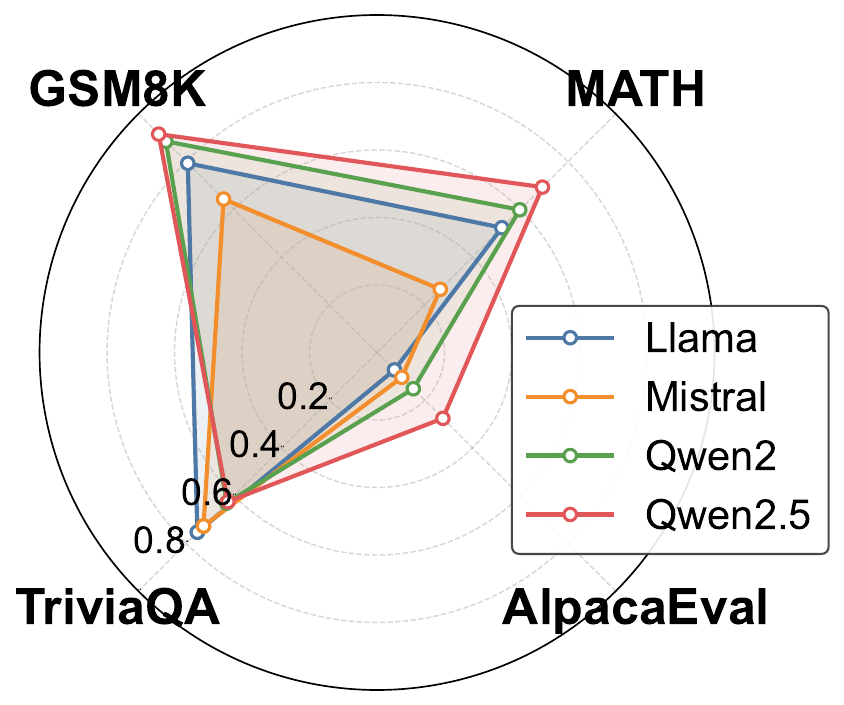} 
    \end{minipage}
    \vspace{0.5cm}  
    \\[-0.3cm]  
    \begin{minipage}[b]{\textwidth}
        \centering
        \includegraphics[width=1.\textwidth]{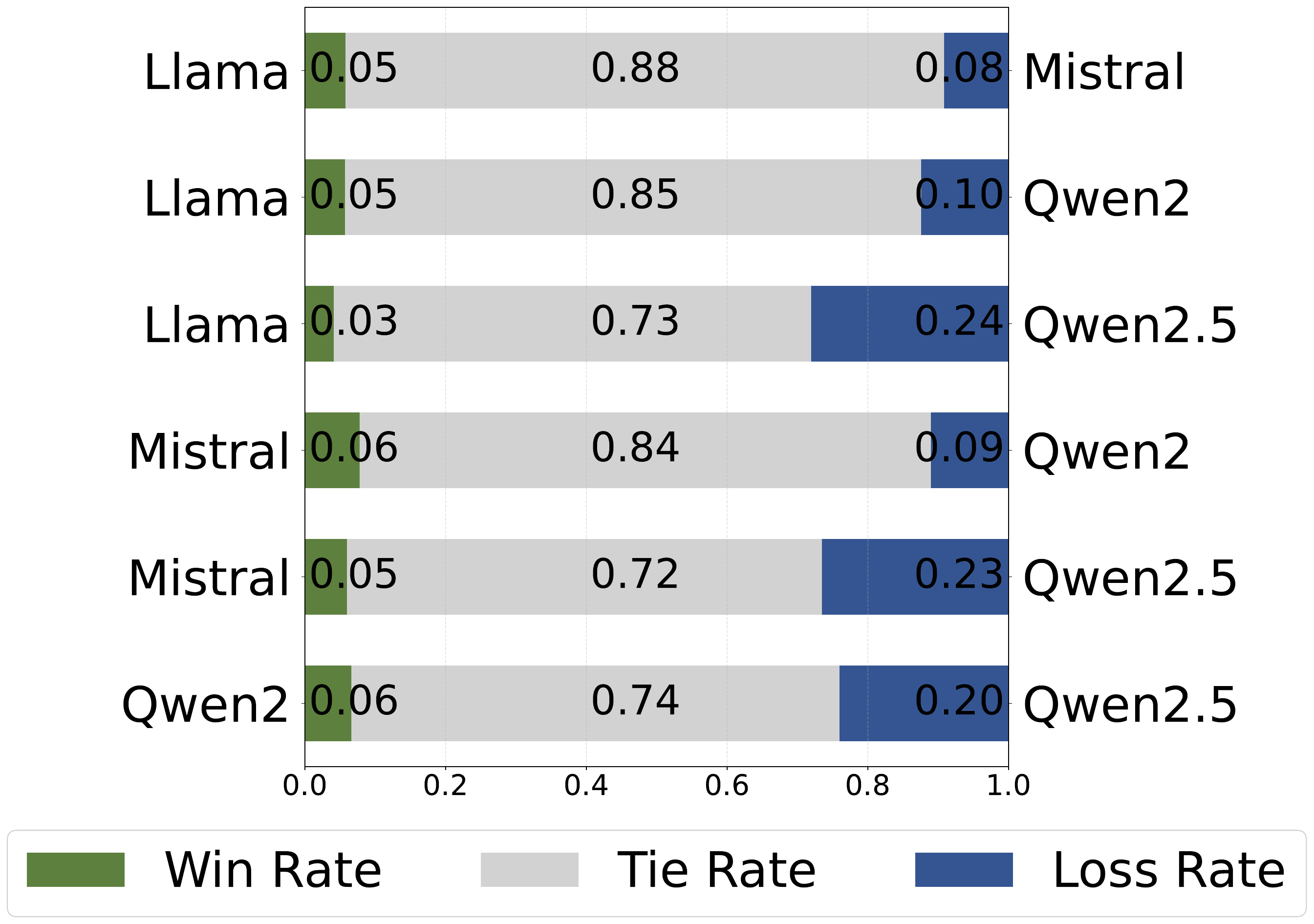} 
    \end{minipage}
  \caption{LLM performances (bottom:  AlpacaEval).}
  \label{fig: llm performances}
\end{minipage}
\end{figure*}

\section{Assessment: Experiment Setup}
\label{sec: Experimental Setup}
We provide the implementation of our LLM-PeerReview
and all baselines for the used datasets:
\pdflink{https://github.com/zeyuji/LLM-PeerReview/}{\textcolor{blue!60!black}{\faGithub\enspace Code}}.

% % We provide more details in the Appendix, including experimental setup, additional results, and the prompts used.
% We provide the implementation of our LLM-PeerReview and all baselines for the used datasets in the Appendix.
% We commit to publicly releasing this project, including all code needed to reproduce the main experiments in Table~\ref{table: main results}.
%  % in our repository: 
%  % \href{https://github.com/zeyuji/LLM-PeerReview}{\textcolor{blue!60!black}{\faGithub\enspace{Code}}}.

\paragraph{Datasets and evaluation.}
We evaluate four widely used datasets spanning three task categories:
\textbf{(1) Factual Recall:}
TriviaQA~\cite{joshi2017triviaqa,guha2024smoothie} evaluates the \textit{accuracy} of model responses to factual questions across various domains, including history, science, and geography;
\textbf{(2) Arithmetic Reasoning:}
GSM8k~\cite{chen2021evaluating,hu2024language} and MATH~\cite{hendrycks2021measuring, hu2024language} assess basic arithmetic and more advanced mathematical reasoning, respectively, with \textit{accuracy} as the evaluation metric, focusing on correct numerical answers;
\textbf{(3) Instruction Following:}
AlpacaEval~\cite{dubois2023alpacafarm,hu2024language} tests models' ability to follow various instructions. We use GPT-4o-mini to evaluate the \textit{accuracy} of model responses, assessing whether the model’s response exceeds the reference answer in the dataset.
We use the same dataset versions as prior studies~\cite{guha2024smoothie,hu2024language}.

\paragraph{Seed LLMs and baselines.}
% Considering 7B-scale models are widely used by researchers and generally regarded as having acceptable judging capabilities~\cite{wang2025improving, kim2024prometheus}, 
% We use these well-established 7B models for ensemble: 
We evaluate our method using several widely adopted open-source LLMs:
Llama-3.1-8B-Instruct, Mistral-7B-Instruct, Qwen2-7B-Instruct, and Qwen2.5-7B-Instruct.
% Furthermore, we employ the instruction-tuned versions of these models rather than their base counterparts. This choice allows us to leverage their instruction-following capabilities, which are essential for the LLM-as-a-judge tasks.
% \paragraph{Baselines.}
We compare the proposed LLM-PeerReview with the  two categories of baselines. 
\textbf{(1) Single LLMs:}
The four LLMs;
% , Llama-3.1-8B-Instruct, Mistral-7B-Instruct,
% Qwen2-7B-Instruct, and Qwen2.5-7B-Instruct.
\textbf{(2) LLM Ensemble baselines:}
(\textit{i}) Random~\cite{lu2024blending,guha2024smoothie} is a random-selection baseline that simply returns the response from a randomly chosen LLM. 
As one of the simplest ensemble strategies, 
% for large language models, 
this method has previously been applied to dialogue tasks~\cite{lu2024blending};
(\textit{ii}) Smoothie-Global~\cite{guha2024smoothie}, Smoothie-Local~\cite{guha2024smoothie}, and Agent-Forest~\cite{li2024more} are recently proposed, strong similarity-based ensemble methods, as introduced in detail in Section~\ref{sec: Introduction};    
(\textit{iii}) 
GaC~\cite{yu2024breaking} is a representative \textit{token-level ensemble-during-inference} approach. 
It constructs a unified vocabulary that merges the individual dictionaries of LLMs. During inference,  sampling is performed by observing the distributions from these models across the unified vocabulary;
(\textit{iv}) SAFE~\cite{yun2026when} and CoRE~\cite{zeng2026harnessing} are two recent \textit{ensemble-during-inference} methods. SAFE adopts the popular idea of speculative decoding and can be coupled with different base ensemble methods, yielding SAFE-GaC and SAFE-UniTE. CoRE uses token- and model-level consistency to assign weights during aggregation, yielding CoRE-GaC and CoRE-UniTE;
(\textit{v}) MAD~\cite{liang2024encouraging} is a classic multi-model collaboration method that can be applied to LLM Ensemble, where models refine their responses through debate and select the final answer by majority voting.

\paragraph{Configurations.}
(1) For each individual LLM, we follow the setup of Smoothie~\cite{guha2024smoothie}, where the model responds once to each query. 
The responses from all models are stored for integration by the LLM Ensemble methods;
(2) For the two variants of the baseline Smoothie~\cite{guha2024smoothie}, we set the number of neighbors as specified in the original paper. 
Agent-Forest~\cite{li2024more} does not require any hyperparameter configuration. 
For our method, we set the model temperature to 0 during the scoring process; 
% to eliminate suboptimal results caused by randomness;
% Additionally, the scoring prompts used across the four datasets are provided in the Appendix.
(3) All experiments were performed using 6 parallel Nvidia V100 32GB GPUs.
All experiments with stochastic outputs were conducted three times.

\begin{figure*}[t] 
	\centering  
	\vspace{-0.0cm} 
	\subfigtopskip=-0pt 
	\subfigbottomskip=0pt 
	\subfigcapskip=-5pt

    \subfigure[TriviaQA]{
		\label{level.sub.3}
		\includegraphics[width=0.5\linewidth]{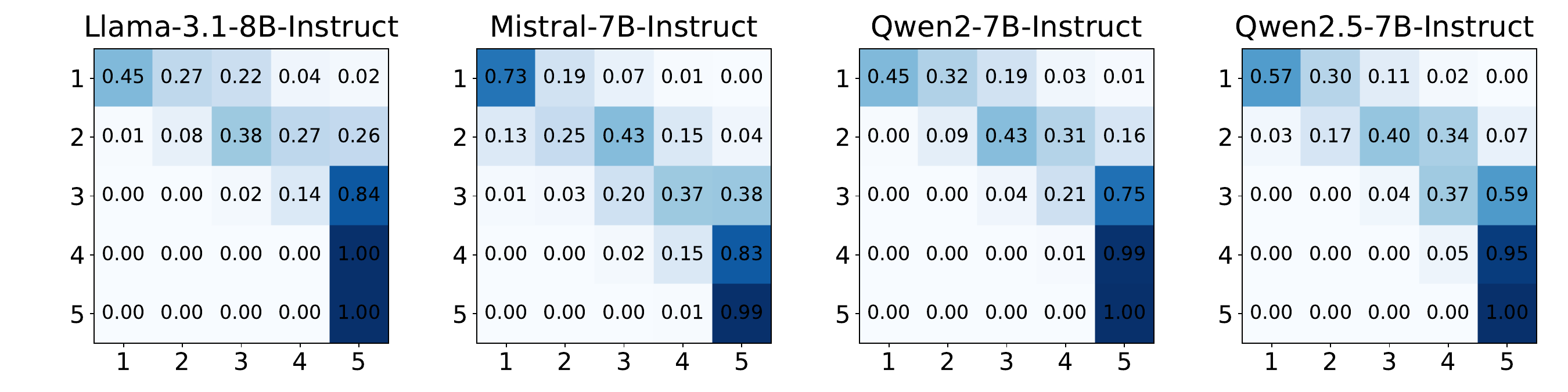}}
        \hspace{16mm}
	\subfigure[TriviaQA]{
		\label{level.sub.4}
		\includegraphics[width=0.16\linewidth]{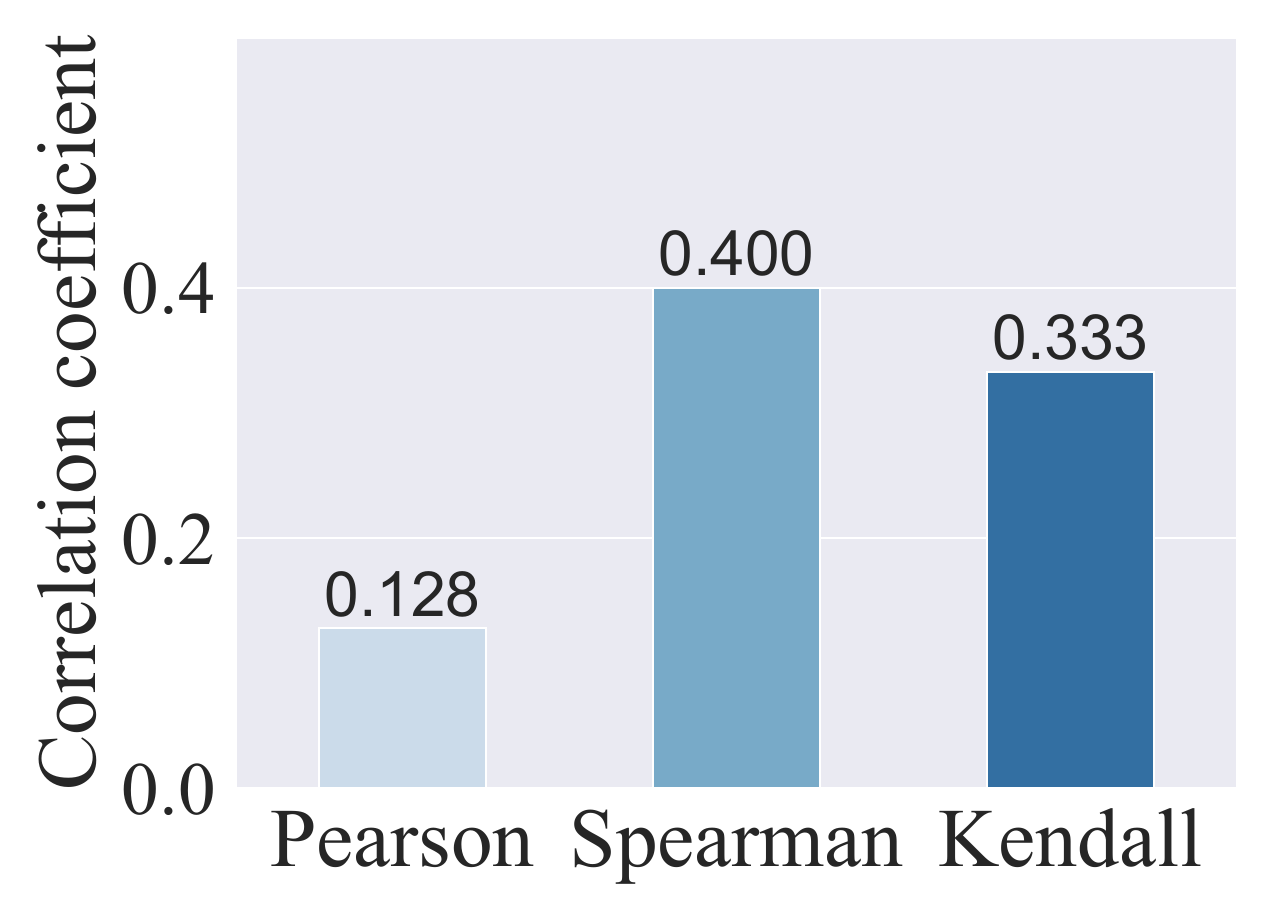}}
    
  	\subfigure[GSM8k]{
		\label{level.sub.3}
		\includegraphics[width=0.5\linewidth]{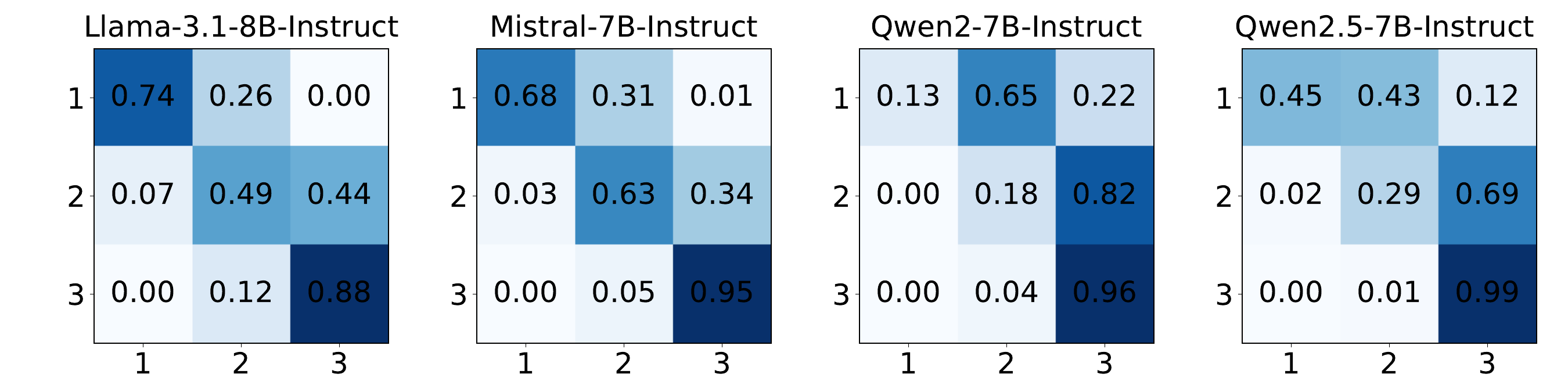}}
        \hspace{16mm} 
	\subfigure[GSM8k]{
		\label{level.sub.4}
		\includegraphics[width=0.16\linewidth]{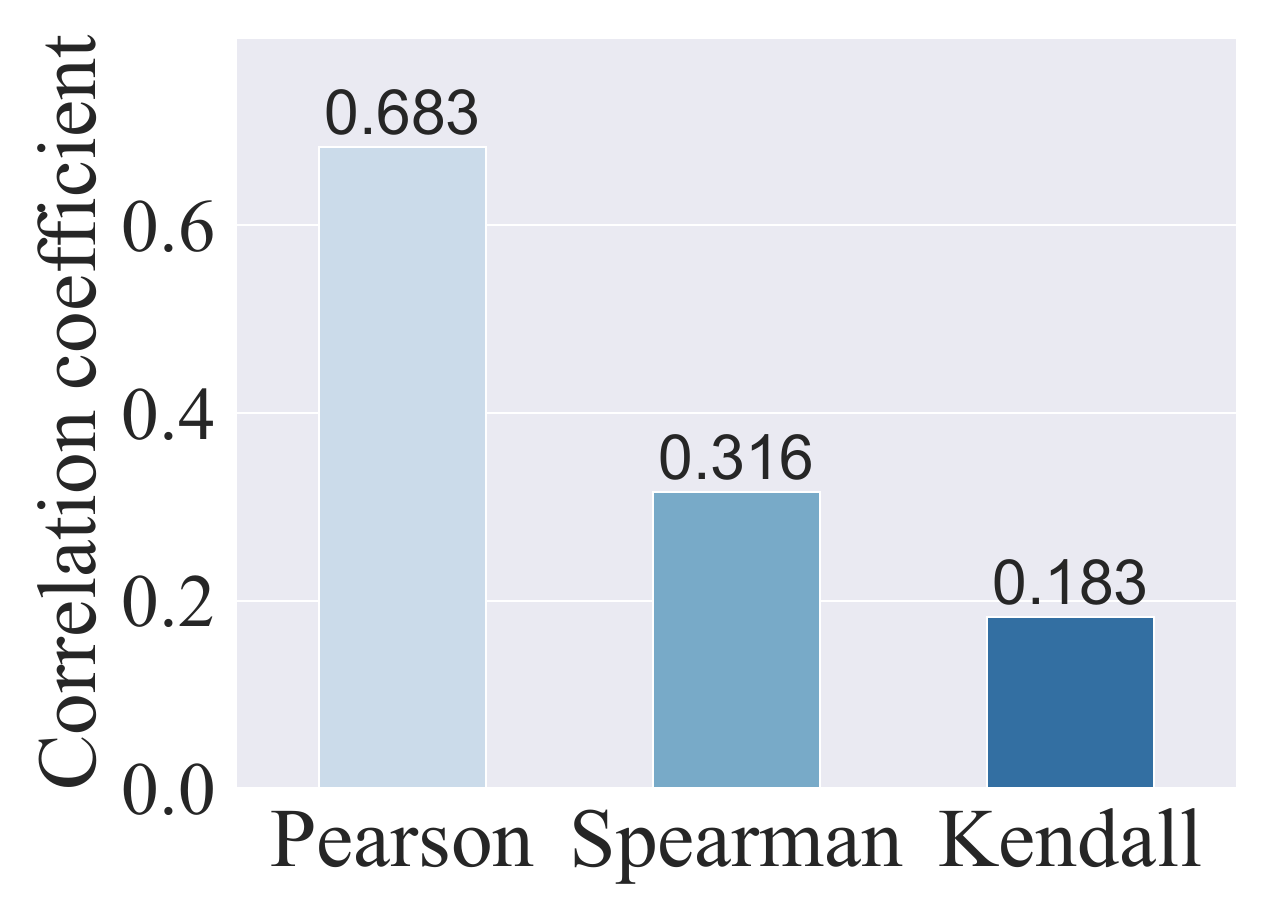}}

 %  	\subfigure[MATH]{
	% 	\label{level.sub.3}
	% 	\includegraphics[width=0.4\linewidth]{figures/matrix-math.pdf}}
 %        \hspace{16mm}
	% \subfigure[MATH]{
	% 	\label{level.sub.4}		
 %        \includegraphics[width=0.13\linewidth]{figures/correlation-math.pdf}}

 %  	\subfigure[AlpacaEval]{
	% 	\label{level.sub.3}
	% 	\includegraphics[width=0.4\linewidth]{figures/matrix-alpacaeval.pdf}}
 %        \hspace{16mm}
	% \subfigure[AlpacaEval]{
	% 	\label{level.sub.4}
	% 	\includegraphics[width=0.13\linewidth]{figures/correlation-alpaca.pdf}}

\caption{Analysis on TriviaQA and GSM8k (more results is provided in the Appendix). \textbf{Left:}  The transition matrix of each LLM estimated by LLM-PeerReview-W. \textbf{Right:} 
Correlation between matrix diagonal information of each LLM and its performance as a single judge (corresponding to ``our variants'' in Table~\ref{table: main results}). 
% For the first three datasets with ground-truth answers, diagonal information is represented by extreme values $(\pi_{1 1}^{(j^{\prime})} + \pi_{K K}^{(j^{\prime})})$.
% for the instruction-following dataset AlpacaEval, the sum of all diagonal values is used.
} 
\label{fig: transition matrix and correlation}
\end{figure*}

\section{Assessment: Results and Analysis}
\label{sec: Results}

\subsection{Main Results}
\label{sec: Main Results}

\textbf{The ensemble of the proposed LLM-PeerReview is effective.}
The main results are shown in Table~\ref{table: main results}.
First, by examining the results in the ``Single LLM'' and ``LLM Ensemble'' sections of Table~\ref{table: main results}, one key finding is that both LLM-PeerReview and LLM-PeerReview-W consistently outperform any single LLM and all LLM Ensemble baselines across all datasets.
In the last column, which presents the average performance, our two variant methods (with results of 67.4\% and 67.8\%) surpass the strongest single model, Qwen2.5, by 4.7\% and 5.1\%, respectively, and outperform the advanced ensemble method, Smoothie-Global, by 6.9\% and 7.3\%.
These results directly demonstrate the effectiveness of our method, as it achieves superior performance by integrating the collective knowledge of multiple models across factual-recall QA tasks, math reasoning tasks, and instruction-following tasks.
% Also, the ensemble task across these datasets is challenging, as the performance of these LLMs varies significantly for each dataset.
Also, the ensemble task across these four datasets is challenging, as the performance of the four LLMs varies significantly for
each dataset.

\textbf{Each LLM has its strengths and weaknesses.}
In Figure~\ref{fig: llm performances}, the upper subplot presents a radar chart of individual LLM performance, while the lower subplot displays the win-tie-loss chart for models on the challenging instruction-following dataset, AlpacaEval. This figure highlights that models with the best overall performance may underperform on specific tasks compared to those with weaker overall results.
In summary, the results in Table~\ref{table: main results} and Figure~\ref{fig: llm performances} demonstrate that a strong LLM does not excel across all datasets. Each model has its strengths and weaknesses, highlighting the substantial practical significance of LLM Ensemble.

\subsection{Advantages of Aggregating Multiple Judges}
\label{sec: Advantages of Aggregating Multiple Judges}

\textbf{Simply averaging the scores from multiple judges is quite effective.}
In the ``Our variants'' of Table~\ref{table: main results}, we present the performance of using a single LLM as a judge to select the optimal response.
From the average performances in the last column of Table~\ref{table: main results}, we observe that these variants perform quite well (surpassing the overall best model, Qwen2.5, in 3/4 cases).
On the other hand, when comparing the performance of these variants with that of our prototype LLM-PeerReview, it becomes clear that aggregating and averaging the scores from multiple judges is  beneficial.
% , compared to relying on the score of a single large model.

\textbf{The weighted truth inference has the potential for further performance improvement.}
By observing the average results in Table~\ref{table: main results}, we find that LLM-PeerReview-W leads to further performance gains compared to LLM-PeerReview.
In the left subplot of Figure~\ref{fig: transition matrix and correlation}, we observe subtle variations in the transition matrices learned for each model. 
On the other hand, the right subplot in Figure~\ref{fig: transition matrix and correlation}, displaying positive correlation coefficients, demonstrates that our method can effectively identifies stronger and weaker judges.
% We naturally expect that adopting a Bayesian perspective and providing distinct prior information for the transition matrices of different judges will further enhance  LLM-PeerReview-W.

\subsection{Alternative Designs and Efficiency Analysis.}
\label{Alternative Designs for the Peer-Review Process}

\begin{figure*}[htb]
\centering
\begin{minipage}{0.5758\textwidth}
\begin{center}
\resizebox{1\linewidth}{!}{
\begin{tabular}{lcccc  >{\columncolor{lightgray!17}}c}
\toprule
   \textbf{Method}          &  \textbf{TriviaQA}          & \textbf{GSM8k}      & \textbf{MATH}                 & \textbf{AlpacaEval}                        &      \textbf{Average}
     \\ \midrule
\multicolumn{6}{l}{\cellcolor[HTML]{DDDDDD}\textbf{(a)} Ours: Variant Performance ($\uparrow$): } \\
% \quad Random &  65.7 \scriptsize{$\pm$ 1.3}  &  81.3 \scriptsize{$\pm$ 0.6}  &  51.2 \scriptsize{$\pm$ 1.8}    &  14.7 \scriptsize{$\pm$ 0.5}  &  53.2 \\
\quad Single &  69.2 \scriptsize{$\pm$ 0.6}  &   85.5 \scriptsize {$\pm$ 2.1}  & 60.3 \scriptsize{$\pm$ 1.6}    &     23.8 \scriptsize{$\pm$ 1.0}    &   59.7   \\
\quad Double & 73.3 \scriptsize{$\pm$ 0.5}   &   90.0 \scriptsize{$\pm$ 0.7}    & 71.3 \scriptsize{$\pm$ 0.2}&    29.2 \scriptsize{$\pm$ 0.2}      &    66.0   \\
\quad Flipped-triple &  74.5 \scriptsize{$\pm$ 0.0}    &  90.8 \scriptsize{$\pm$ 0.2}   &  71.5 \scriptsize{$\pm$ 0.4} &    30.5 \scriptsize{$\pm$ 0.0}     &   66.8  \\
\quad Quadruple-half &  74.7 \scriptsize{$\pm$ 0.2}   &  91.5 \scriptsize{$\pm$ 0.4}    & 73.3 \scriptsize{$\pm$ 0.2}  &   29.2 \scriptsize{$\pm$ 0.2}  &   67.2     \\
\multicolumn{6}{l}{\cellcolor[HTML]{DDDDDD}\textbf{(b)}  Ours: Efficiency for Scoring (GPU hours/1 judge scoring the all generated responses;  $\downarrow$): } \\ 
% \quad Single ($\mathcal{O}(J)$)
% & 7.89 & 10.2  & 10.6 & 16.9 & 11.4
% \\
% \quad Double ($\mathcal{O}(J^2)$)
% & 37.1  &  49.4  & 51.6 & 77.4 & 53.9
% \\
% % \texttt{Early stopping \hl{patient}}

% \quad Flipped-triple ($\mathcal{O}(J)$)
% & 29.7  & 43.4  & 47.1 & 74.3 & 48.6
% \\
% \quad Quadruple-half ($\mathcal{O}(J!)$)
%   &  51.3 &  83.8 & 90.0 & 137.65 & 90.7 \\   
\quad Single ($\mathcal{O}(J)$)
& 2.63 & 3.4 & 3.5 & 5.6 & 3.8
\\
\quad Double ($\mathcal{O}(J^2)$)
& 12.4 & 16.5 & 17.2 & 25.8 & 18.0
\\
% \texttt{Early stopping \hl{patient}}
\quad Flipped-triple ($\mathcal{O}(J)$)
& 9.9 & 14.5 & 15.7 & 24.8 & 16.2
\\
\quad Quadruple-half ($\mathcal{O}(J!)$)
  & 17.1 & 27.9 & 30.0 & 45.9 & 30.2 \\
  % \vspace{5pt}
  \multicolumn{6}{l}{\cellcolor[HTML]{DDDDDD}\textbf{(c)}  Ours: Efficiency for Reasoning (seconds;  $\downarrow$): } \\ 
\quad Averaging/Weighted
& 0.7/2.0 & 0.6/0.8  & 0.9/0.8 & 1.0/0.7 & 0.8/1.1
\\
  \multicolumn{6}{l}{\cellcolor[HTML]{DDDDDD}\textbf{(d)} Efficiency Comparison: LLM-PeerRivew v.s. Baseline MAD (GPU hours;  $\downarrow$): } \\ 
% \quad  LLM-PeerReivew Efficient Reasoning
% & 17.8 & 29.0  & 31.4 & 49.5 & 32.0\\ 
\quad  LLM-PeerReivew cost
& 35.6 & 57.9 & 62.8 & 99.1 & 63.8\\ 
\quad Baseline MAD cost
& N/A & 104.7  & 80.4 & N/A & -\\ 
 \bottomrule 
\end{tabular}}
\end{center}
\captionof{table}{Efficiency/cost analysis.
\textbf{(a):} Performance of the LLM-PeerReview with different scoring strategies. Note that, for computation efficiency, we use 200 samples from each dataset (all shuffled as necessary).
% (\textit{i}) \textit{Random} refers to the same baseline in Table~\ref{table: main results}; 
(\textit{i}) \textit{Single} refers to scoring a single response at a time;
(\textit{ii}) \textit{Double} refers to scoring all \textit{response pairs} within the response set at a time;
(\textit{iii}) \textit{Quadruple-half} refers to scoring all possible \textit{response quadruple sequences} in the response set; given the high computational cost, we used a relaxed version and only calculated half of the cases.
\textbf{(b)/(c)/(d):}
Computation efficiency for LLM-PeerReview's and baseline MAD.
}
\label{table: more variants}
\end{minipage}%
\hfill
\begin{minipage}{0.39\textwidth}
\centering
\includegraphics[width=\textwidth]{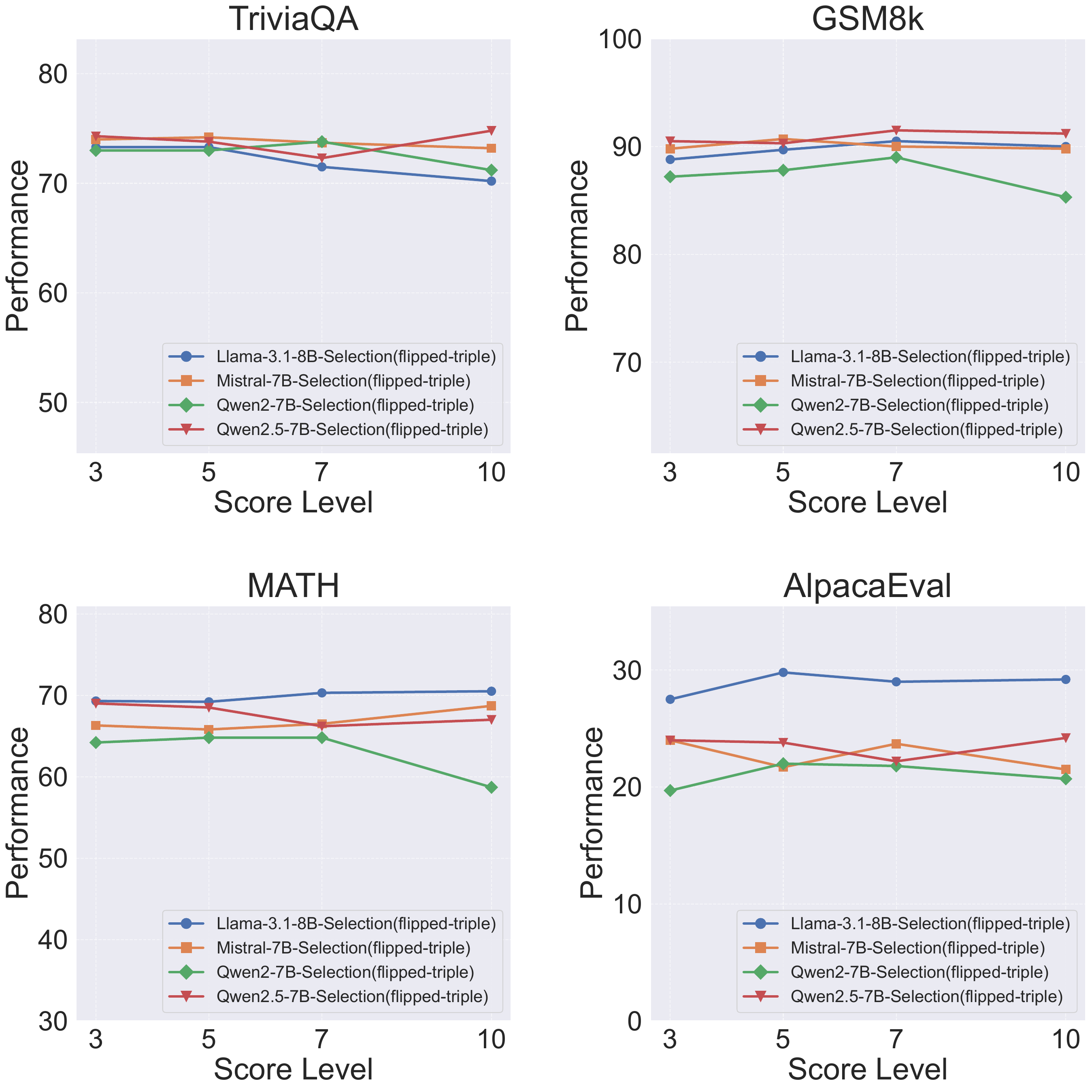}
\caption{Performance of various variants across different scoring levels.}
\label{fig: different scoring levels}
\end{minipage}
\end{figure*}

\textbf{The flipped-triple scoring trick embodies a performance-efficiency trade-off and is vital to the framework's overall efficacy.}
(1)
It is intuitive that, in addition to the our recommended \textit{flipped-triple scoring} method, several variant scoring methods could be employed.
(their definitions are provided in the caption of Table~\ref{table: more variants}.
Overall, the performance of these four variants follows the order: $\textit{quadruple-half} > \textit{flipped-triple}  > \textit{double} > \textit{single}$.
Variants quadruple-half, flipped-triple, and double all offer noticeable \textit{de-biasing} performance advantages over the single-scoring strategy.
On the other hand, in terms of theoretical computational complexity, the complexities for de-biased strategies double/flipped-triple/quadruple-half  are $\mathcal{O}(J)$/$\mathcal{O}(J^2)$/$\mathcal{O}(J)$/$\mathcal{O}(J!)$, with strategy flipped-triple having the low computational complexity.
Further, in Table~\ref{table: more variants}, we present the scoring efficiency of these four strategies. 
Compared to strategies double and quadruple-half, strategy flipped-triple is the most time-efficient.
(2)
\textit{Furthermore, the remarkable performance gain—from 59.7 with the original \textit{single}  to 66.8 with the proposed \textit{flipped triple} —validates the profound debiasing impact of our scoring technique.
This technique serves as a crucial cornerstone of LLM-PeerReview.}

% \textbf{LLM-PeerReview is more efficient than debate-based methods.} Table~\ref{table: more variants}(d) compares the additional inference cost of LLM-PeerReview and MAD, excluding their shared response-generation stage. On GSM8K and MATH, LLM-PeerReview requires substantially less GPU time. This is mainly because each judge only outputs a few scalar \textit{scores} in LLM-PeerReview, whereas MAD requires every model to generate longer \textit{revised full responses} over two debate rounds. Further details are provided in the Appendix.

\textbf{LLM-PeerReview is substantially more computationally efficient than debate-based methods.}
Table~\ref{table: more variants}(d) compares their post-generation inference costs, excluding the shared response-generation stage. 
% Compared with MAD, LLM-PeerReview reduces GPU time by 44.7\% on GSM8K and 21.9\% on MATH. 
On GSM8K and MATH, LLM-PeerReview requires substantially less GPU time. 
\textit{This advantage mainly comes from its simpler interaction pattern: our scoring stage requires only one round, in which each judge outputs a few scalar scores, whereas MAD performs two debate rounds and requires every model to generate much longer revised full responses in each round. }
Also, as shown in Table~\ref{table: more variants}, our final selecting step incurs virtually no additional computational cost. 
% Further details are provided in the Appendix.

\textbf{Reducing the number of judges further realizes the performance-efficiency trade-off.}
As discussed in Section~\ref{sec: others}, our framework allows for flexibility in reducing the judge count during deployment to further enhance efficiency, building upon the efficient flipped-triple scoring strategy.
The Appendix also shows that LLM-PeerReview remains effective in a lightweight setting: using fewer judges during scoring still outperforms the baselines.

% When employing only a single model as a judge—where LLM-PeerReview simply selects the highest-scoring response without the reasoning process—all four variants of our method still substantially outperform all baselines. This includes the computationally intensive GaC, with average leads of $6.2\%$, $6.5\%$, $1.9\%$, and $5.3\%$. 
% Regarding computational cost, as shown in 
% Table~\ref{table: more variants},
% % Table~\ref{table: more variants}, 
% LLM-PeerReview (1 judge) is more cost efficiency than GaC.~\footnote{For instance, on the TriviaQA dataset, the combined time of 29.7s for LLM-PeerReview and 6.1s for response generation is less than the 46.8s required by GaC. Moreover, the computational overhead incurred by the reasoning phase within LLM-PeerReview is virtually negligible.} 

\textbf{Common scoring levels can generally be attempted.}
In Figure~\ref{fig: different scoring levels}, we conduct a further analysis of how different scoring levels influence the performance of flipped-triple  judges.
For each scoring level, we have carefully crafted meaningful descriptions and corresponding prompts. 
% We use the basic variant LLM-PeerReview for this analysis.
Under these conditions, these judges exhibit  varying performance, showing no consistent tendencies across the levels of 3, 5, 7, and 10.
In addition, as indicated by the prompts in the Appendix, for the main experiment in Table~\ref{table: main results}, the scoring levels for the four datasets were 5, 3, 3, and 10, respectively.

\subsection{Further Analysis in the Appendix.}
\label{sec: Further Analysis}

We provide more comprehensive experimental analyses in the Appendix, including
% all of which further validate the findings presented in the main sections. The supplementary content includes: 
 performance results of LLM-PeerReview variants using the ``Single'' scoring strategy, win-tie-loss charts across all models on four datasets, 
full analysis for the transition matrix, 
additional results using varying numbers of judges, the performance of LLM-PeerReview with 13B-scale LLMs, more efficiency/cost analysis for LLM-PeerReview and four case studies across the four datasets respectively.

\section{Related Work}
\label{sec: Related Work}

\textbf{LLM Ensemble}, 
% as outlined before,
can be broadly categorized into three approaches.
% ~\cite{chen2025harnessing}.
The first category, \textit{ensemble-before-inference} approach~\cite{shnitzer2023large,srivatsa2024harnessing,ong2024routellm}, typically necessitates custom-labeled data to pretrain a classifier that routes each query.
The mandatory pretraining phase and the dependency on labeled data are the primary inconveniences of these methods.
The second category, \textit{ensemble-during-inference} approach, can be subdivided into \textit{token-level}~\cite{yu2024breaking,huang2024ensemble,xu2024bridging,zou2025transformer,chen2025llmboostmakelargelanguage}, \textit{span-level}~\cite{liu2024cool,xu2025hit}, and \textit{process-level}~\cite{park2024ensembling} methods, depending on the level of granularity of the information considered in the ensemble.
% These methods, however, entail substantial computational costs and require the local deployment of every LLM in the ensemble.
Among them, some methods~\cite{zou2025transformer,chen2025llmboostmakelargelanguage} require post-training~\cite{yang2026grouprelativeadvantagebiased}.
Lastly, 
as mentioned before, 
\textit{ensemble-after-inference} approach mainly includes selection-then-regeneration-based~\cite{lu2024blending} and aggregation-based methods~\cite{li2024more,guha2024smoothie}.
Also, there are several approaches closely related to LLM Ensemble
% , such as using multiple prompts
% for ensemble inference
~\cite{he2025llm}.

\textbf{Weak Supervision}~\cite{zhang2021wrench,chen2023neural}, also commonly referred to as \textbf{Learning from Crowds}~\cite{chen2021structured,chen2022adversarial}, is a research problem that closely resembles post-inference LLM ensemble.
The key distinction is that Weak Supervision methods either focus on learning classifiers directly from imperfectly labeled data~\cite{zhang2021wrench} or on aggregating weak label information~\cite{zheng2017truth,chen2022adversarial}, whereas ensemble-after-inference LLM Ensemble methods are exclusively concerned with aggregation and do not involve the learning of classifier models.
% Additionally, a significant difference is that most weakly supervised learning methods are primarily focused on classification 
% % scenarios 
% with closed answer sets, rather than open-ended generation. 
% tasks. 
% text generation tasks that involve open-ended output spaces.
Also, most weak supervision methods are primarily focused on traditional classification,  
% scenarios 
% with closed answer sets, 
rather than open-ended generation.

\textbf{LLM-as-a-Judge} approaches has received a lot of attention recently. 
As task complexity increases and model outputs become more diverse, traditional evaluation methods—such as matching-based or embedding-based metrics—often fail to capture subtle attributes. 
% and provide reliable results.
% ~\cite{gu2024survey}. 
The recent emergence of LLMs
% large language models
has led to 
% the development of 
the ``LLM-as-a-judge'' paradigm, where LLMs assess the quality of model outputs. 
These methods can be broadly classified into three categories: Single-LLM-based, Multi-LLM-based, and Human-AI collaboration-based evaluation approaches. Single-LLM-based methods primarily focus on prompt design~\cite{fu2023gptscore}, 
% design~\cite{fu2023gptscore,kotonya2023little}, 
fine-tuning~\cite{chen2023adaptation}, 
% fine-tuning~\cite{chen2023adaptation, wu2024meta}, 
or post-processing~\cite{daynauth2024aligning}. 
Notably, Multi-LLM-based methods, which include collaborative~\cite{zhang2023wider}, competitive~\cite{owens2024multi}, and aggregation-based~\cite{verga2024replacing,chen2024automatic} 
% aggregation-based~\cite{verga2024replacing,chen2024automatic,chu2024pre} 
strategies, are closely related to our approach.
% (especially for aggregation methods).

\section{Conclusion}
This paper presents LLM-PeerReview, an  unsupervised peer-review-inspired method, to address the LLM ensemble problem.
PeerReview benefits both from the analytical capabilities of the powerful large language models at hand when evaluating response quality, and from the straightforward averaging
strategy or the principled, fine-grained score aggregation when inferring final scores.
LLM-PeerReview---embedded with the well-established techniques of LLM-as-a-Judge---closely emulates the real-world human process of selecting the best text, offering a clear and interpretable mechanism.
Empirical evaluations on four datatests demonstrate that LLM-PeerReview significantly improves the 
advanced model Smoothie-Global and provides a new solution to LLM Ensemble.
Additionally, we hope that our open-source release will help researchers reproduce LLM Ensemble methods.

% \newpage

% \section*{Acknowledgments}
% AAAI is especially grateful to Peter Patel Schneider for his work in implementing the original aaai.sty file, liberally using the ideas of other style hackers, including Barbara Beeton. We also acknowledge with thanks the work of George Ferguson for his guide to using the style and BibTeX files --- which has been incorporated into this document --- and Hans Guesgen, who provided several timely modifications, as well as the many others who have, from time to time, sent in suggestions on improvements to the AAAI style. We are especially grateful to Francisco Cruz, Marc Pujol-Gonzalez, and Mico Loretan for the improvements to the Bib\TeX{} and \LaTeX{} files made in 2020.

% The preparation of the \LaTeX{} and Bib\TeX{} files that implement these instructions was supported by Schlumberger Palo Alto Research, AT\&T Bell Laboratories, Morgan Kaufmann Publishers, The Live Oak Press, LLC, and AAAI Press. Bibliography style changes were added by Sunil Issar. \verb+\+pubnote was added by J. Scott Penberthy. George Ferguson added support for printing the AAAI copyright slug. Additional changes to aaai2027.sty and aaai2027.bst have been made by Francisco Cruz, Marc Pujol-Gonzalez, and Mico Loretan.

% \bigskip
% \noindent Thank you for reading these instructions carefully. We look forward to receiving your electronic files!

% \bibliography{aaai2027}
\bibliography{ref}

\newpage
\appendix
\onecolumn

\appendix

\section{Appendix}
\label{sec: appendix}

\setcounter{table}{0}
\renewcommand{\thetable}{A.\arabic{table}}

\setcounter{figure}{0}
\renewcommand{\thefigure}{A.\arabic{figure}}

In this appendix, we present the following sections to further support the content of the main paper.
\begin{itemize}

\item 
\textbf{Section~\ref{sec: Proof of the Optimization}:} The optimization process of the graphical model constructed in the main text;
\item 
\textbf{Section~\ref{sec: Theoretical Analysis of Scoring Phase}:} 
% Theoretical Analysis of Scoring Phase;
 Theoretical analysis for the scoring phase: error-ambiguity  decomposition;
\item 
\textbf{Section~\ref{sec: Summary of notations used in LLM-PeerReview}:} Notation and algorithm overview;
\item 
\textbf{Section~\ref{sec: More Experiment Setup}:} More details on the experimental setup;
\item 
\textbf{Section~\ref{sec: More Experimental Results}:} More experimental results;
\item 
\textbf{Section~\ref{sec: Prompts}:} The prompts used in the experiments.
\end{itemize}

\subsection{Proof of the Optimization}
\label{sec: Proof of the Optimization}
The optimization objective is to maximize the log conditional likelihood of the observed scoring labels $ \mathbf{Y} = \{y^{(i,j;j^{\prime})} \mid 1 \le i \le I,\ 1 \le j \le J ,\ 1 \le j^{\prime} \le J \}$ contributed by $J$ LLM judges, i.e., $\log p(\mathbf{Y} ; \Theta)$.
In brief, as with most latent variable models~\cite{bishop2006pattern,dawid1979maximum}, we apply the Expectation-Maximization (EM) algorithm~\cite{dempter1977maximum} to solve the optimization problem.

First, the log-likelihood is:
% \begingroup
% \small
% \allowdisplaybreaks[4]
% \begin{align}
% \log p(\mathbf{Y};\Theta)
% &=
% \log\prod_{i=1}^{I}\prod_{j=1}^{J}
% p(\mathbf{y}^{(i,j)};\Theta)
% \notag\\
% &=
% \sum_{i=1}^{I}\sum_{j=1}^{J}
% \log p(\mathbf{y}^{(i,j)};\Theta)
% \notag\\
% &=
% \sum_{i=1}^{I}\sum_{j=1}^{J}
% \log\sum_{t^{(i,j)}}
% p(\mathbf{y}^{(i,j)},t^{(i,j)};\Theta)
% \notag\\
% &=
% \sum_{i=1}^{I}\sum_{j=1}^{J}
% \log\!\left[
% q(t^{(i,j)})
% \sum_{t^{(i,j)}}
% \frac{
% p(\mathbf{y}^{(i,j)},t^{(i,j)};\Theta)
% }{
% q(t^{(i,j)})
% }
% \right]
% \notag\\
% &\geq
% \sum_{i=1}^{I}\sum_{j=1}^{J}
% \sum_{t^{(i,j)}}q(t^{(i,j)})
% \log
% \frac{
% p(\mathbf{y}^{(i,j)},t^{(i,j)};\Theta)
% }{
% q(t^{(i,j)})
% }
% \notag\\
% &\geq
% \sum_{i=1}^{I}\sum_{j=1}^{J}
% \Bigg[
% \mathbb{E}_{q(t^{(i,j)})}
% \log p(t^{(i,j)};\boldsymbol{\alpha})
% \notag\\
% &\qquad
% +\mathbb{E}_{q(t^{(i,j)})}
% \log\prod_{j^{\prime}=1}^{J}
% p\!\left(
% y^{(i,j;j^{\prime})}\mid t^{(i,j)};
% \boldsymbol{\Pi}^{(j^{\prime})}
% \right)
% \notag\\
% &\qquad
% -\mathbb{E}_{q(t^{(i,j)})}
% \log q(t^{(i,j)})
% \Bigg].
% \tag{A.1}
% \label{eq: appendix a1}
% \end{align}
% \endgroup

\begin{small}
\begin{equation}
\begin{aligned}
\log p(\mathbf{Y}; \Theta)=&   \log \prod_{i=1}^{I}  \prod_{j=1}^{J}  p(\mathbf{y}^{(i,j)}  ; \Theta) \\
= & \sum_{i=1}^{I}  \sum_{j=1}^{J}  \log p(\mathbf{y}^{(i,j)}  ; \Theta)\\
= & \sum_{i=1}^{I}  \sum_{j=1}^{J} \log  \sum_{t^{(i,j)}} p(\mathbf{y}^{(i,j)}, t^{(i,j)}  ; \Theta),\\
= & \sum_{i=1}^{I}  \sum_{j=1}^{J} \log  [q(t^{(i,j)})  \sum_{t^{(i,j)}} \frac{p(\mathbf{y}^{(i,j)}, t^{(i,j)}  ; \Theta)}{q(t^{(i,j)})}],\\
\geq & \sum_{i=1}^{I}  \sum_{j=1}^{J}  \sum_{t^{(i,j)}} q(t^{(i,j)}) \log \frac{p(\mathbf{y}^{(i,j)}, t^{(i,j)}  ; \Theta)}{q(t^{(i,j)})}\\
\geq & \sum_{i=1}^{I}  \sum_{j=1}^{J}  [\mathbb{E}_{q(t^{(i,j)})} \operatorname{log} p(t^{(i,j)} ; \boldsymbol{\alpha} )+  
\mathbb{E}_{q(t^{(i,j)})} \log\prod_{j^{\prime}=1}^{J}  
 p({y}^{(i,j;j^{\prime})} \mid t^{(i,j)} ; \boldsymbol{\Pi}^{(j^{\prime})}) - 
 \mathbb{E}_{q(t^{(i,j)})} \log q({t}^{(i,j)})],\\
\end{aligned}
\label{eq: appendix a1}
\tag{A.1}
\end{equation}
\end{small}
where $ \mathbf{y}^{(i,j)} = \{y^{(i,j;j^{\prime})}\mid1 \le j^{\prime} \le J \}$ denotes the set of scores assigned to response $\mathbf{r}^{(i,j)}$ by the $J$ models.
The derivation which obtains the \textit{Evidence Lower Bound} (ELBO)
\begin{equation}
\sum_{i=1}^{I}  \sum_{j=1}^{J}  \sum_{t^{(i,j)}} q(t^{(i,j)}) \log \frac{p(\mathbf{y}^{(i,j)}, t^{(i,j)}  ; \Theta)}{q(t^{(i,j)})}
\label{eq: appendix elbo}
\tag{A.2}
\end{equation}
used Jensen’s inequality~\cite{bishop2006pattern}; 
and $q(t^{(i,j)})$ is a discrete distribution over the variable $t^{(i,j)}$.
% Note that we use $q(t^{(l)})$  to denote $q_{l}(t^{(l)})$  in the following derivation for simplicity.
In the following, we then proceed to apply the general EM recipe to perform iterative calculations
(concerning E-step and M-step) to solve the optimization problem:  $\Theta :=
\underset{\Theta}{\operatorname{argmax}} \log p(\mathbf{Y} ; \Theta)$.

% \begin{equation}
% \begin{aligned}
% \textbf{E-step (inference):    } q(t^{(i,j)}=k)& :=p(t^{(i,j)}=k \mid \mathbf{y}^{(i,j)}; \Theta) \\
% \ \qquad  \propto p(t^{(i,j)}=k; \boldsymbol{\alpha} ) \cdot  \prod_{j^{\prime}=1}^{J} & p(y^{(i, j;j^{\prime})} \mid t^{(i,j)}=k ; \boldsymbol{\Pi}^{(j^{\prime})}).
% \end{aligned}
% \label{eq: appendix posterior 1}
% \tag{A.3}
% \end{equation}

\begin{equation}
\begin{aligned}
\textbf{E-step (inference):}\qquad 
q(t^{(i,j)}=k)& :=p(t^{(i,j)}=k \mid \mathbf{y}^{(i,j)}; \Theta) \\
 &  \propto p(t^{(i,j)}=k; \boldsymbol{\alpha} ) \cdot  \prod_{j^{\prime}=1}^{J}  p(y^{(i, j;j^{\prime})} \mid t^{(i,j)}=k ; \boldsymbol{\Pi}^{(j^{\prime})}).
\end{aligned}
\label{eq: appendix posterior 1}
\tag{A.3}
\end{equation}
The posterior $q(t^{(i,j)})$
is obtained by using of Bayes's theorem given the parameters $\Theta=\{\boldsymbol{\alpha}, \boldsymbol{\Pi}^{(1)}, \ldots, \boldsymbol{\Pi}^{(J)}\}$ learned on the last M-step.
Given that we are likely to obtain decimal values rather than integers on $y^{(i, j;j^{\prime})}$ after the scoring phase, we make the adaptation:
% \begin{equation}
% \begin{aligned}
% q(t^{(i,j)}=k)
% &\propto
% p\!\left(t^{(i,j)}=k;\boldsymbol{\alpha}\right)
% \prod_{j^{\prime}=1}^{J}\Bigl[
% \\[-1pt]
% &\qquad
% \phi_l\,
% p\!\left(
% y_l^{(i,j;j^{\prime})}
% \mid t^{(i,j)}=k;
% \boldsymbol{\Pi}^{(j^{\prime})}
% \right)
% \\[-1pt]
% &\qquad
% +\phi_u\,
% p\!\left(
% y_u^{(i,j;j^{\prime})}
% \mid t^{(i,j)}=k;
% \boldsymbol{\Pi}^{(j^{\prime})}
% \right)
% \Bigr].
% \end{aligned}
% \tag{A.4}
% \label{eq: appendix posterior 2}
% \end{equation}
% \begin{equation}
% \begin{aligned}
% q(t^{(i,j)}=k)
% &\propto
% p\!\left(t^{(i,j)}=k;\boldsymbol{\alpha}\right)
% \\[-1pt]
% &\quad {}\times
% \prod_{j^{\prime}=1}^{J}
% \sum_{s\in\{l,u\}}
% \phi_s\,
% p\!\left(
% y_s^{(i,j;j^{\prime})}
% \mid t^{(i,j)}=k;
% \boldsymbol{\Pi}^{(j^{\prime})}
% \right).
% \end{aligned}
% \tag{A.4}
% \label{eq: appendix posterior 2}
% \end{equation}
\begin{small}
\begin{equation}
\begin{aligned}
q(t^{(i,j)}=k)\propto p(t^{(i,j)}=k; \boldsymbol{\alpha} ) \cdot \prod_{j^{\prime}=1}^{J}  [\phi_{l}\cdot
 p(y^{(i, j;j^{\prime})}_{l} \mid t^{(i,j)}=k ; \boldsymbol{\Pi}^{(j^{\prime})}) + \phi_{u}\cdot p(y^{(i, j;j^{\prime})}_{u} \mid t^{(i,j)}=k ; \boldsymbol{\Pi}^{(j^{\prime})})],
\end{aligned}
\label{eq: appendix posterior 2}
\tag{A.4}
\end{equation}
\end{small}
where $s\in\{l,u\}$ indexes the lower and upper nearest integer
neighbors of the decimal score $y^{(i,j;j^{\prime})}$, and $\phi_s$
denotes the corresponding interpolation weight.
% \begin{small}
% \begin{equation}
% \begin{aligned}
% q(t^{(i,j)}=k)\propto  & p(t^{(i,j)}=k; \boldsymbol{\alpha} ) \cdot \prod_{j^{\prime}=1}^{J}  [\phi_{l}\cdot
%  p(y^{(i, j;j^{\prime})}_{l} \mid t^{(i,j)}=k ; \boldsymbol{\Pi}^{(j^{\prime})})\\
%  &  + \phi_{u}\cdot p(y^{(i, j;j^{\prime})}_{u} \mid t^{(i,j)}=k ; \boldsymbol{\Pi}^{(j^{\prime})})],
% \end{aligned}
% \label{eq: appendix posterior 2}
% \tag{A.4}
% \end{equation}
% \end{small}
% \begin{small}
% \begin{equation}
% \begin{aligned}
% q(t^{(i,j)}=k)\propto p(t^{(i,j)}=k; \boldsymbol{\alpha} ) \cdot \prod_{j^{\prime}=1}^{J}  [\phi_{l}\cdot
%  p(y^{(i, j;j^{\prime})}_{l} \mid t^{(i,j)}=k ; \boldsymbol{\Pi}^{(j^{\prime})}) + \phi_{u}\cdot p(y^{(i, j;j^{\prime})}_{u} \mid t^{(i,j)}=k ; \boldsymbol{\Pi}^{(j^{\prime})})],
% \end{aligned}
% \label{eq: appendix posterior 2}
% \tag{A.4}
% \end{equation}
% \end{small}
where $\phi_{l}$ and $\phi_{u}$ represent the confidences  for the decimal $y^{(i, j;j^{\prime})}$ corresponding to its lower and upper nearest integer neighbors (i.e., $y^{(i, j;j^{\prime})}_{l}$, $y^{(i, j;j^{\prime})}_{u}$).

\begin{small}
\begin{equation}
\begin{aligned}
\textbf{M-}&\textbf{step (learning):}
\\
\Theta :=&\underset{\Theta}{\operatorname{argmax}} \sum_{i=1}^{I}  \sum_{j=1}^{J}  [\mathbb{E}_{q(t^{(i,j)})} \operatorname{log} p(t^{(i,j)} ; \boldsymbol{\alpha} )+ 
\mathbb{E}_{q(t^{(i,j)})} \log\prod_{j^{\prime}=1}^{J}  
 p({y}^{(i,j;j^{\prime})} \mid t^{(i,j)} ; \boldsymbol{\Pi}^{(j^{\prime})})-  \mathbb{E}_{q(t^{(i,j)})} \log q({t}^{(i,j)})]\\
:=&\underset{\Theta}{\operatorname{argmax}} \sum_{i=1}^{I}  \sum_{j=1}^{J}  [\mathbb{E}_{q(t^{(i,j)})} \operatorname{log} p(t^{(i,j)} ; \boldsymbol{\alpha} )+ \mathbb{E}_{q(t^{(i,j)})} \log\prod_{j^{\prime}=1}^{J}  
 p({y}^{(i,j;j^{\prime})} \mid t^{(i,j)} ; \boldsymbol{\Pi}^{(j^{\prime})})].\\
\end{aligned}
\tag{A.5}
\label{eq: appendix m-step}
\end{equation}
\end{small}

% \begin{small}
% \begin{equation}
% \begin{aligned}
% \textbf{M-}&\textbf{step (learning):}
% \\
% \Theta :=&\underset{\Theta}{\operatorname{argmax}} \sum_{i=1}^{I}  \sum_{j=1}^{J}  [\mathbb{E}_{q(t^{(i,j)})} \operatorname{log} p(t^{(i,j)} ; \boldsymbol{\alpha} )+ 
% \mathbb{E}_{q(t^{(i,j)})} \log\prod_{j^{\prime}=1}^{J}  
%  p({y}^{(i,j;j^{\prime})} \mid t^{(i,j)} ; \boldsymbol{\Pi}^{(j^{\prime})})-  \mathbb{E}_{q(t^{(i,j)})} \log q({t}^{(i,j)})]\\
% :=&\underset{\Theta}{\operatorname{argmax}} \sum_{i=1}^{I}  \sum_{j=1}^{J}  [\mathbb{E}_{q(t^{(i,j)})} \operatorname{log} p(t^{(i,j)} ; \boldsymbol{\alpha} )+ \mathbb{E}_{q(t^{(i,j)})} \log\prod_{j^{\prime}=1}^{J}  
%  p({y}^{(i,j;j^{\prime})} \mid t^{(i,j)} ; \boldsymbol{\Pi}^{(j^{\prime})})].\\
% \end{aligned}
% \tag{A.5}
% \label{eq: appendix m-step}
% \end{equation}
% \end{small}

Furthermore, by maximizing optimization objective in Equation~\ref{eq: appendix m-step} and using the standard Lagrange multiplier method~\cite{bishop2006pattern}, we can obtain the closed-form solution for $\boldsymbol{\alpha}=\{\alpha^{(k)}\mid1 \le k \le K \}$ in Equation~\ref{eq: appendix alpha} shown below; and by equating the gradient of Equation~\ref{eq: appendix m-step} to zero, we can obtain the closed-form solution for $\{\boldsymbol{\Pi}^{(j^{\prime})}\}_{j^{\prime}=1}^J$ in Equation~\ref{eq: appendix pi} shown below.
\begin{equation}
\alpha_k=\frac{\sum_{i=1}^I \sum_{j=1}^J q(t^{(i,j)}=k)}{I\cdot J},
\label{eq: appendix alpha}
\tag{A.6}
\end{equation}
\begin{equation}
\pi_{m n}^{(j^{\prime})}=\frac{\sum_{i=1}^{I}\sum_{j=1}^{J}  q(t^{(i,j)}=m) \cdot\Psi(y^{(i,j;j^{\prime})},n) }{\sum_{i=1}^{I}\sum_{j=1}^{J} q(t^{(i,j)}=m) }\mathrm{,}
\label{eq: appendix pi}
\tag{A.7}
\end{equation}
where $\Psi(y^{(i,j;j^{\prime})})=[\phi_{l}\cdot \mathbb{I}(y^{(i,j;j^{\prime})}_l=n) + \phi_{u}\cdot \mathbb{I}(y^{(i,j;j^{\prime})}_u=n]$, and $\mathbb{I}(\cdot)$ is an indicator function that takes the value $1$ when the internal declaration is true, and $0$ otherwise.

\subsection{Theoretical Analysis for the Scoring Phase: Error-Ambiguity Decomposition}
\label{sec: Theoretical Analysis of Scoring Phase}

In this section, we provide the formal derivation of the Error-Ambiguity Decomposition, which serves as the theoretical foundation for the Scoring phase in LLM-PeerReview and LLM-PeerReview-W.

\begin{theorem}[Error-Ambiguity Decomposition for Scoring Phase of LLM-PeerReview]
Consider a scoring task with a target function $f$. Let $\{\mathcal{M}_1, \mathcal{M}_2, \dots, \mathcal{M}_J\}$ be a set of $J$ individual models. The ensemble output $\bar{g}$ is defined as the straightforward average of these models:
\begin{equation}
    \bar{g}(\mathbf{r}) = \frac{1}{J} \sum_{j^{'}=1}^J \mathcal{M}_{j^{'}}(\mathbf{r}).
% \label{eq: appendix pi}
\tag{A.8}
\end{equation}
The relationship between the ensemble error $E_{ens}$, the average individual error $\bar{E}$, and the ambiguity (diversity) $\bar{A}$ is given by the identity:
\begin{equation}
    E_{ens} = \bar{E} - \bar{A},
    \tag{A.9}
\end{equation}
where $\bar{E} = \frac{1}{J} \sum_{j^{'}=1}^J (\mathcal{M}_{j^{'}} - f)^2$ and $\bar{A} = \frac{1}{J} \sum_{t=1}^J (\mathcal{M}_{j^{'}} - \bar{g})^2$.
\end{theorem}

\begin{proof}
Starting from the definition of the average individual generalization error $\bar{E}$:
\begin{equation}
    \bar{E} = \frac{1}{J} \sum_{j^{'}=1}^J (\mathcal{M}_{j^{'}} - f)^2.
    \tag{A.10}
\end{equation}
By introducing the ensemble term $\bar{g}$ into the expression $(\mathcal{M}_{j^{'}} - f)$, we have:
\begin{equation}
    \bar{E} = \frac{1}{J} \sum_{j^{'}=1}^J \left[ (\mathcal{M}_{j^{'}} - \bar{g}) + (\bar{g} - f) \right]^2.
    \tag{A.11}
\end{equation}
Expanding the quadratic term $(a+b)^2 = a^2 + b^2 + 2ab$:
% \begin{equation}
% \begin{aligned}
% \bar{E}
% &=
% \frac{1}{J}\sum_{j^{'}=1}^{J}
% (\mathcal{M}_{j^{'}}-\bar{g})^2
% +
% \frac{1}{J}\sum_{j^{'}=1}^{J}
% (\bar{g}-f)^2
% \\
% &\quad+
% \frac{2}{J}\sum_{j^{'}=1}^{J}
% (\mathcal{M}_{j^{'}}-\bar{g})(\bar{g}-f).
% \end{aligned}
% \tag{A.12}
% \label{eq: main eq for the theoretical analysisa on averging method}
% \end{equation}
\begin{equation}
    \bar{E} = \frac{1}{J} \sum_{j^{'}=1}^J (\mathcal{M}_{j^{'}} - \bar{g})^2 + \frac{1}{J} \sum_{j^{'}=1}^J (\bar{g} - f)^2 + \frac{2}{J} \sum_{j^{'}=1}^J (\mathcal{M}_{j^{'}} - \bar{g})(\bar{g} - f).
    \tag{A.12}
    \label{eq: main eq for the theoretical analysisa on averging method}
\end{equation}
We now analyze each term individually in the following.

 The first term in Equation~\ref{eq: main eq for the theoretical analysisa on averging method} is the average ambiguity of the individual models:
    \begin{equation*}
        \frac{1}{J} \sum_{j^{'}=1}^J (\mathcal{M}_{j^{'}} - \bar{g})^2 = \bar{A}.
            \tag{A.13}
    \end{equation*}
 The second term in Equation~\ref{eq: main eq for the theoretical analysisa on averging method} is the error of the ensemble, as $(\bar{g} - f)^2$ is constant with respect to $j^{'}$:
    \begin{equation*}
        \frac{1}{J} \sum_{j^{'}=1}^J (\bar{g} - f)^2 = (\bar{g} - f)^2 =  E_{ens}.
             \tag{A.14}
    \end{equation*}
 For the third term (the cross-term) in Equation~\ref{eq: main eq for the theoretical analysisa on averging method}, since $(\bar{g} - f)$ is independent of $j^{'}$, we can pull it out of the summation:
    \begin{equation*}
        \frac{2}{J} (\bar{g} - f) \sum_{j^{'}=1}^J (\mathcal{M}_{j^{'}} - \bar{g}) = \frac{2}{J} (\bar{g} - f) \left( \sum_{j^{'}=1}^J \mathcal{M}_{j^{'}} - J\bar{g} \right).
             \tag{A.15}
    \end{equation*}
    Given that $\bar{g} = \frac{1}{J} \sum_{j^{'}=1}^J \mathcal{M}_{j^{'}}$, the term $\left( \sum_{j^{'}=1}^J \mathcal{M}_{j^{'}} - J\bar{g} \right)$ equals zero. Thus, the entire cross-term vanishes.

Combining these results, we obtain:
\begin{equation}
    \bar{E} = \bar{A} + E_{ens} \implies E_{ens} = \bar{E} - \bar{A}.
         \tag{A.16}
\end{equation}
This completes the proof.
\end{proof}

% In this section, we extend the theoretical derivation to the weighted ensemble scenario, which directly supports our \textit{LLM-PeerReview-W}.

\begin{theorem}[Error-Ambiguity Decomposition for the Scoring Phase of LLM-PeerReview-W]
Consider a scoring task with a target function $f$. 
Let $\{\mathcal{M}_1, \mathcal{M}_2, \dots, \mathcal{M}_J\}$ be a set of $J$ individual models
 with corresponding non-negative weights $\{w_1, w_2, \dots, w_J\}$ such that $\sum_{j^{'}=1}^J w_{j^{'}} = 1$. The weighted ensemble output $H$ is defined as:
\begin{equation}
    H(\mathbf{r}) = \sum_{j^{'}=1}^J w_{j^{'}} \mathcal{M}_{j^{'}}(\mathbf{r}).
         \tag{A.17}
\end{equation}
The relationship between the ensemble error $E_{ens}$, the weighted average individual error $\bar{E}$, and the weighted ambiguity $\bar{A}$ is given by:
\begin{equation}
    E_{ens} = \bar{E} - \bar{A},
         \tag{A.18}
\end{equation}
where $\bar{E} = \sum_{j^{'}=1}^J w_{j^{'}} (\mathcal{M}_{j^{'}} - f)^2$ and $\bar{A} = \sum_{j^{'}=1}^J w_{j^{'}} (\mathcal{M}_{j^{'}} - H)^2$.
\end{theorem}

\begin{proof}
Starting from the definition of the weighted average individual generalization error $\bar{E}$:
\begin{equation}
    \bar{E} = \sum_{j^{'}=1}^J w_{j^{'}} (\mathcal{M}_{j^{'}} - f)^2.
         \tag{A.19}
\end{equation}
By introducing the weighted ensemble term $H$ into the expression $(\mathcal{M}_{j^{'}} - f)$, we have:
\begin{equation}
    \bar{E} = \sum_{j^{'}=1}^J w_{j^{'}} \left[ (\mathcal{M}_{j^{'}} - H) + (H - f) \right]^2.
    \tag{A.20}
\end{equation}
Expanding the quadratic term using $(a+b)^2 = a^2 + b^2 + 2ab$:
\begin{equation}
    \bar{E} = \sum_{j^{'}=1}^J w_{j^{'}} (\mathcal{M}_{j^{'}} - H)^2 + \sum_{j^{'}=1}^J w_{j^{'}} (H - f)^2 + 2 \sum_{j^{'}=1}^J w_{j^{'}} (\mathcal{M}_{j^{'}} - H)(H - f).
     \label{eq: main eq for the theoretical analysisa on weighted averging method}
     \tag{A.21}
\end{equation}·
% \begin{equation}
% \begin{aligned}
%     \bar{E} &= \sum_{j^{'}=1}^J w_{j^{'}} (\mathcal{M}_{j^{'}} - H)^2 + \sum_{j^{'}=1}^J w_{j^{'}} (H - f)^2 \\
%     &\quad + 2 \sum_{j^{'}=1}^J w_{j^{'}} (\mathcal{M}_{j^{'}} - H)(H - f).
% \end{aligned}
% \label{eq: main eq for the theoretical analysisa on weighted averging method}
% \tag{A.21}
% \end{equation}
We analyze each of the three terms individually in the following.

    The first term in Equation~\ref{eq: main eq for the theoretical analysisa on weighted averging method} is the weighted average ambiguity of the individual models:
    \begin{equation*}
        \sum_{j^{'}=1}^J w_{j^{'}} (\mathcal{M}_{j^{'}} - H)^2 = \bar{A}.
         \tag{A.22}
    \end{equation*}
The second term in Equation~\ref{eq: main eq for the theoretical analysisa on weighted averging method} is the error of the weighted ensemble. Since $(H - f)^2$ is constant with respect to the summation index $j^{'}$:
    \begin{equation*}
        \sum_{j^{'}=1}^J w_{j^{'}} (H - f)^2 = (H - f)^2 \sum_{j^{'}=1}^J w_{j^{'}} = (H - f)^2 \cdot 1 = E_{ens}.
         \tag{A.23}
    \end{equation*}
 For the third term (the cross-term) in Equation~\ref{eq: main eq for the theoretical analysisa on weighted averging method} , since $(H - f)$ is independent of $j^{'}$, we can pull it out of the summation:
    \begin{equation*}
        2(H - f) \sum_{j^{'}=1}^J w_{j^{'}} (\mathcal{M}_{j^{'}} - H) = 2(H - f) \left( \sum_{j^{'}=1}^J w_{j^{'}} \mathcal{M}_{j^{'}} - H \sum_{j^{'}=1}^J w_{j^{'}} \right).
         \tag{A.24}
    \end{equation*}
% \begin{equation*}
% \begin{aligned}
%     &2(H - f) \sum_{j^{'}=1}^J w_{j^{'}} (\mathcal{M}_{j^{'}} - H) \\
%     &= 2(H - f) \left( \sum_{j^{'}=1}^J w_{j^{'}} \mathcal{M}_{j^{'}} - H \sum_{j^{'}=1}^J w_{j^{'}} \right).
% \end{aligned}
% \tag{A.24}
% \end{equation*}
    Given that $H = \sum_{j^{'}=1}^J w_{j^{'}} \mathcal{M}_{j^{'}}$ and $\sum_{j^{'}=1}^J w_{j^{'}} = 1$, the term $\left( H - H \cdot 1 \right)$ equals zero. Thus, the entire cross-term vanishes.

Combining these results, we obtain the identity:
\begin{equation}
    \bar{E} = \bar{A} + E_{ens} \implies E_{ens} = \bar{E} - \bar{A}.
     \tag{A.25}
\end{equation}
This completes the proof.
\end{proof}

% \subsection{Summary of Notations Used in LLM-PeerReview and the Algorithm}
\subsection{Notation and Algorithm Overview}
\label{sec: Summary of notations used in LLM-PeerReview}

We provide the mathematical notations used in LLM-PeerReview in Table \ref{table: LLM-PeerReview Symbols}.
Also, we provide the overall procedure for our proposed LLM-PeerReview in Algorithm~\ref{alg:algorithm}.

\refstepcounter{algorithm}
\begin{center}
\begin{tcolorbox}[
    width=0.88\textwidth,
    colback=algoback,
    colframe=algoframeline,
    arc=0pt,
    boxrule=0.6pt,
    left=6pt, right=6pt, top=4pt, bottom=4pt,
    title={\textbf{Algorithm \thealgorithm:} \textsc{LLM-PeerReview}},
    fonttitle=\small\bfseries,
    coltitle=black,
    colbacktitle=white,
    toprule at break=0pt,
    bottomrule at break=0pt
]
\label{alg:algorithm}
\footnotesize
\textbf{Input}:
Data $\mathcal{D}=\{\mathbf{x}^{(i)}, \mathbf{R}^{(i)}\}_{i=1}^I$, where for each query $\mathbf{x}^{(i)}$, we have responses $\mathbf{R}^{(i)}=\{\mathbf{r}^{(i,j)}\mid 1 \le j \le J \}$ from heterogeneous LLMs $\left\{\mathcal{M}_j\right\}_{j=1}^J$
\\
\textbf{Output}: results $\{ \mathbf{r}_{\text{ensemble}}^{(i)} \}_{i=1}^I$
\begin{algorithmic}[1]
\footnotesize
\STATE \textcolor{DeepBlue}{ \textit{\# \textbf{1) Scoring:}}}
\STATE Each LLM $\mathcal{M}_{j^{\prime}}$ acts as a judge and assigns a score $y^{(i,j;j^{\prime})}$ to each response $\mathbf{r}^{(i,j)}$
\vspace{3pt}
\STATE \textcolor{SoftBrown}{\textit{\# \textbf{2) Reasoning:}}}
\IF{default base variant (LLM-PeerReview)}
    \STATE Calculate each $Score(\mathbf{r}^{(i,j)})$ by using Eq.~10.
    \vspace{2pt}
\ELSIF{weighted variant (LLM-PeerReview-W)}
    \STATE \textcolor{gray}{\textit{\# 2.1) Reasoning the posterior probabilities $q(t^{(i,j)})$ over score categories for each response:}}
    \STATE Initialize posterior $\{q(t^{(i,j)})\mid 1 \le i \le I, 1 \le j \le J \}$ by averaging scores on $\textbf{r}^{(i,j)}$
    \WHILE{not converge}
        \STATE Update $\boldsymbol{\alpha}=\{\alpha^{(k)}\mid 1 \le k \le K \}$ by using Eq.~8
        \STATE Update $\{\boldsymbol{\Pi}^{(j^{\prime})}\}_{j^{\prime}=1}^J$ by using Eq.~9
        \STATE Update the posterior $\{q(t^{(i,j)})\mid 1 \le i \le I, 1 \le j \le J \}$ by using Eq.~7
    \ENDWHILE
    \STATE \textcolor{gray}{\textit{\# 2.2) Obtain the final score value for each response:}}
    \STATE Obtain the final score value $Score(\mathbf{r}^{(i,j)})$ for each individual response $\mathbf{r}^{(i, j)}$ by using Eq.~10
\ENDIF
\vspace{3pt}
\STATE \textcolor{DarkGreen}{\textit{\# \textbf{3) Selecting the best:}}}
\STATE Obtain the final results $\{ \mathbf{r}_{\text{ensemble}}^{(i)} \}_{i=1}^I$ by using Eq.~11
\end{algorithmic}
\end{tcolorbox}
\end{center}

\begin{table*}[t] % 建议用 [t] 放在页面顶部，[h!] 在跨栏表中容易失效
    \centering
    \small % 替代 resizebox，保持字体与正文比例协调
    \begin{tabular}{ll}
     \textbf{Symbol}& \textbf{Description}  \\ \toprule
   % \multirow{4}{*}{\makecell{Single\\LLM}} 

         $\{\mathbf{x}^{(i)}\}_{i=1}^I$ & Set of input queries $\{\mathbf{x}^{(i)}\}_{i=1}^I$ indexed from $1$ to $I$\\
    $i$ & Query index\\

          $\{\mathcal{M}_j\}_{j=1}^J$ & Set of large language models indexed from $1$ to $J$\\
             $j$ & Index of different LLM for response generation\\
          $\mathbf{r}^{(i,j)}$ & Response generated by LLM $\mathcal{M}_j$ for query $\mathbf{x}^{(i)}$
        \\
            $\mathbf{R}^{(i)}$ & Collection of responses $\mathbf{R}^{(i)}=[\mathbf{r}^{(i,1)}, \ldots, \mathbf{r}^{(i,J)}]$ from all LLMs for query $\mathbf{x}^{(i)}$    \\
              $\mathbf{y}^{(i)}$ & Latent truth/reference response for query $\mathbf{x}^{(i)}$ \\
               $y^{(i,j;j^{\prime})}$ & Judging score assigned by LLM indexed by $j^{'}$ to response  $\mathbf{r}^{(i,j)}$ \\
            $\{y^{(i,j;j^{\prime})}\}_{j^{\prime}=1}^J$ & Judging scores assigned by all $J$ LLMs to response  $\mathbf{r}^{(i,j)}$ \\
             $j^{'}$ & Index of different LLM for score generation in the Scoring phase\\
              $Score(\mathbf{r}^{(i,j)})$ & Final ensemble score for  response $\mathbf{r}^{(i,j)}$ after the  Reasoning phase \\
                $t^{(i,j)}$ & Latent true score category for response $\mathbf{r}^{(i,j)}$\\
                $\alpha$ &  Prior distribution for the score category $t^{(i,j)}$ 
        
                \\
                  $K$ &  Total number of score levels (categories)\\
        $\boldsymbol{\Pi}^{(j^{\prime})}$ &  Transition matrix for judge  LLM indexed by $j^{'}$      \\
          $\pi_{m n}^{(j^{\prime})}$ & Element of $\boldsymbol{\Pi}^{(j^{\prime})}$ (probability of judge $j^{\prime}$ scoring $n$ given true $m$)       \\
                 $s_k$ & Numerical value corresponding to the $k$-th score level \\
                     $\mathbf{r}_{\text {ensemble}}^{(i)}$ & Final ensemble result for  query $\mathbf{x}^{(i)}$  \\               
  
         \bottomrule
    \end{tabular}
    % \vspace{6pt}

     % \caption*{Table A.1: LLM-PeerReview Symbols.}
     \caption{LLM-PeerReview Symbols.}
 \label{table: LLM-PeerReview Symbols}
\end{table*}

\subsection{More Experiment Setup}
\label{sec: More Experiment Setup}

% \begin{table*}[h!]
%     \centering
%     \resizebox{0.62\textwidth}{!}{
%     \begin{tabular}{ll}
%          \textbf{Dataset }& \textbf{Huggingface URL}  \\ \toprule
%          TriviaQA & \url{https://huggingface.co/datasets/mandarjoshi/trivia_qa}\\
%          GSM8K & \url{https://huggingface.co/datasets/openai/GSM8K} \\ 
%          MATH & \url{https://huggingface.co/datasets/EleutherAI/hendrycks_math} \\ 
%          AlpacaEval & \url{https://huggingface.co/datasets/tatsu-lab/alpaca_eval}\\ \bottomrule
%     \end{tabular}
%     }
%     \caption*{Table A.1: The datasets used in our experiments.}
%     \label{table: appendix datasets}
% \end{table*}

\begin{table*}[h!]
    \centering
     \resizebox{0.73\textwidth}{!}{
    \begin{tabular}{lll}
      \textbf{Scale}&     \textbf{Model}& \textbf{Huggingface URL}  \\ \toprule
   % \multirow{4}{*}{\makecell{Single\\LLM}} 
   \multirow{4}{*}{7B}
    &
         \texttt{Llama-3.1-8B-Instruct} & \url{https://huggingface.co/meta-llama/Llama-3.1-8B-Instruct}\\
          &\texttt{Mistral-7B-Instruct} & \url{https://huggingface.co/mistralai/Mistral-7B-Instruct-v0.3}\\
         & \texttt{Qwen2-7B-Instruct} & \url{https://huggingface.co/Qwen/Qwen2-7B-Instruct}\\
         & \texttt{Qwen2.5-7B-Instruct} & \url{https://huggingface.co/Qwen/Qwen2.5-7B-Instruct}\\
          % \midrule
          \addlinespace[1.5ex] % 这里的 [1ex] 可以根据需要调整高度，不加括号则是默认间距
       \multirow{4}{*}{13B}  & \texttt{Qwen2.5-14B-Instruct} & \url{https://huggingface.co/Qwen/Qwen2.5-14B-Instruct}\\
         &  \texttt{DeepSeek-R1-Distill-Qwen-14B} & \url{https://huggingface.co/deepseek-ai/DeepSeek-R1-Distill-Qwen-14B}\\
         &    \texttt{Qwen2.5-14B-Instruct-1M} & \url{https://huggingface.co/Qwen/Qwen2.5-14B-Instruct-1M}\\
       &           \texttt{Phi-3-medium-4k-instruct}  & \url{https://huggingface.co/microsoft/Phi-3-medium-4k-instruct}\\        
         \bottomrule
    \end{tabular}
    }
    % \vspace{6pt}
    % \caption*{Table A.2: The models used in our experiments and their corresponding URLs.}
     \caption{The models used in our experiments and their corresponding URLs.}
    \label{table: The models used in our experiments and their corresponding URLs}
\end{table*}

Beyond the setup described in Section~3 of the main text, we use the authors' released implementations and the recommended hyperparameter settings for GaC, SAFE-GaC, SAFE-UniTE, CoRE-GaC, CoRE-UniTE, and MAD. No additional hyperparameter tuning is performed. For MAD, we use two debate rounds, following the configuration used and recommended in the original paper.
The models used in the experiments are shown in Table
 \ref{table: The models used in our experiments and their corresponding URLs}.
In the following, we provide further descriptions of the four datasets used.
As mentioned in Section~3 of the main text, the Appendix provides the code for LLM-PeerReview and all evaluated baselines on the datasets considered. We will publicly release this project, including all code
needed to reproduce the main experiments in Table 1.

\textbf{TriviaQA~\cite{dubois2023alpacafarm}.}
The TriviaQA dataset consists of 950K question-answer pairs from 662K documents on Wikipedia and the web. It is designed for complex question answering, where answers cannot always be directly obtained through span-based methods. The dataset includes both human-verified and machine-generated pairs.
We utilize the dataset from the Smoothie~\cite{guha2024smoothie}.
% containing 1000 randomly selected samples from the original dataset.

\textbf{GSM8K~\cite{chen2021evaluating}.}
GSM8K (Grade School Math 8K) is a dataset of 8.5K math word problems designed for grade school students, with 7.5K training and 1K test problems.  Each problem requires 2 to 8 steps to solve, focusing on basic arithmetic operations.  The natural language solutions allow for evaluating models' reasoning abilities.  
Given the high resource cost associated with using the full test set, we adopt existing approaches and utilize the publicly available dataset in Hu et al. \citeyearpar{hu2024language}, which consist of 400 randomly selected samples from the original full test set.

\textbf{MATH~\cite{hendrycks2021measuring}.}
The MATH dataset consists of 12,500 challenging, competition-level math problems across topics like algebra, geometry, probability, and number theory. Each problem is accompanied by a detailed, step-by-step solution, making it ideal for training models to not only find answers but also generate reasoning and logical explanations. This dataset is highly valuable for advancing and evaluating models in solving complex mathematical problems. 
% Also, given the high resource cost associated with using the full test set, 
We adopt existing approaches and utilize the publicly available dataset in Hu et al. \citeyearpar{hu2024language}, which consist of 400 randomly selected samples from the original full test set.

\textbf{AlpacaEval~\cite{dubois2023alpacafarm}.}
AlpacaEval consists of 805 instructions~\cite{hu2024language}, including 252 from the self-instruct test set~\cite{wang2022self}, 188 from the Open Assistant (OASST) test set, 129 from Anthropic’s helpful test set~\cite{zhou2023lima}, 80 from the Vicuna test set~\cite{chiang2023vicuna}, and 156 from the Koala test set~\cite{vu2023koala}.

\subsection{More Experimental Results}
\label{sec: More Experimental Results}

\definecolor{LightPink}{rgb}{0.97,0.92,0.92}

\begin{table*}[!h]
\centering
\begin{minipage}[t]{0.78\textwidth}
\begin{center}
\resizebox{0.99\linewidth}{!}{
\begin{tabular}{lccccc}
\toprule
\textbf{Method} &
\textbf{TriviaQA}$\uparrow$ &
\textbf{GSM8K}$\uparrow$ &
\textbf{MATH}$\uparrow$ &
\textbf{AlpacaEval}$\uparrow$ &
\textbf{Average}$\uparrow$ \\
\midrule

\multicolumn{6}{l}{\textit{Single LLM}} \\
Llama-3.1-8B-Instruct & 75.3 & 79.3 & 52.3 & 7.3 & 53.5 \\
Mistral-7B-Instruct & 72.7 & 64.3 & 26.5 & 10.4 & 43.5 \\
Qwen2-7B-Instruct & 63.0 & 88.5 & 59.8 & 15.2 & 56.6 \\
Qwen2.5-7B-Instruct & 62.5 & 91.5 & 69.3 & 27.6 & 62.7 \\
% Qwen2.5-7B-Instruct & 62.5 & 91.5 & 69.3 & 27.6 & 62.7 \\
Theoretical average & 68.4 & 80.9 & 51.9 & 15.1 & 54.1 \\
\midrule

\multicolumn{6}{l}{\textit{LLM Ensemble}} \\
Random \scriptsize{\cite{guha2024smoothie}}
& 68.4 \scriptsize{$\pm$ 0.3}
& 81.2 \scriptsize{$\pm$ 1.2}
& 52.2 \scriptsize{$\pm$ 1.1}
& 15.2 \scriptsize{$\pm$ 0.6}
& 54.2 \\

Smoothie-Global \scriptsize{\cite{guha2024smoothie}}
& 63.0 & 91.5 & 59.8 & 27.6 & 60.5 \\

Smoothie-Local \scriptsize{\cite{guha2024smoothie}}
& 73.6 & 85.5 & 61.8 & 18.3 & 59.8 \\

Agent-Forest \scriptsize{\cite{li2024more}}
& 70.5 & 86.8 & 61.0 & 22.1 & 60.1 \\

GaC \scriptsize{\cite{yu2024breaking}}
& 71.5 & 91.8 & 54.0 & 23.6 & 60.2 \\

SAFE-GaC \scriptsize{\cite{yun2026when}}
& 63.2 & 90.5 & 41.3 & 20.1 &  53.8\\

SAFE-UniTE \scriptsize{\cite{yun2026when}}
& 63.3 & 91.5 & 48.5 & 16.5 & 55.0 \\

CoRE-GaC \scriptsize{\cite{zeng2026harnessing}}
& 64.0 & 92.0 & 63.5 & 8.7 & 57.1 \\

CoRE-UniTE \scriptsize{\cite{zeng2026harnessing}}
& 64.0 & 92.0 & 63.3 & 8.7 & 57.0 \\

MAD \scriptsize{\cite{liang2024encouraging}}
& - & 91.8 & 47.3 & - & - \\

\rowcolor[rgb]{0.867,0.922,0.969}
LLM-PeerReview
& 76.9 \scriptsize{$\pm$ 0.1}
& 92.7 \scriptsize{$\pm$ 0.3}
& 69.5 \scriptsize{$\pm$ 0.2}
& \textbf{30.4} \scriptsize{$\pm$ 0.1}
& 67.4 \\

\rowcolor[rgb]{0.867,0.922,0.969}
LLM-PeerReview-W
& \textbf{77.0} \scriptsize{$\pm$ 0.1}
& \textbf{93.0} \scriptsize{$\pm$ 0.2}
& \textbf{71.0} \scriptsize{$\pm$ 0.2}
& 30.2 \scriptsize{$\pm$ 0.1}
& \textbf{67.8} \\
\midrule
\midrule
\multicolumn{6}{l}{\textit{Our variants (flipped-triple)}} \\
Llama-3.1-8B-Selection
& 76.5 \scriptsize{$\pm$ 0.2}
& 90.8 \scriptsize{$\pm$ 0.6}
& 68.8 \scriptsize{$\pm$ 0.5}
& 29.6 \scriptsize{$\pm$ 0.3}
& 66.4 \\

Mistral-7B-Selection
& 75.6 \scriptsize{$\pm$ 0.3}
& 90.8 \scriptsize{$\pm$ 0.1}
& 66.4 \scriptsize{$\pm$ 0.3}
& 25.9 \scriptsize{$\pm$ 0.4}
& 64.7 \\

Qwen2-7B-Selection
& 74.2 \scriptsize{$\pm$ 0.2}
& 88.8 \scriptsize{$\pm$ 0.6}
& 61.7 \scriptsize{$\pm$ 0.7}
& 23.7 \scriptsize{$\pm$ 0.3}
& 62.1 \\

Qwen2.5-7B-Selection
& 75.5 \scriptsize{$\pm$ 0.2}
& 92.1 \scriptsize{$\pm$ 0.4}
& 66.2 \scriptsize{$\pm$ 0.6}
& 28.1 \scriptsize{$\pm$ 0.1}
& 65.5 \\
\midrule

\multicolumn{6}{l}{\textit{Our variants (single)}} \\

\rowcolor{LightPink}
Llama-3.1-8B-Selection
& 69.8 \scriptsize{$\pm$ 0.3}
& 83.7 \scriptsize{$\pm$ 1.3}
& 56.8 \scriptsize{$\pm$ 0.4}
& 21.5 \scriptsize{$\pm$ 0.6}
& 58.0 \\

\rowcolor{LightPink}
Mistral-7B-Selection
& 71.1 \scriptsize{$\pm$ 0.9}
& 82.9 \scriptsize{$\pm$ 0.5}
& 57.3 \scriptsize{$\pm$ 0.5}
& 18.6 \scriptsize{$\pm$ 0.1}
& 57.5 \\

\rowcolor{LightPink}
Qwen2-7B-Selection
& 70.9 \scriptsize{$\pm$ 0.5}
& 81.7 \scriptsize{$\pm$ 0.6}
& 53.4 \scriptsize{$\pm$ 0.2}
& 16.9 \scriptsize{$\pm$ 0.3}
& 55.7 \\

\rowcolor{LightPink}
Qwen2.5-7B-Selection
& 71.0 \scriptsize{$\pm$ 0.8}
& 83.2 \scriptsize{$\pm$ 0.9}
& 55.4 \scriptsize{$\pm$ 0.5}
& 23.8 \scriptsize{$\pm$ 0.2}
& 58.4 \\

\bottomrule
\end{tabular}
}
\end{center}
% \phantomsection
% \caption*{Table A.3: Main results (\%).}
\caption{Main results (\%). “-”: not applicable to the corresponding task.}
\label{table: Main results-appendix}
\end{minipage}
\end{table*}
% \end{center}
% \captionsetup{
%   width=\linewidth,
%   justification=justified,
%   singlelinecheck=false
% }
% \caption{Main results (\%).}
% \label{table: Main results-appendix}
% \end{minipage}
% \end{table*}

\begin{figure*}[h!]
\centering
\includegraphics[width=0.57\textwidth]{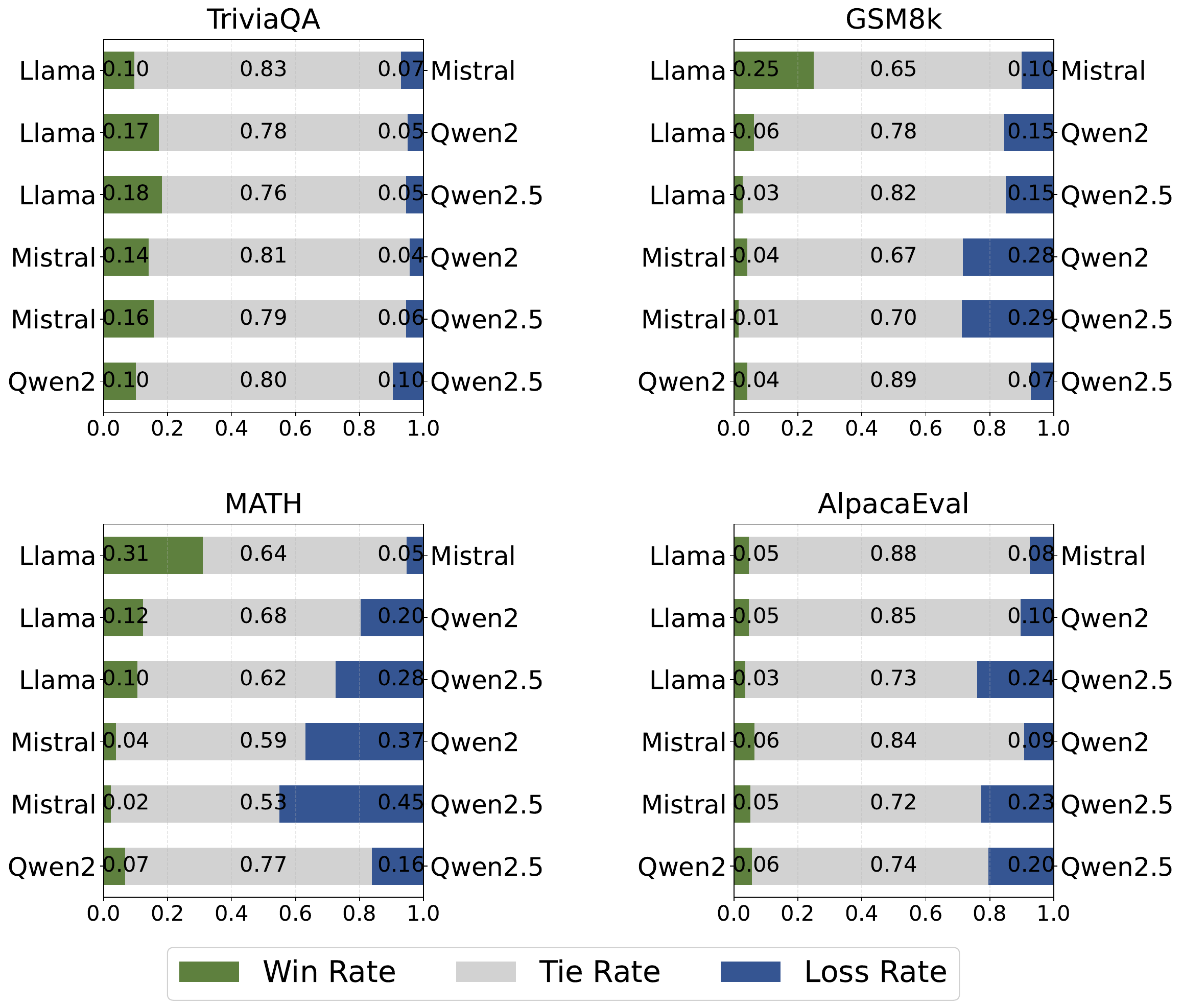} % Reduce the figure size so that it is slightly narrower than the column.
\phantomsection % <<< 在 \caption* 之前添加此命令
% \caption*{Figure A.1: Win-tie-loss charts for the 7B LLMs on four datasets.}
\caption{Win-tie-loss charts for the 7B LLMs on four datasets.}
\label{fig: Win-tie-loss charts of the 7B LLMs on four datasets}
\end{figure*}

\begin{figure*}[h!]
\centering
\includegraphics[width=0.8\textwidth]{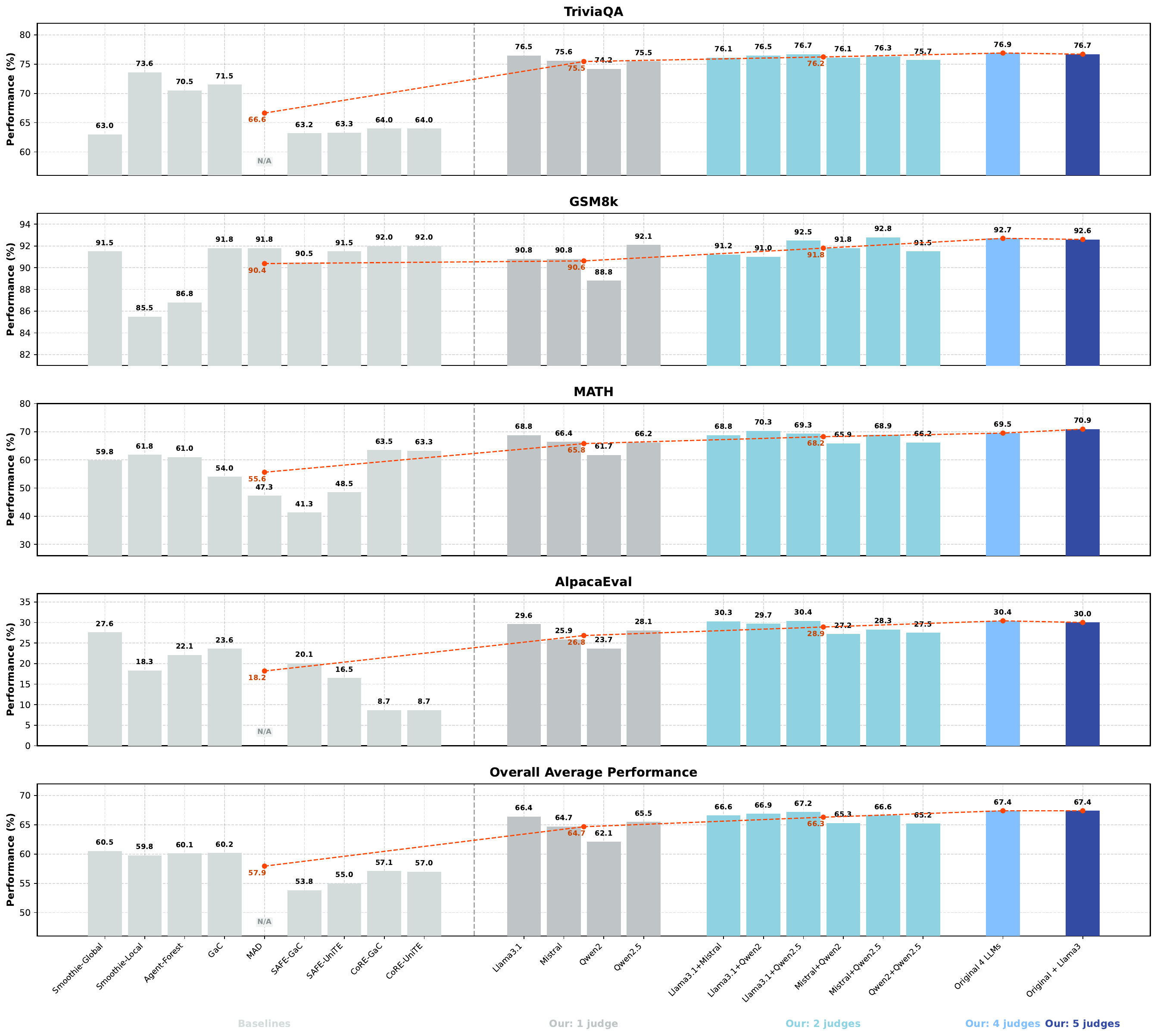} % Reduce the figure size so that it is slightly narrower than the column.
\phantomsection % <<< 在 \caption* 之前添加此命令
% \caption*{Figure A.2: Comparison of key LLM Ensemble baselines versus LLM-PeerReview variants with different numbers of judges.}
\caption{Comparison of LLM Ensemble baselines versus LLM-PeerReview variants with different numbers of judges.}
\label{fig: Comparison of key LLM Ensemble baselines versus LLM-PeerReview variants with different numbers of judges}
\end{figure*}

% \caption*{Figure A.1: Win-tie-loss charts on four datasets.}
% \label{fig: appendix win-tie-loss}
% \end{figure*}

\paragraph{(a) More variant results.}
By examining Table~\ref{table: Main results-appendix}, we observe the following:
(\textit{i}) The four variants of ``single'' obtained through the ``naive point-wise scoring'' method outperform the ``theoretical average'' across all four datasets. This suggests that these variant methods have a distinct advantage over random selection from model responses, implying they are useful;
(\textit{ii}) However, these four variants are still insufficiently useful when compared to the flipped-triple variant methods in 
Table ~\ref{table: Main results-appendix}, 
indicating a performance disadvantage.
By contrasting the performance of ``variants (single)'' and ``variants (flipped-triple)'', it becomes clear that our proposed flipped-triple scoring strategy is indeed effective. This conclusion supports the assertion that the flipped-triple strategy, as designed in 
% Section~\ref{sec: Scoring}
Section~2.1
of the main paper, mitigates the bias introduced by the naive point-wise scoring method. 
Furthermore, it is precisely upon this superior ``flipped-triple'' scoring strategy that the performance of our two primary variants, LLM-PeerReview and LLM-PeerReview-W, exhibits its ultimate superiority.
In addition: 
(\textit{i}) 
MAD results on TriviaQA and AlpacaEval are unavailable because its final majority-vote aggregation assumes extractable answers, limiting its direct applicability to free-form generation tasks;
(\textit{ii}) Compared with Table~1 in the main paper, we additionally report AlpacaEval results for SAFE-GaC, SAFE-UniTE, CoRE-GaC, and CoRE-UniTE after resolving an API endpoint issue.

% \caption*{Figure A.1: Win-tie-loss charts for the 7B LLMs on four datasets.}
% \label{fig: Win-tie-loss charts of the 7B LLMs on four datasets}
% \end{figure*}

\paragraph{(b) Win-tie-loss charts.}
As a supplement to Figure 3 in the main text, we provide 
% Figure~\hyperref[fig: Win-tie-loss charts of the 7B LLMs on four datasets]{A.1} 
Figure~A.1 
here, showing the win-tie-loss charts for models across the four datasets.
(\textit{i}) Overall, on the TriviaQA dataset, Llama outperforms the other three models; on the GSM8K, MATH, and AlpacaEval datasets, Qwen2.5 consistently performs the best, although its advantage varies in strength across datasets;
(\textit{ii}) However, consistent with the analysis in Section 4.1 of the main text, observing the win-tie-loss charts across the four datasets reveals that, on a specific dataset, despite a model having a significant overall performance advantage, its performance on individual queries (i.e., samples) may not be better than a model with an overall suboptimal performance. 
This intuitive reality suggests that attempting LLM Ensemble makes a lot of sense, and a sufficiently sophisticated method could lead to even higher performance than the all individual models.

\paragraph{(c) Transition matric.}
% As a supplement to Figure~4 in the main text, we provide Figure \ref{fig: Win-tie-loss charts of the 7B LLMs on four datasets} here, showing the win-tie-loss charts for models across the four datasets.

% As a complement to Figure~4 in the main paper, Figure\ref{fig: Win-tie-loss charts of the 7B LLMs on four datasets}  presents the learned transition matrices (and corresponding corelation information) for all datasets. The results are consistent with the findings reported in Figure~4.

\begin{figure*}[t] 
	\centering  
	\subfigtopskip=-0pt 
	\subfigbottomskip=0pt 
	\subfigcapskip=-5pt

    \subfigure[TriviaQA]{
		\label{level.sub.3}
		\includegraphics[width=0.478\linewidth]{figures/matrix-triviaqa.pdf}}
        \hspace{16mm}
	\subfigure[TriviaQA]{
		\label{level.sub.4}
		\includegraphics[width=0.16\linewidth]{figures/correlation-qa.pdf}}
    
  	\subfigure[GSM8k]{
		\label{level.sub.3}
		\includegraphics[width=0.478\linewidth]{figures/matrix-gsm8k.pdf}}
        \hspace{16mm} 
	\subfigure[GSM8k]{
		\label{level.sub.4}
		\includegraphics[width=0.16\linewidth]{figures/correlation-gsm8k.pdf}}

  	\subfigure[MATH]{
		\label{level.sub.3}
		\includegraphics[width=0.478\linewidth]{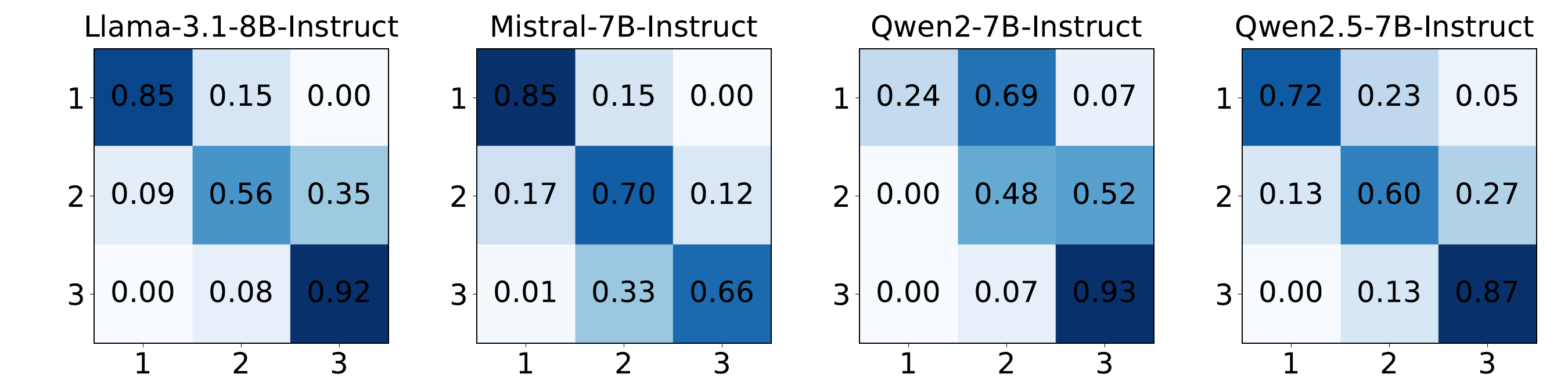}}
        \hspace{16mm}
	\subfigure[MATH]{
		\label{level.sub.4}		
        \includegraphics[width=0.16\linewidth]{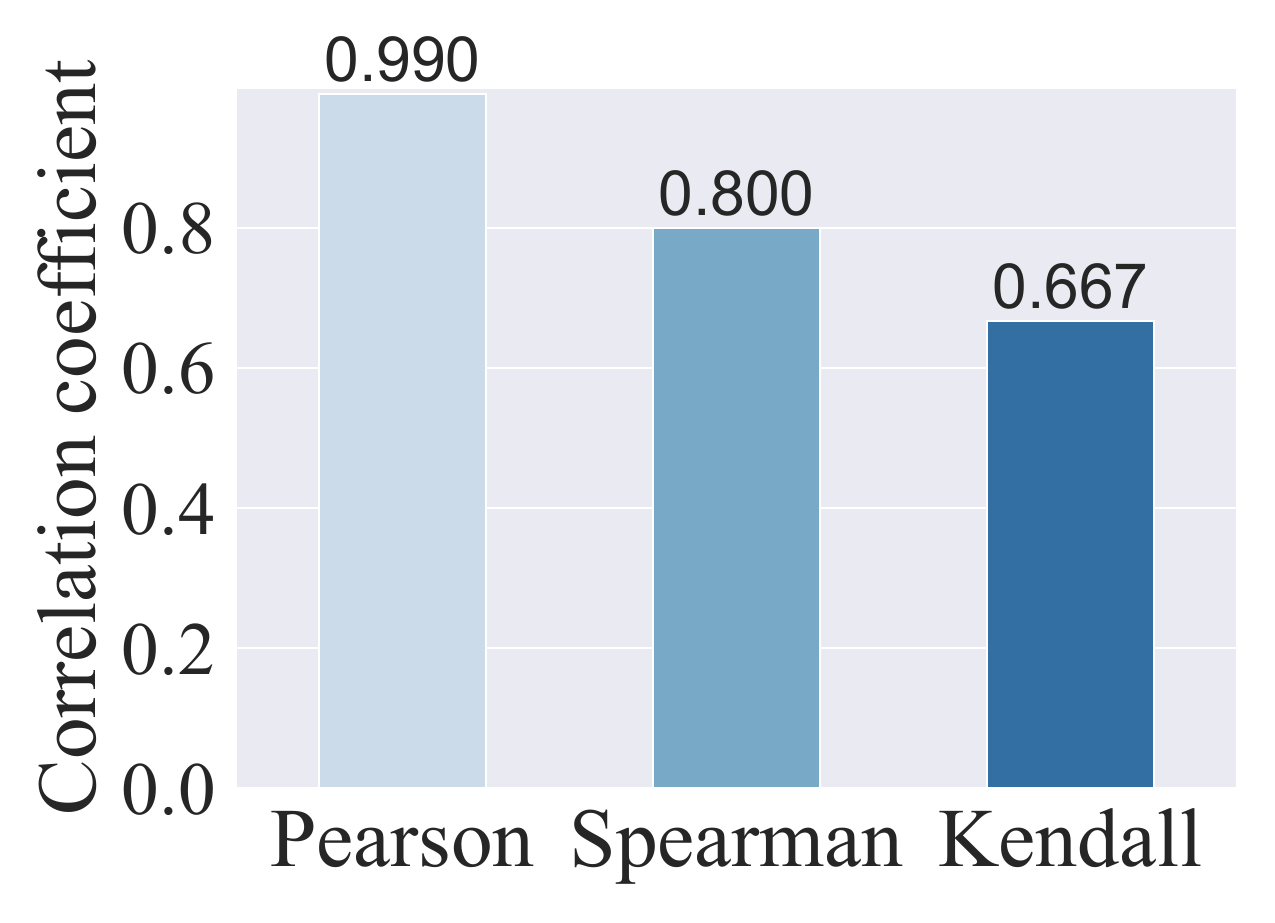}}

  	\subfigure[AlpacaEval]{
		\label{level.sub.3}
		\includegraphics[width=0.478\linewidth]{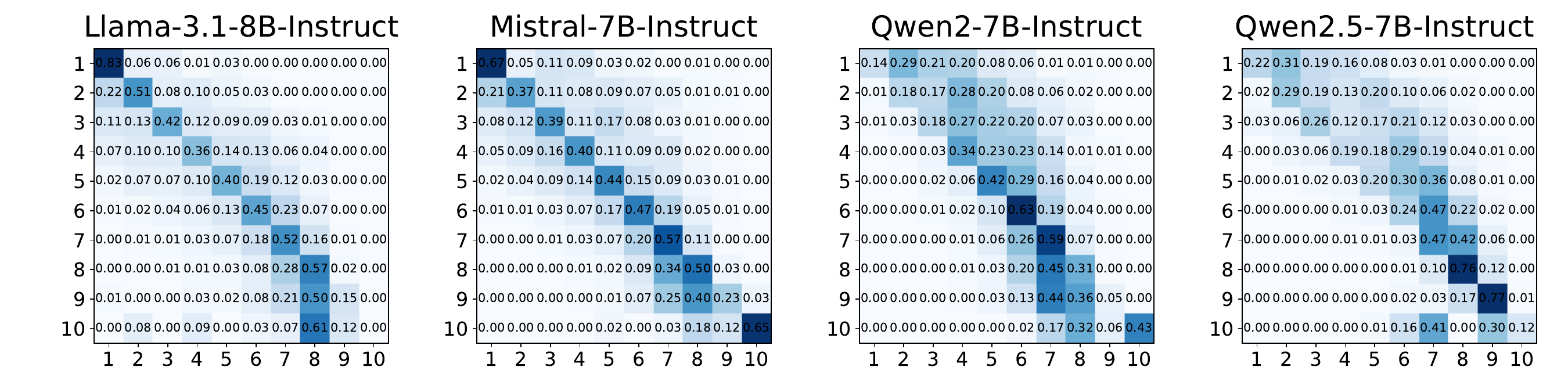}}
        \hspace{16mm}
	\subfigure[AlpacaEval]{
		\label{level.sub.4}
		\includegraphics[width=0.16\linewidth]{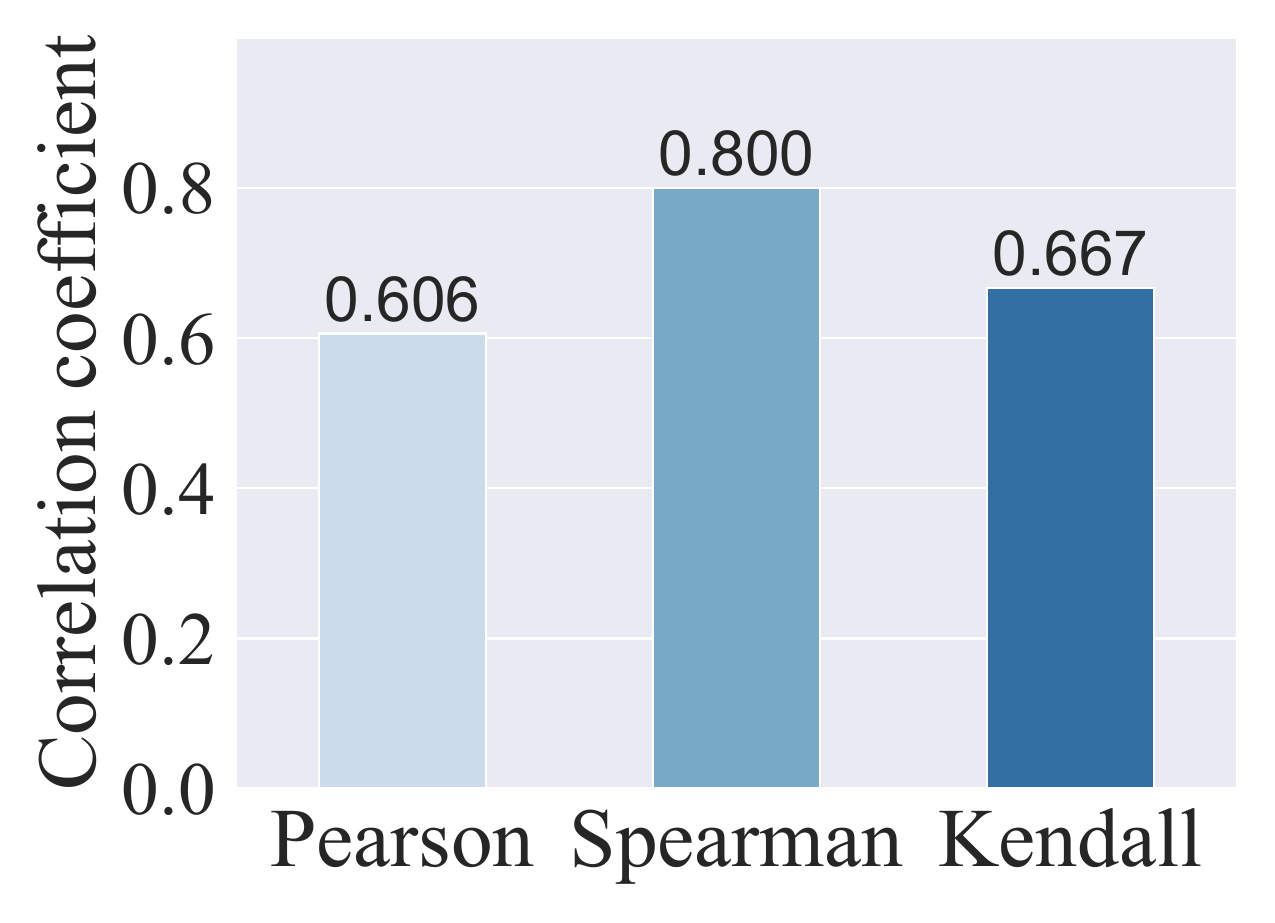}}

\caption{\textbf{Left:}  The transition matrix of each LLM estimated by LLM-PeerReview-W. \textbf{Right:} 
Correlation between matrix diagonal information of each LLM and its performance as a single judge (corresponding to ``our variants'' in Table 1 of the main text). 
For the first three datasets with ground-truth answers, diagonal information is represented by extreme values $(\pi_{1 1}^{(j^{\prime})} + \pi_{K K}^{(j^{\prime})})$.
for the instruction-following dataset AlpacaEval, the sum of all diagonal values is used.
} 
\label{fig: transition matrix and correlation-appendix}
\end{figure*}

As a complement to Figure~4 in the main paper, Figure~\ref{fig: transition matrix and correlation-appendix} presents the learned transition matrices and the corresponding correlation coefficients for all datasets. These results are consistent with the findings reported in Figure~4.
For the first three datasets with ground-truth answers, diagonal information is represented by extreme values $(\pi_{1 1}^{(j^{\prime})} + \pi_{K K}^{(j^{\prime})})$.
for the instruction-following dataset AlpacaEval, the sum of all diagonal values is used.

\paragraph{(d) Comparison of key LLM Ensemble baselines versus LLM-PeerReview variants with different numbers of judges.}
The experimental results are illustrated in Figure~\ref{fig: Comparison of key LLM Ensemble baselines versus LLM-PeerReview variants with different numbers of judges}. As a supplementary evaluation to Section~4.3 in the main paper, we have the following findings:
(\textit{i}) For LLM-PeerReview, the  average performance across four datasets (bottom sub-figure in Figure~\ref{fig: Comparison of key LLM Ensemble baselines versus LLM-PeerReview variants with different numbers of judges}) shows that increasing the number of judges generally improves performance (specifically, $our: 1\text{ judge} < 2\text{ judges} < 4\text{ judges} \approx 5\text{ judges}$). However, adding Llama-3-8B-Instruct~\footnote{Huggingface URL: \url{https://huggingface.co/meta-llama/Meta-Llama-3-8B-Instruct}.} as a fifth judge did not yield further gains, primarily because performance improvements become increasingly marginal as the ensemble size grows. Additionally, we found that Llama-3-8B-Instruct's individual average performance (62.1\%) is the lowest among the four candidate judges; for comparison, Llama-3.1-8B-Instruct achieved 66.4\%;
(\textit{ii}) Furthermore, the overall average performance indicates that LLM-PeerReview significantly outperforms all classic LLM Ensemble baselines,  across all judge configurations.

\begin{figure*}[h!]
\centering
\includegraphics[width=0.75\textwidth]{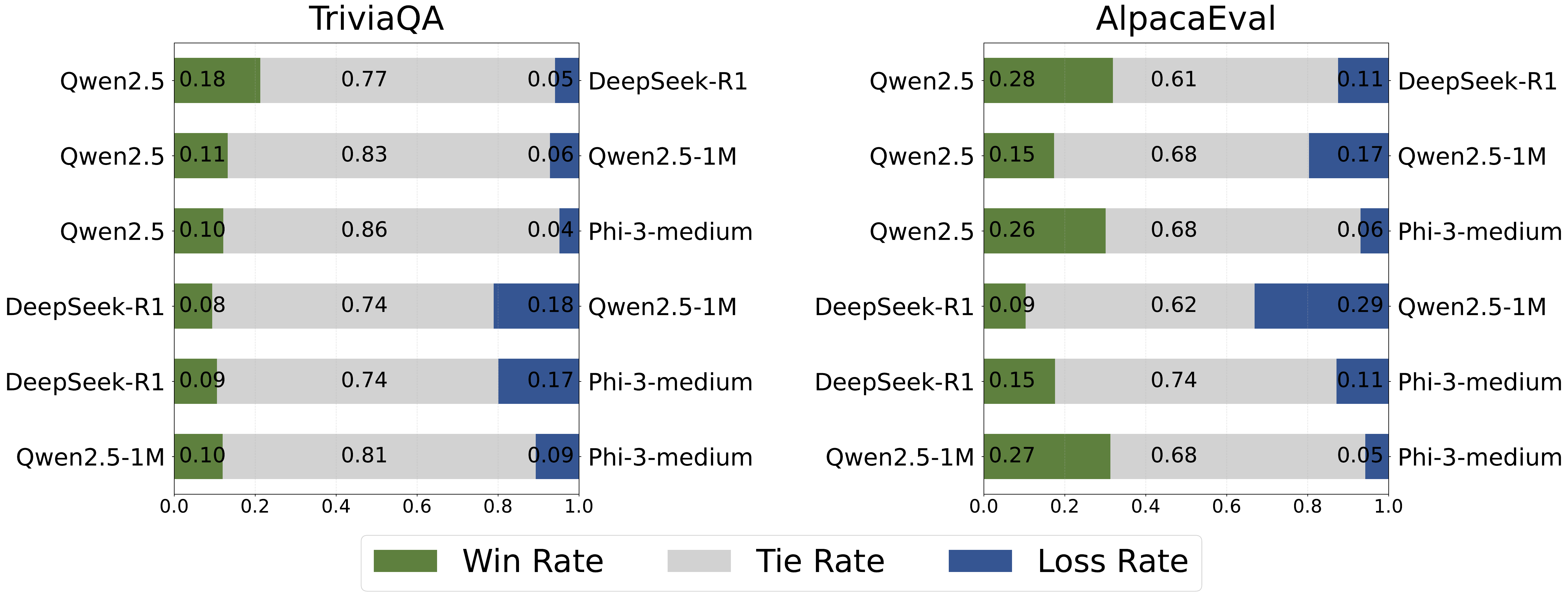} % Reduce the figure size so that it is slightly narrower than the column.
\phantomsection % <<< 在 \caption* 之前添加此命令
% \caption*{Figure A.3: Win-tie-loss charts for 13B-scale LLMs.}
\caption{Win-tie-loss charts for 13B-scale LLMs.}
\label{fig: Win-tie-loss charts for 13B-scale LLMs}
\end{figure*}

\paragraph{(e) More experiments on 13B-scale LLMs.}
In our main paper, we conducted experimental analysis using popular 7B-scale models. Here, we further investigate the performance of our method and  baselines on 13B-scale LLMs. 
The corresponding experimental results and win-tie-loss charts for the LLMs are presented in Table~\ref{table: Main results for 13B-scale LLMs} and Figure~\ref{fig: Win-tie-loss charts for 13B-scale LLMs}.
Our primary findings are as follows:
(\textit{i}) Similar to the 7B-scale results in Table 1 of the main paper, we obtained consistent findings for the 13B-scale models. The proposed methods, LLM-PeerReview and LLM-PeerReview-W, demonstrate significant advantages over both individual LLMs and established LLM Ensemble baselines;
(\textit{ii}) For the win-tie-loss charts, we observe a pattern similar to Figure~\ref{fig: Win-tie-loss charts of the 7B LLMs on four datasets}: different models exhibit varying performance levels, manifesting dynamic and diverse strengths and weaknesses across different datasets.

\begin{table*}[!h]
\centering
\begin{minipage}[t]{0.82\textwidth}
\begin{center}
\resizebox{0.63\linewidth}{!}{
\begin{tabular}{lcc  c}
\toprule
 \textbf{Method}  & 
\textbf{TriviaQA}$\uparrow$ & \textbf{AlpacaEval}$\uparrow$ & \textbf{Average}$\uparrow$    \\ 
     
\midrule
Qwen2.5-14B-Instruct            & 73.5  & 34.5  & 54.0  \\
DeepSeek-R1-Distill-Qwen-14B    & 59.5  & 17.5  & 38.5  \\
Qwen2.5-14B-Instruct-1M         & 69.5  & 36.5  & 53.0  \\
Phi-3-medium-4k-instruct        & 68.0  & 14.0  & 41.0  \\
 Theoretical average            & 67.6  & 25.6  & 46.6  \\
 
\midrule
 Random         & 67.7 \scriptsize{$\pm$ 0.6}   & 26.0 \scriptsize{$\pm$ 1.5}   & 46.9  \\
Smoothie-Global & 73.5  & 34.5  & 54.0 \\
Smoothie-Local  & 74.0  & 33.5  & 53.8 \\
Agent-Forest    & 73.0  & 40.5  & 56.8 \\
% GaC             & -     & -     & - \\
LLM-PeerReview  &\textbf{77.3} \scriptsize{$\pm$ 0.2}           & 46.5 \scriptsize{$\pm$ 0.0}                   & 61.9 \\
LLM-PeerReview-W& \textbf{77.3} \scriptsize{$\pm$ 0.2}  & \textbf{47.3} \scriptsize{$\pm$ 0.2} & \textbf{62.3} \\
 \bottomrule 
\end{tabular}
}
\end{center}
% \caption*{Table A.4: Main results for 13B-scale LLMs (\%).}
\caption{Main results for 13B-scale LLMs (\%).}
\label{table: Main results for 13B-scale LLMs}
\end{minipage}
\end{table*}

% \paragraph{(g) Case studies.}
% \paragraph{(f) More efficiency/cost analysis.}
\paragraph{(f) Further efficiency and cost analysis.}

\textit{Methodologically,} as discussed in the paragraph \textit{More efficient scoring} in Section~2 and in Section~2.4 of the main paper, the scoring stage supports two simple efficiency controls. First, it can use fewer LLM judges. Second, flipped-triple scoring can be sparsified for each judge by applying a sliding window over candidate responses, thereby avoiding exhaustive enumeration of all response triplets.

\textit{Empirically,} taken together, Table~A.3, Figure~A.2, and Table~2 of the main paper show that our method provides a substantially better performance--efficiency trade-off than MAD~\cite{liang2024encouraging} on the two mathematical reasoning datasets, GSM8K and MATH. Even with only two judges, our method achieves an average score of 68.2 on MATH, compared with 47.3 for MAD, yielding a gain of 20.9 percentage points while further reducing the scoring cost. Moreover, our method applies naturally to both exact-match and free-form generation tasks. In contrast, the final majority-vote aggregation used by MAD~\cite{liang2024encouraging} is not directly applicable to outputs from datasets such as TriviaQA and AlpacaEval without an additional answer-matching mechanism.

\begin{figure*}[t!]
\centering
\includegraphics[width=0.95\textwidth]{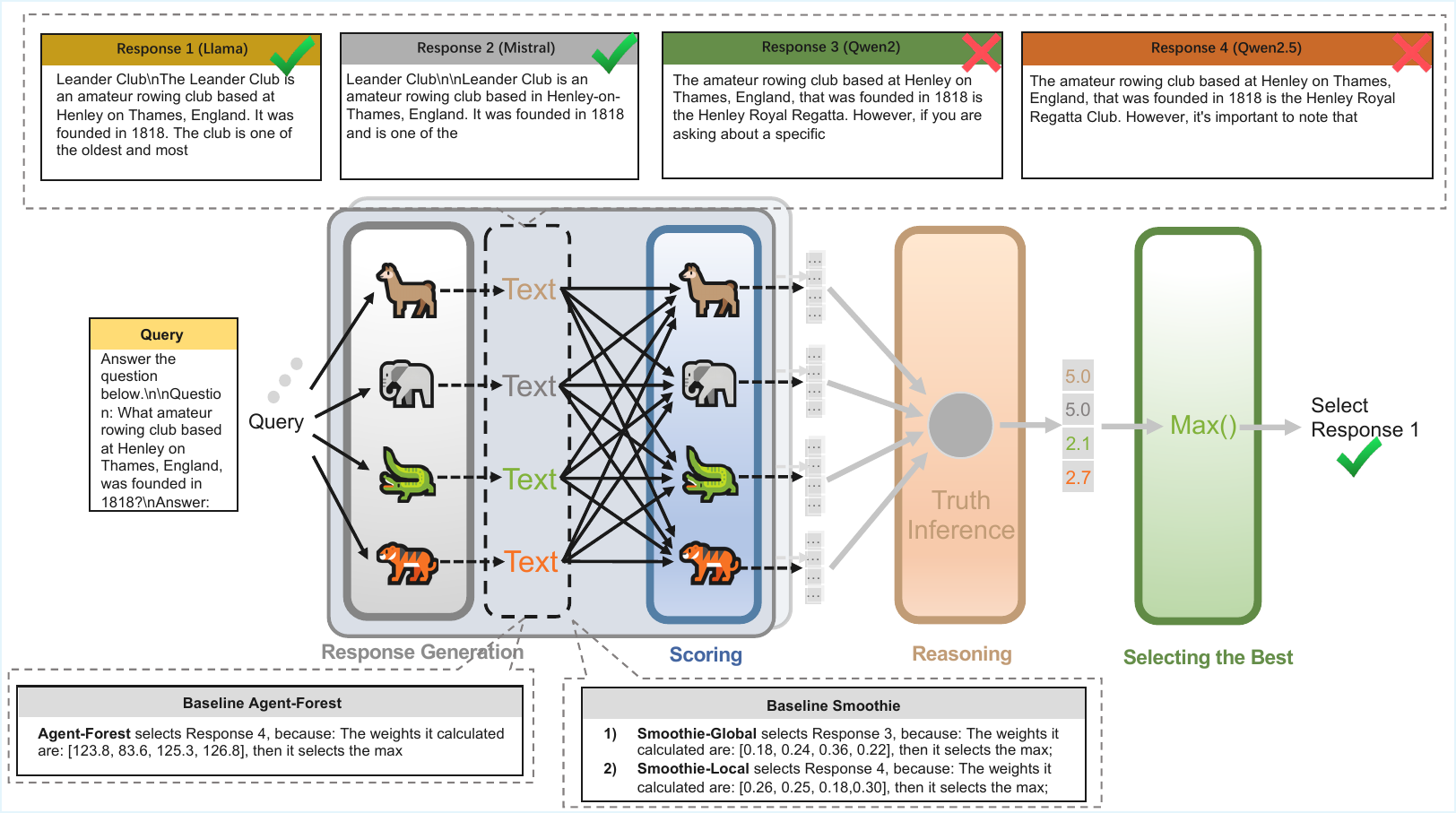} % Reduce the figure size so that it is slightly narrower than the column.
\caption{Case study on dataset TriviaQA.  \textbf{Analysis}: For this query, the correct answer is ``Leander Club''.
For each response generation, we followed the approach in ~\citet{guha2024smoothie} and applied truncation techniques under the same configuration.
(\textit{i})
For our method, the results of our variant, LLM-PeerReview, are shown in the Figure; also,  the selected Response after truth inference remains correct, with the four scalar values obtained as: [2.39, 2.39, 1.18, 1.18];
(\textit{ii}) For the baseline Agent-Forest, it focuses on ``which response has the highest cumulative BLEU similarity to all other responses''.
Since Responses 1 and 2 are correct, while Responses 3 and 4 are incorrect, the cumulative BLEU similarity of Response 4 with all others is the highest, so Agent-Forest incorrectly selects Response 4;
(\textit{iii}) For the baseline Smoothie, its two variants do not consider the four responses for the current query during the calculation. Instead, they either focus on identifying the best overall model or the responses from the most similar neighboring query. As a result, both variants of Smoothie also make errors.}
\label{fig: appendix case qa}
\end{figure*}

\begin{figure*}[t!]
\centering
\includegraphics[width=0.95\textwidth]{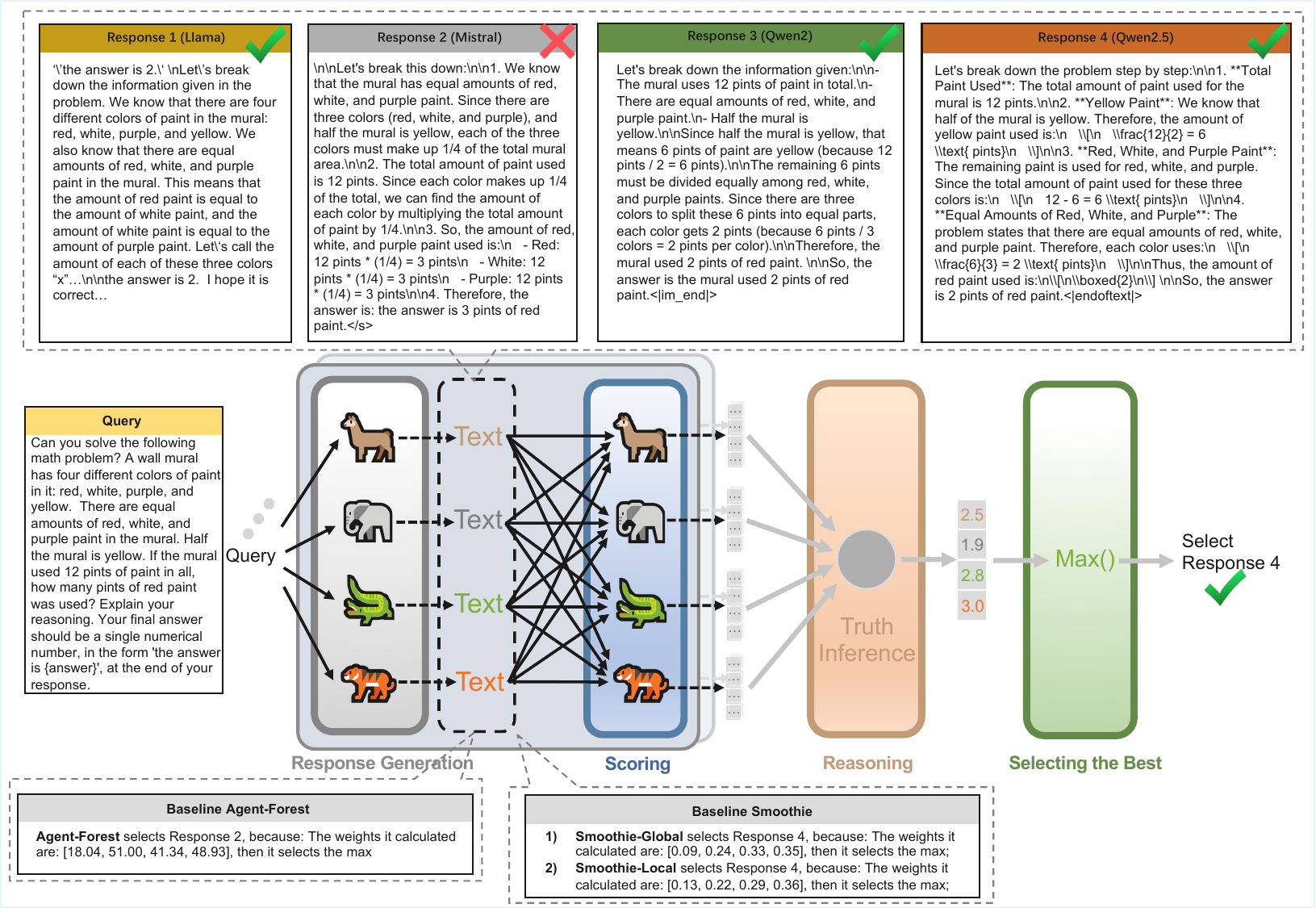} 
\caption{Case study on dataset GSM8K. 
The ellipses ``...'' in the Responses represent the omission of less important content.
\textbf{Analysis}: The correct answer for this query is ``2''.
(\textit{i}) For our method, the results of our variant, LLM-PeerReview, are shown in the Figure;
also,  the selected Response after truth inference remains correct, with the four scalar values obtained as: [1.12, 1.00, 2.01, 2.90];
(\textit{ii})
For the Agent-Forest method, it focuses on determining ``which response has the highest cumulative BLEU similarity to all other responses''.
Despite Response 2 not providing the correct final answer, Agent-Forest selects it because the weight (i.e., cumulative BLEU similarity score) is the highest. 
This is likely due to the high similarity between many words in the response compared to other responses. 
In this case, the simple analysis of ``which response has the highest cumulative BLEU similarity'' fails;
(\textit{iii})
For the Smoothie method, its two variants do not consider the four responses for the current query during the computation. 
Instead, they either focus on identifying the best overall model or consider the four responses of the most similar neighboring query. In contrast to the results shown in Figure \ref{fig: appendix case qa}, here both variants of Smoothie obtained the correct result, despite the computation process being flawed.}
\label{fig: appendix case gsm8k}
\end{figure*}

\begin{figure*}[t!]
\centering
\includegraphics[width=0.95\textwidth]{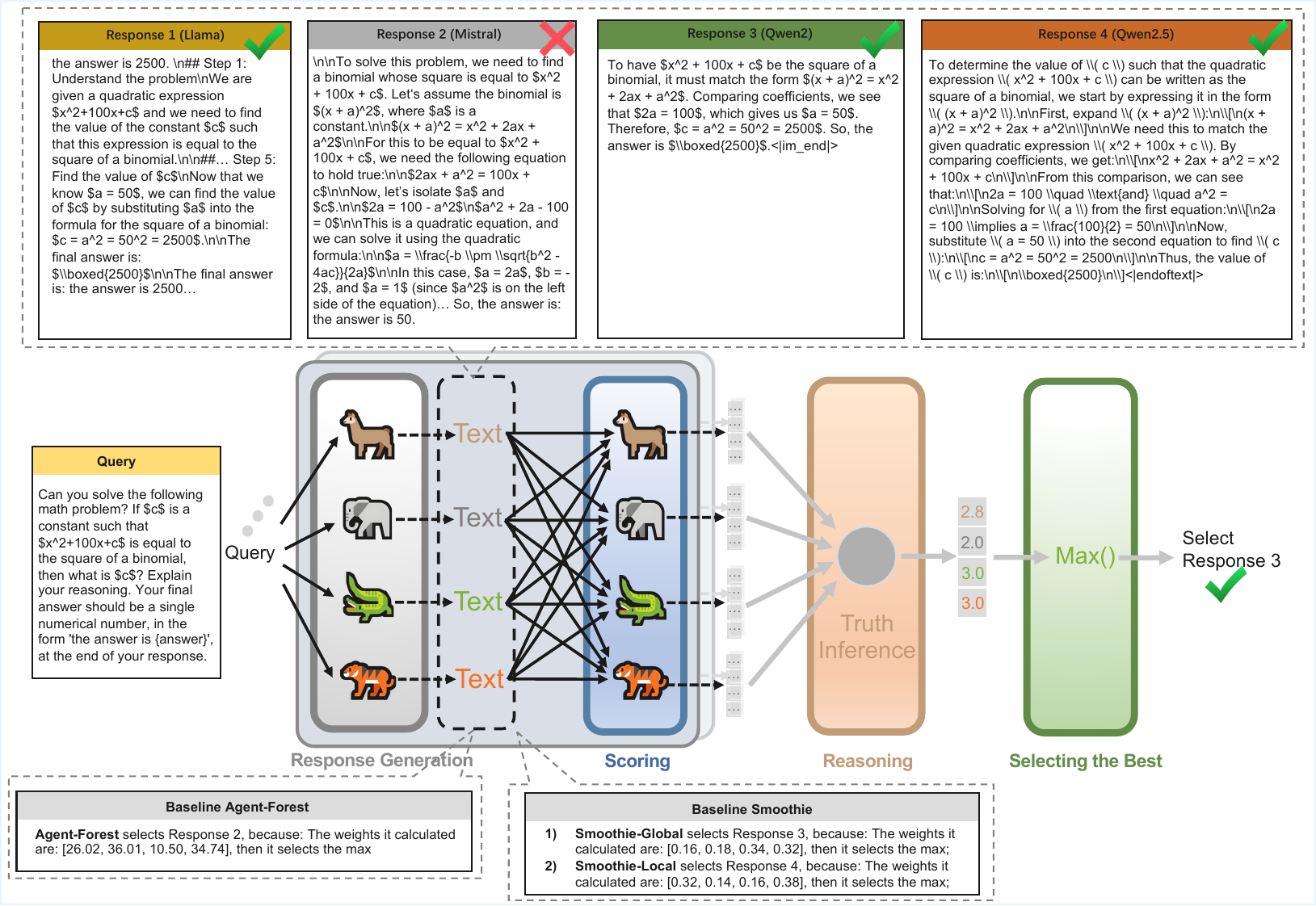}
\caption{Case study on dataset MATH.
For this query, the correct answer is “2500”.
The ellipses “...” in the Responses represent the omission of less important content.}
\label{fig: appendix case math}
\end{figure*}

\begin{figure*}[t!]
\centering
\includegraphics[width=0.95\textwidth]{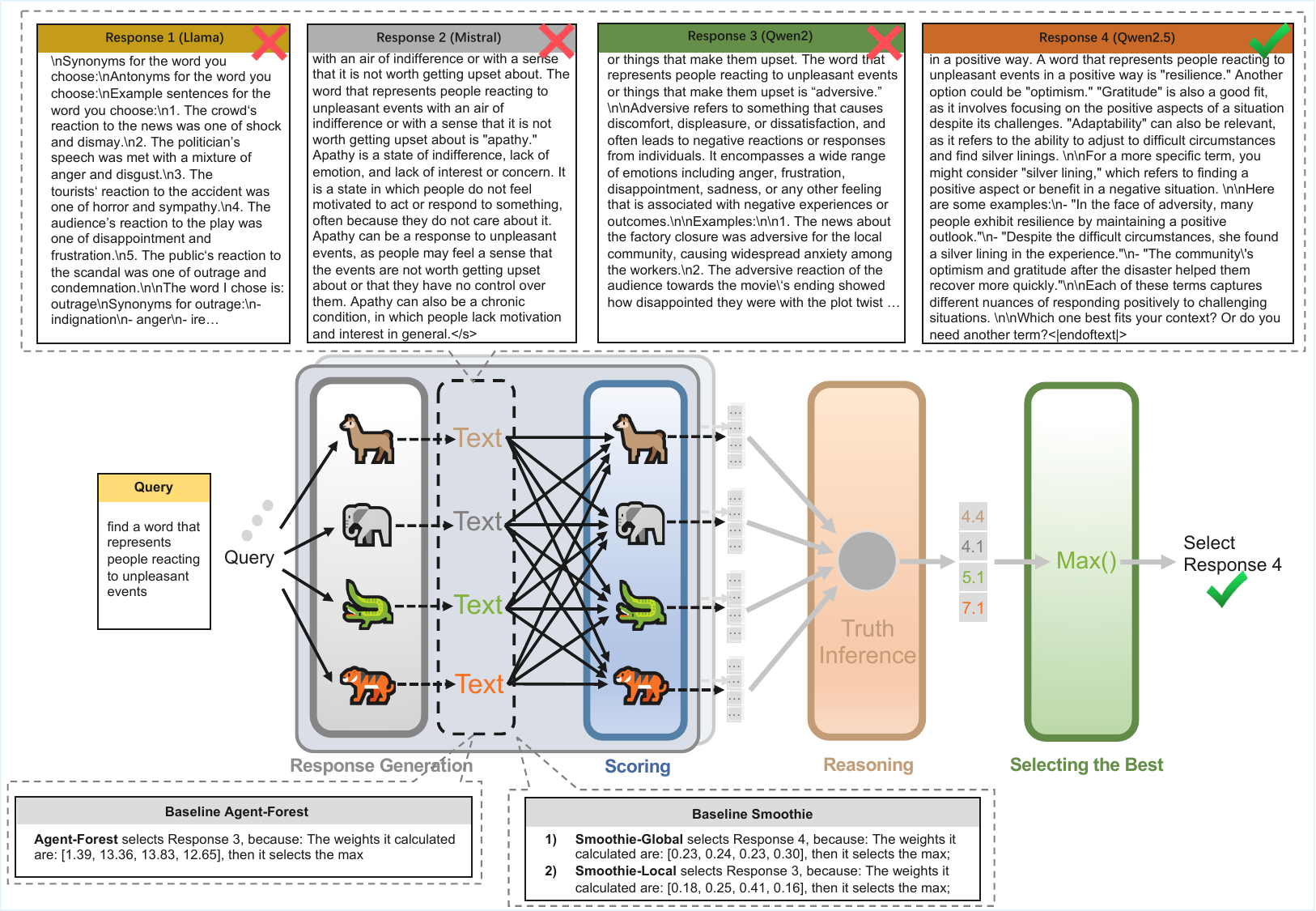} 
\caption{Case study on dataset AlpacaEval.
The ellipses “...” in the Responses represent the omission of less important content.
For this query, the best response is Response 4, and the reference answer is ``A word that represents people reacting to unpleasant events is "resilience." Resilience refers to the ability of individuals to cope with, adapt to, and recover from stress, adversity, trauma, or tragedy. It implies a form of mental toughness and flexibility that allows people to face challenges and bounce back from them.''}
\label{fig: appendix case alpacaeval}
\end{figure*}

\paragraph{(g) Case studies.}
% \paragraph{(f) Case studies.}
Subsequently, in Figures \ref{fig: appendix case qa}, \ref{fig: appendix case gsm8k},  \ref{fig: appendix case math}, and \ref{fig: appendix case alpacaeval}, we present a case study for each corresponding dataset. 
Each case study illustrates the core computational process and final results of our method, as well as those of the comparison methods, Agent-Forest~\cite{li2024more} and Smoothie~\cite{guha2024smoothie}, highlighting the key computational steps and outcomes.
We provide a more detailed analysis of the first two datasets, as shown in the captions; the patterns observed in the remaining two datasets are consistent with those in the first two datasets.

\clearpage % 强制刷新所有待处理的浮动体，并另起新页

\subsection{Prompts}
\label{sec: Prompts}
In this section, we first introduce the prompts for each task across the corresponding datasets. 
We then introduce the scoring prompts, which are unique to our method in comparison to the baseline, designed specifically for scoring responses using the LLM-as-a-Judge technique.

\begin{table*}[!hb] % 使用 [t] 确保放在页面顶部
\renewcommand{\arraystretch}{1.4}
\centering

% 1.1 TriviaQA Template
\begin{tabular}{m{0.9\linewidth}}
\begin{tcolorbox}[mybox, title=Prompt template for dataset TriviaQA.]
\vspace{-4pt}
{\fontsize{9}{9}\selectfont
Answer the question below.

Question: [placeholder of question]

Answer:
}
\vspace{-4pt}
\end{tcolorbox}
\end{tabular}
\vspace{-1mm}
% \caption*{Table A.5: Prompt template for dataset TriviaQA.}
\caption{Prompt template for dataset TriviaQA.}
\vspace{2mm} % 控制表格之间的间距

% 1.2 TriviaQA Example
\begin{tabular}{m{0.9\linewidth}}
\begin{tcolorbox}[mybox, title=A prompt example (after query instantiation) for dataset TriviaQA.]
\vspace{-4pt}
{\fontsize{9}{9}\selectfont
Answer the question below.

Question: What amateur rowing club based at Henley on Thames, England, was founded in 1818?

Answer:
}
\vspace{-4pt}
\end{tcolorbox}
\end{tabular}
\vspace{-1mm}
% \caption*{Table A.6: A prompt example (after query instantiation) for dataset TriviaQA.}
\caption{A prompt example (after query instantiation) for dataset TriviaQA.}
\vspace{2mm}

% 2.1 GSM8K Template
\begin{tabular}{m{0.9\linewidth}}
\begin{tcolorbox}[mybox, title=Prompt template for dataset GSM8K.]
\vspace{-4pt}
{\fontsize{9}{9}\selectfont
Can you solve the following math problem? [placeholder of question]. 
Explain your reasoning. Your final answer should be a single numerical number, in the form `the answer is \{answer\}', at the end of your response.
}
\vspace{-4pt}
\end{tcolorbox}
\end{tabular}
\vspace{-1mm}
% \caption*{Table A.7: Prompt template for dataset GSM8K.}
\caption{Prompt template for dataset GSM8K.}
\vspace{2mm}

% 2.2 GSM8K Example
\begin{tabular}{m{0.9\linewidth}}
\begin{tcolorbox}[mybox, title=A prompt example (after query instantiation) for dataset GSM8K.]
\vspace{-4pt}
{\fontsize{9}{9}\selectfont
Can you solve the following math problem? A wall mural has four different colors of paint in it: red, white,
purple, and yellow. There are equal amounts of red, white, and purple paint in the mural. Half the mural
is yellow. If the mural used 12 pints of paint in all, how many pints of red paint was used? Explain your
reasoning. Your final answer should be a single numerical number, in the form `the answer is \{answer\}', at the
end of your response.
}
\vspace{-4pt}
\end{tcolorbox}
\end{tabular}
\vspace{-1mm}
% \caption*{Table A.8: A prompt example (after query instantiation) for dataset GSM8K.}
\caption{A prompt example (after query instantiation) for dataset GSM8K.}
\vspace{2mm}

% 3.1 MATH Template
\begin{tabular}{m{0.9\linewidth}}
\begin{tcolorbox}[mybox, title=Prompt template for dataset MATH.]
\vspace{-4pt}
{\fontsize{9}{9}\selectfont
Can you solve the following math problem? [placeholder of question]. Explain your reasoning. Your final answer should be a single numerical number, in the form `the answer is \{answer\}', at the end of your response.
}
\vspace{-4pt}
\end{tcolorbox}
\end{tabular}
\vspace{-1mm}
% \caption*{Table A.9: Prompt template for dataset MATH.}
\caption{Prompt template for dataset MATH.}
\vspace{2mm}

% 3.2 MATH Example
\begin{tabular}{m{0.9\linewidth}}
\begin{tcolorbox}[mybox, title=A prompt example (after query instantiation) for dataset MATH.]
\vspace{-4pt}
{\fontsize{9}{9}\selectfont
Can you solve the following math problem? 
If $x$ is a constant such that $x^2+100x+c$ is equal to the square of a binomial, then what is $c$?
Explain your reasoning. Your final answer should be a single numerical number, in the form `the answer is \{answer\}’, at the end of your response.
}
\vspace{-4pt}
\end{tcolorbox}
\end{tabular}
\vspace{-1mm}
% \caption*{Table A.10: A prompt example (after query instantiation) for dataset MATH.}
\caption{A prompt example (after query instantiation) for dataset MATH.}
\vspace{2mm}

% 4.2 AlpacaEval Example
\begin{tabular}{m{0.9\linewidth}}
\begin{tcolorbox}[mybox, title=A prompt example (after query instantiation) for dataset AlpacaEval.]
\vspace{-4pt}
{\fontsize{9}{9}\selectfont
Find a word that represents people reacting to unpleasant events.
}
\vspace{-4pt}
\end{tcolorbox}
\end{tabular}
\vspace{-1mm}
% \caption*{Table A.11: A prompt example (after query instantiation) for dataset AlpacaEval.}
\caption{A prompt example (after query instantiation) for dataset AlpacaEval.}

\end{table*}

% \newpage

% \clearpage

% 1.3
\begin{table*}[!t]
\renewcommand{\arraystretch}{1.4}
\centering
\begin{tabular}{m{0.9\linewidth}} %
\begin{tcolorbox}[mybox, title=Scoring prompt template for  dataset TriviaQA.]
\vspace{-4pt}
{\fontsize{8}{8}\selectfont
You are an expert evaluator tasked with assessing qualities of the three Responses generated for a Question/Instruction from the TriviaQA (factual recall) dataset in NLP.

TriviaQA (factual recall) is a dataset designed for factual recall, containing a large number of question-answer pairs based on multiple sources such as Wikipedia and books, covering a wide range of topics including history, geography, culture, entertainment, science, and more. **Its purpose is to test how well models provide the correct facts/concepts when given factual questions.**\\

You will be provided with a Question/Instruction and three Responses generated from different models.

**Your task is to evaluate the qualities of the three Responses by focusing on the following evaluation criteria:

For each Response, regardless of whether the entire Response contains irrelevant or nonsensical content, focus solely on whether the fact/concept (including different expressions referring to the same fact/concept, such as variations in capitalization, aliases, or minor formatting issues) that can correctly answer the Question is mentioned/appears anywhere in the entire Response**:

Simply search for whether the fact/concept that can correctly answer the Question is mentioned/appears anywhere in the entire Response. If it does, the Response is considered correct.

For example, the fact ``McDonnell Douglas'' is correct for answering the Question ``First flown on the 27th July 1972, who built the F15 Eagle fighter?''. In this case, regardless of whether the entire Response contains irrelevant or nonsensical content, you should only check for the mention of the correct fact/concept in various possible expressions (e.g., ``McDonnell Douglas'', ``mcdonell douglas'', ``mcdonnell douglas aerospace'',  ``mcdonnell douglas corp'', ``mc donell douglas'', etc.). If any of these variations are present in the Response, it is considered correct.\\

In line with the above evaluation criteria, here are the scoring guidelines between [1, 2, 3, 4, 5]. Please provide **three precise integer scores between [1, 2, 3, 4, 5]** by **strictly and meticulously adhering to the following scoring guidelines**:  

1 = Regardless of whether the entire Response contains irrelevant or nonsensical content, the Response **does not contain** the fact/concept (including different expressions referring to the same fact/concept, such as variations in capitalization, aliases, or minor formatting issues) that can correctly answer the Question.

2 = Regardless of whether the entire Response contains irrelevant or nonsensical content, the Response **most likely does not contain** the fact/concept (including different expressions referring to the same fact/concept, such as variations in capitalization, aliases, or minor formatting issues)  that can correctly answer the Question.

3 = Regardless of whether the entire Response contains irrelevant or nonsensical content, the Response **seems likely (20\%-60\%) to contain**  the fact/concept (including different expressions referring to the same fact/concept, such as variations in capitalization, aliases, or minor formatting issues) that can correctly answer the Question.

4 = Regardless of whether the entire Response contains irrelevant or nonsensical content, the Response **is highly likely (60\%-99\%) to contain**  the fact/concept (including different expressions referring to the same fact/concept, such as variations in capitalization, aliases, or minor formatting issues) that can correctly answer the Question.

5 = Regardless of whether the entire Response contains irrelevant or nonsensical content, the Response **100\% contains** the fact/concept (including different expressions referring to the same fact/concept, such as variations in capitalization, aliases, or minor formatting issues) that can correctly answer the Question.\\

Note:

1) **Output format: Please return only a Python dictionary in the format {{``Score for the Response One'': x, ``Score for the Response Two'': y, ``Score for the Response Three'': z}} where x,y,z are the placeholders that you should replace with the integer scores for each Response. Do NOT include any additional opening, closing, explanation, or formatting.**

2) **Objectivity**:

- **Do not let the order of the Responses to introduce any bias in your scoring**: As an expert evaluator, you should strictly follow the scoring guidelines, ensuring that the order of the Responses does not influence your judgment in any way. You can reconsider and evaluate the Responses multiple times internally to absolutely ensure that ``order'' does not affect your final scores.

- **Please avoid any bias: Do NOT let your judgment be swayed by any conservatism or exaggeration. Carefully consider the content of the three Responses,and strictly adhere to the scoring guidelines to provide three accurate scores.**\\

Now here are the Question/Instruction, the Responses, and your judge result:

Question/Instruction: [placeholder of question]

Response One: [placeholder of response1]

Response Two: [placeholder of response2]

Response Three: [placeholder of response3]

Your Python dictionary containing the three scores:
}

\vspace{-4pt}
\end{tcolorbox}
\end{tabular}
% \vspace{-5mm}
% \caption*{Table A.12: Scoring prompt template for dataset TriviaQA.}
\caption{Scoring prompt template for dataset TriviaQA.}
\label{table:PromptTriviaQA}
\end{table*}

% 2.3
\begin{table*}[t]
\renewcommand{\arraystretch}{1.4}
\centering
\begin{tabular}{m{0.9\linewidth}} %
\begin{tcolorbox}[mybox, title=Scoring prompt template for  dataset GSM8K.]
\vspace{-4pt}
{\fontsize{8}{8}\selectfont
You are an expert evaluator tasked with assessing the qualities of three Responses to math problems from the GSM8K dataset in NLP.

You will be provided with an Instruction/Question and three Responses generated by different models.\\

Your task is to evaluate the qualities of three Responses by focusing on the following evaluation criteria:

After carefully analyzing the math problem and deriving a final answer yourself, for each Response, regardless of whether the mathematical reasoning or analysis in the Response is correct or reasonable, or whether the Response contains irrelevant, redundant or repetitive content, your evaluation should focus solely on two possible scenarios (**for each Response, only one of the following two cases applies**):

**(Case A) If the Response contains a sentence/sentences containing ``the answer is'', ``The answer is'', ``answer'', or ``Answer'':
Simply search for whether the correct answer to the math Question is mentioned/appears in any of these sentences.**

**(Case B) If the Response does not contain any sentences containing ``the answer is'', ``The answer is'', ``answer'', or ``Answer'':
Simply check whether **the LAST single numerical number in the entire Response** is the correct answer (usually a single numerical value) to the math Question.**\\

In line with the above evaluation criteria, here are the scoring guidelines between [1, 2, 3]. Please provide **three precise integer scores between [1, 2, 3]** by **strictly and meticulously adhering to the following scoring guidelines**: 

1 = Regardless of how irrelevant, unreasonable, or unformatted other content in the Response is: 

(Case A) If the Response contains a sentence/sentences containing ``the answer is'', ``The answer is'', ``answer'', or ``Answer'':
These ``answer''  sentences **do not contain** the correct answer (usually a single numerical value) to the math Question.

(Case B) If the Response does not contain any sentence containing ``the answer is'', ``The answer is'', ``answer'', or ``Answer'':
**The last numerical value** that appears in the entire Response **is not** the correct answer (usually a single numerical value) to the math Question.

2 = Regardless of how irrelevant, unreasonable, or unformatted other content in the Response is: 

(Case A) If the Response contains a sentence/sentences containing ``the answer is'', ``The answer is'', ``answer'', or ``Answer'':
These ``answer''  sentences **seems likely or very likely to contain** the correct answer (usually a single numerical value) to the math Question.

(Case B) If the Response does not contain any sentence containing ``the answer is'', ``The answer is'', ``answer'', or ``Answer'':
**The last numerical value** that appears in the entire Response **seems likely or very likely to be** the correct answer (usually a single numerical value) to the math Question.

3 = Regardless of how irrelevant, unreasonable, or unformatted other content in the Response is: 

(Case A) If the Response contains a sentence/sentences containing ``the answer is'', ``The answer is'', ``answer'', or ``Answer'':
These ``answer''  sentences **100\% contain** the correct answer (usually a single numerical value) to the math Question.

(Case B) If the Response does not contain any sentence containing ``the answer is'', ``The answer is'', ``answer'', or ``Answer'':
**The last numerical value** that appears in the entire Response **100\% is** the correct answer (usually a single numerical value) to the math Question.\\

Note:

1) **Output format: Please return only a Python dictionary in the format \{``Score for the Response One'': x, ``Score for the Response Two'': y, ``Score for the Response Three'': z\} where x,y,z are the placeholders that you should replace with the integer scores for each Response. Do NOT include any additional opening, closing, explanation, or formatting.**

2) **Objectivity**:

- **Do not let the order of the Responses to introduce any bias in your scoring**: As an expert evaluator, you should strictly follow the scoring guidelines, ensuring that the order of the Responses does not influence your judgment in any way. You can reconsider and evaluate the Responses multiple times internally to absolutely ensure that ``order'' does not affect your final scores.

- **Please avoid any bias: Do NOT let your judgment be swayed by any conservatism or exaggeration. Carefully consider the content of the three Responses,and strictly adhere to the scoring guidelines to provide three accurate scores.**\\

Now here are the Instruction/Question, the Responses, and your judge result:

Instruction/Question: [placeholder of question]

Response One: [placeholder of response 1]

Response Two: [placeholder of response 2]

Response Three: [placeholder of response 3]

Your Python dictionary containing the three scores:
}
\vspace{-4pt}
\end{tcolorbox}
\end{tabular}
% \vspace{-5mm}
% \caption*{Table A.13: Scoring prompt template for  dataset GSM8K.}
\caption{Scoring prompt template for  dataset GSM8K.}
\label{table:PromptGSM8K}
\end{table*}

% 3.3
\begin{table*}[t]
\renewcommand{\arraystretch}{1.4}
\centering
\begin{tabular}{m{0.9\linewidth}} %
\begin{tcolorbox}[mybox, title=Scoring prompt template for  dataset MATH.]
\vspace{-4pt}
{\fontsize{8}{8}\selectfont
You are an expert evaluator tasked with assessing the qualities of three Responses to math problems from the MATH dataset in NLP.

You will be provided with an Instruction/Question and three Responses generated by different models.\\

Your task is to evaluate the qualities of three Responses by focusing on the following evaluation criteria:

After carefully analyzing the math problem and deriving a final answer yourself, for each Response, regardless of whether the mathematical reasoning or analysis in the Response is correct or reasonable, or whether the Response contains irrelevant, redundant or repetitive content, your evaluation should focus solely on two possible scenarios (**for each Response, only one of the following two cases applies**):

**(Case A) If the Response contains a sentence/sentences containing ``the answer is'', ``The answer is'', ``answer'', or ``Answer'':
Simply search for whether the correct answer to the math Question is mentioned/appears in any of these sentences.**

**(Case B) If the Response does not contain any sentences containing ``the answer is'', ``The answer is'', ``answer'', or ``Answer'':
Simply check whether **the LAST single numerical number in the entire Response** is the correct answer (usually a single numerical value) to the math Question.**\\

In line with the above evaluation criteria, here are the scoring guidelines between [1, 2, 3]. Please provide **three precise integer scores between [1, 2, 3]** by **strictly and meticulously adhering to the following scoring guidelines**: 

1 = Regardless of how irrelevant, unreasonable, or unformatted other content in the Response is: 

(Case A) If the Response contains a sentence/sentences containing ``the answer is'', ``The answer is'', ``answer'', or ``Answer'':
These ``answer''  sentences **do not contain** the correct answer (usually a single numerical value) to the math Question.

(Case B) If the Response does not contain any sentence containing ``the answer is'', ``The answer is'', ``answer'', or ``Answer'':
**The last numerical value** that appears in the entire Response **is not** the correct answer (usually a single numerical value) to the math Question.

2 = Regardless of how irrelevant, unreasonable, or unformatted other content in the Response is: 

(Case A) If the Response contains a sentence/sentences containing ``the answer is'', ``The answer is'', ``answer'', or ``Answer'':
These ``answer''  sentences **seems likely or very likely to contain** the correct answer (usually a single numerical value) to the math Question.

(Case B) If the Response does not contain any sentence containing ``the answer is'', ``The answer is'', ``answer'', or ``Answer'':
**The last numerical value** that appears in the entire Response **seems likely or very likely to be** the correct answer (usually a single numerical value) to the math Question.

3 = Regardless of how irrelevant, unreasonable, or unformatted other content in the Response is: 

(Case A) If the Response contains a sentence/sentences containing ``the answer is'', ``The answer is'', ``answer'', or ``Answer'':
These ``answer''  sentences **100\% contain** the correct answer (usually a single numerical value) to the math Question.

(Case B) If the Response does not contain any sentence containing ``the answer is'', ``The answer is'', ``answer'', or ``Answer'':
**The last numerical value** that appears in the entire Response **100\% is** the correct answer (usually a single numerical value) to the math Question.\\

Note:

1) **Output format: Please return only a Python dictionary in the format \{``Score for the Response One'': x, ``Score for the Response Two'': y, ``Score for the Response Three'': z\} where x,y,z are the placeholders that you should replace with the integer scores for each Response. Do NOT include any additional opening, closing, explanation, or formatting.**

2) **Objectivity**:

- **Do not let the order of the Responses to introduce any bias in your scoring**: As an expert evaluator, you should strictly follow the scoring guidelines, ensuring that the order of the Responses does not influence your judgment in any way. You can reconsider and evaluate the Responses multiple times internally to absolutely ensure that ``order'' does not affect your final scores.

- **Please avoid any bias: Do NOT let your judgment be swayed by any conservatism or exaggeration. Carefully consider the content of the three Responses,and strictly adhere to the scoring guidelines to provide three accurate scores.**\\

Now here are the Instruction/Question, the Responses, and your judge result:

Instruction/Question: [placeholder of question]

Response One: [placeholder of response 1]

Response Two: [placeholder of response 2]

Response Three: [placeholder of response 3]

Your Python dictionary containing the three scores:
}
\vspace{-4pt}
\end{tcolorbox}
\end{tabular}
% \vspace{-5mm}
% \caption*{Table A.14: Scoring prompt template for dataset MATH.}
\caption{Scoring prompt template for dataset MATH.}
\label{table:PromptMATH}
\end{table*}

% 4.3(1)
\begin{table*}[t]
\renewcommand{\arraystretch}{1.4}
\centering
\begin{tabular}{m{0.9\linewidth}} %
\begin{tcolorbox}[mybox, title=Scoring prompt template for dataset AlpacaEval (Part 1).]
\vspace{-4pt}
{\fontsize{8}{8}\selectfont
You are an expert evaluator tasked with assessing the qualities of three Responses generated for an Instruction from the well-known instruction-following dataset AlpacaEval in NLP.

You will be provided with an Instruction and three Responses generated by different models.\\
 
Your task is to evaluate the overall qualities of three Responses by focusing on the following evaluation criteria:

Always keep in mind the ``Core Evaluation Criteria'' — ``Evaluate whether the Response effectively addresses the Instruction from a human perspective.'' Also, please focus on the following **FOUR Specific Evaluation Criteria**:

**Basic Requirement: Understanding the Question \& Response Relevance**: Does the Response understand and follow the Instruction without going off-topic or irrelevant?

**Core Requirement: Correctness, Reasonableness, Logical Consistency, and Overall Usefulness/Effectiveness**: Does the Response correctly, reasonably, and logically answer the Instruction? Overall, does the Response meet the core requirements of the Instruction in terms of usefulness and effectiveness?

**Content-Related Advanced Requirement: Completeness, Thoroughness, Depth, Creativity, and Added Value**: In cases where needed (especially for more open-ended Instructions), does the Response thoroughly cover all aspects of the Instruction's requirements, including all key points, relevant aspects, necessary information, steps for problem-solving needed to perfectly answer the Instruction? Does the Response go beyond surface-level answers and provide valuable, useful, creative, more in-depth information, offering added value and making the response more perfect?

**Language-Related Advanced Requirement: Clarity, Naturalness, Coherence, Fluency, Conciseness, and Ease of Understanding**: Does the Response use appropriate language that is clear, natural, coherent, fluent, concise, and easy to understand? Does the Response avoid vague, unnecessarily verbose, or repetitive statements? \\

In line with the above evaluation criteria, here are the scoring guidelines between [1, 2, 3, 4, 5, 6, 7, 8, 9, 10]. Please provide **three precise integer scores between [1, 2, 3, 4, 5, 6, 7, 8, 9, 10]** by **strictly and meticulously adhering to the following scoring guidelines**:

**1 = From a human perspective, overall, the Response ``fails to address the Instruction’s needs'', and it is a ``Unrelevant, Unacceptable Response''.**
Because: It fails to meet the above ``Basic Requirement: Understanding the Question \& Response Relevance'', let alone the more advanced requirements.  

**2 = From a human perspective, overall, the Response ``partially addresses the Instruction’s needs to a small extent'', and it is a ``Relevant, Low-Utility Response''.**
Because: It meets the above ``Basic Requirement: Understanding the Question \& Response Relevance'',  but  **``has too many serious and obvious shortcomings in some of the the following aspects'': ``Core Requirement: Correctness, Reasonableness, Logical Consistency, and Overall Usefulness/Effectiveness'', ``Content-Related Advanced Requirement: Completeness, Thoroughness, Depth, Creativity, and Added Value'', ``Language-Related Advanced Requirement: Clarity, Naturalness, Coherence, Fluency, Conciseness, and Ease of Understanding''**.

**3 = From a human perspective, overall, the Response ``partially addresses the Instruction’s needs to a moderate extent'', and it is a ``Relevant, Moderately Useful Response''.**
Because: It meets the above ``Basic Requirement: Understanding the Question \& Response Relevance'',  but  **``has several serious and obvious shortcomings in some of the the following aspects'': ``Core Requirement: Correctness, Reasonableness, Logical Consistency, and Overall Usefulness/Effectiveness'', ``Content-Related Advanced Requirement: Completeness, Thoroughness, Depth, Creativity, and Added Value'', ``Language-Related Advanced Requirement: Clarity, Naturalness, Coherence, Fluency, Conciseness, and Ease of Understanding''**.

**4 = From a human perspective, overall, the Response ``partially addresses the Instruction’s needs to a large extent'', and it is a ``Relevant, Useful Response''.**
Because: It meets the above ``Basic Requirement: Understanding the Question \& Response Relevance'',  but  **``has several obvious shortcomings in some of the the following aspects'': ``Core Requirement: Correctness, Reasonableness, Logical Consistency, and Overall Usefulness/Effectiveness'', ``Content-Related Advanced Requirement: Completeness, Thoroughness, Depth, Creativity, and Added Value'', ``Language-Related Advanced Requirement: Clarity, Naturalness, Coherence, Fluency, Conciseness, and Ease of Understanding''**.

**5 = From a human perspective, overall, the Response ``almost addresses the Instruction’s needs'', and it is a ``Relevant, Useful, Minimally Acceptable Response''.**
Because: It meets the above ``Basic Requirement: Understanding the Question \& Response Relevance'',  but  **``has noticeable shortcomings in some of the the following aspects'': ``Core Requirement: Correctness, Reasonableness, Logical Consistency, and Overall Usefulness/Effectiveness'', ``Content-Related Advanced Requirement: Completeness, Thoroughness, Depth, Creativity, and Added Value'', ``Language-Related Advanced Requirement: Clarity, Naturalness, Coherence, Fluency, Conciseness, and Ease of Understanding''**.

**6 = From a human perspective, overall, the Response ``sufficiently addresses the Instruction’s needs'', and it is a ``Relevant, Useful, Acceptable Response.**
Because: It meets the above ``Basic Requirement: Understanding the Question \& Response Relevance'',  and it  **``has minor, subtle shortcomings in some of the the following aspects'': ``Core Requirement: Correctness, Reasonableness, Logical Consistency, and Overall Usefulness/Effectiveness'', ``Content-Related Advanced Requirement: Completeness, Thoroughness, Depth, Creativity, and Added Value'', ``Language-Related Advanced Requirement: Clarity, Naturalness, Coherence, Fluency, Conciseness, and Ease of Understanding''**.

**7 = From a human perspective, overall, the Response ``effectively addresses the Instruction’s needs'', and it is a ``Relevant, Useful, Acceptable, Good Response''.**
Because: It meets the ``Basic Requirement: Understanding the Question \& Response Relevance'', and **``there may be minor room for improvement in the following criteria'': ``Core Requirement: Correctness, Reasonableness, Logical Consistency, and overall Usefulness/Effectiveness'', ``Advanced Requirement One (Content-Related): Completeness, Thoroughness, Depth, Creativity, and Added Value'', ``Advanced Requirement Two (Language-Related): Clarity, Naturalness, Coherence, Fluency, Conciseness, and Ease of Understanding''**.

**8 = From a human perspective, overall, the Response ``excellently addresses the Instruction’s needs'', and it is a ``Relevant, Useful, Strongly Acceptable, Excellent Response''.**
Because: It excels in all ``FOUR Specific Evaluation Criteria'', and there seems to be no room for improvement.

**9 = From a human perspective, overall, the Response ``perfectly addresses the Instruction’s needs``, and it is a ``Relevant, Useful, Strongly Acceptable, Perfect Response''.** 
Because: It fully excels in all above FOUR Specific Evaluation Criteria, with NO room for improvement.

**10 = From a human perspective, overall, the Response ``exceeds expectations and very perfectly addresses'' the Instruction’s needs, and it is a ``Relevant, Useful, Very Strong Acceptable, Very Perfect Response''.**
Because: It fully excels in all above FOUR Specific Evaluation Criteria, with NO room for improvement.
}
\vspace{-4pt}
\end{tcolorbox}
\end{tabular}
% \vspace{-5mm}
% \caption*{Table A.15: Scoring prompt template for dataset AlpacaEval (Part 1).}
\caption{Scoring prompt template for dataset AlpacaEval (Part 1).}
\label{table:PromptAlpacaEval}
\end{table*}

% 4.3(2)
\begin{table*}[t]
\renewcommand{\arraystretch}{1.4}
\centering
\begin{tabular}{m{0.9\linewidth}} %
\begin{tcolorbox}[mybox, title=Scoring prompt template for dataset AlpacaEval (Part 2).]
\vspace{-4pt}
{\fontsize{8}{8}\selectfont
Note:

1) **Output format: Please return only a Python Dictionary, 100\% STRICTLY following in the format {{``Score for the Response One'': x, ``Score for the Response Two'': y, ``Score for the Response Three'': z}} where x,y,z are the placeholders that you should replace with the integer scores for each Response. NOTE: You are NOT allowed to modify the ``key'' of the Dictionary under any circumstances. Do NOT include any additional opening, closing, explanation, or formatting.**

2) **Objectivity**:

- **Do not let the order of the Responses to introduce any bias in your scoring**: As an expert evaluator, you should strictly follow the scoring guidelines, ensuring that the order of the Responses does not influence your judgment in any way. You can reconsider and evaluate the Responses multiple times internally to absolutely ensure that ``order'' does not affect your final scores.

- **Please avoid any bias: Do NOT let your judgment be swayed by any conservatism or exaggeration. Do NOT allow the length of the responses to influence your evaluation. Carefully consider the content of the three Responses, and strictly adhere to the scoring guidelines to provide three precise scores.**\\

Now here are the Instruction, the Responses, and your judge result:

Instruction: [placeholder of question]

Response One: [placeholder of response1]

Response Two: [placeholder of response2]

Response Three: [placeholder of response3]

Your Python dictionary containing the three scores:
}

\vspace{-4pt}

\end{tcolorbox}
\end{tabular}
% \vspace{-5mm}
% \caption*{Table A.16: Scoring prompt template for dataset AlpacaEval (Part 2).}
\caption{Scoring prompt template for dataset AlpacaEval (Part 2).}
\label{table:PromptAlpacaEval}
\end{table*}

%%%%%%%%%%%%%%%%%%%%%%%%%%%%%%%%%%%%%%%%%%%%%%%%%%%%%%%%%%%%%%%%%%%%%%%%%%%%%%%
%%%%%%%%%%%%%%%%%%%%%%%%%%%%%%%%%%%%%%%%%%%%%%%%%%%%%%%%%%%%%%%%%%%%%%%%%%%%%%%


% You can have as much text here as you want. The main body must be at most $8$
% pages long. For the final version, one more page can be added. If you want, you
% can use an appendix like this one.

% The $\mathtt{\backslash onecolumn}$ command above can be kept in place if you
% prefer a one-column appendix, or can be removed if you prefer a two-column
% appendix.  Apart from this possible change, the style (font size, spacing,
% margins, page numbering, etc.) should be kept the same as the main body.
%%%%%%%%%%%%%%%%%%%%%%%%%%%%%%%%%%%%%%%%%%%%%%%%%%%%%%%%%%%%%%%%%%%%%%%%%%%%%%%
%%%%%%%%%%%%%%%%%%%%%%%%%%%%%%%%%%%%%%%%%%%%%%%%%%%%%%%%%%%%%%%%%%%%%%%%%%%%%%%

\newpage

\end{document}